%% file: iclr2027_conference.tex
\documentclass{applemlr} %

\usepackage{caption}

\input{math_commands.tex}

\usepackage{hyperref}
\usepackage{url}
\usepackage{graphicx}
\usepackage{adjustbox}
\usepackage{booktabs}
\usepackage{xspace}

\usepackage[T1]{fontenc}
\usepackage[utf8]{inputenc}

\usepackage{float}
\usepackage{comment}
\usepackage{tablefootnote}
\usepackage{wrapfig}

\usepackage{xstring}

\usepackage{multirow}

\title{How Much of a Harness Does a Strong Agent Need for Autonomous ML Engineering?}

\DeclareRobustCommand{\sfacute}[1]{\ooalign{#1\cr\hidewidth{\fontfamily{phv}\selectfont\char1}\hidewidth\cr}}
\author[1,2,*]{Kirill Brilliantov}
\author[1,2,*,\dagger]{Alejandro Hern\sfacute{a}ndez-Cano}
\author[1,2]{Emmanuel Abb\sfacute{e}}

\affiliation[1]{EPFL}
\affiliation[2]{Apple}

\contribution[*]{Equal contribution}
\contribution[\dagger]{Work done while interning at Apple}

\date{\sffamily\today}

\newcommand{\chat}{Chat\xspace}
\newcommand{\oneshot}{Oneshot\xspace}
\newcommand{\agentic}{Malena\xspace}

\newcommand{\Chain}{Chain\xspace}
\newcommand{\Greedy}{Greedy\xspace}
\newcommand{\UCB}{UCB1\xspace}
\newcommand{\Bootstrap}{Best-of-N\xspace}

\newcommand{\malena}{Malena\xspace}

\newcommand{\tasksplit}[1]{%
	\IfStrEqCase{#1}{%
		{30}{\texttt{fixed30}}%
		{29}{\texttt{default29}}%
		{14}{\texttt{fixed14}}%
		{10}{\texttt{fixed10}}%
	}[unknown]%
}
\usepackage{afterpage}

\definecolor{textgray}{HTML}{6E6E73}
\makeatletter
\newcommand\applefootnote[1]{%
  \begingroup
  \renewcommand\thefootnote{}%
  \renewcommand\@makefntext[1]{\noindent##1}%
  \footnote{#1}%
  \addtocounter{footnote}{-1}%
  \endgroup
}
\makeatother

\abstract{
Recent autonomous machine learning engineering (MLE) agents have made significant progress on public leaderboards.
Often motivated by progress stagnation over long-horizon cycles and limited Large Language Model (LLM) primitives, modern MLE agents are deployed on top of increasingly elaborate machinery: multi-agent orchestrators, dedicated retrieval subagents, and more.
While such harnesses expand, the use of more primitive but improved coding agents --- where LLMs have direct access to the execution environment through \texttt{read}, \texttt{write}, and \texttt{bash} primitives --- has received little attention in the field.
In this paper we find that, under an equal time budget and the same frontier LLM backbone, open-source state-of-the-art harnesses provide no advantages over a single session of a minimal-harness coding agent baseline, pointing to the backbone as the primary driver for performance.
Via a series of large-scale systematic ablation studies, we argue that the machinery layers become redundant in the coding agent setting.
We conclude that the effort spent elaborating hand-crafted harnesses around strong models yields poor returns for current MLE benchmarks.
}

\begin{document}

\maketitle

\section{Introduction}
\label{sec:intro}

Autonomous machine learning engineering (MLE) has become a demanding testbed for LLM-based agents: solutions are open-ended ML pipelines, a single evaluation can take hours on expensive hardware, and the feedback signal --- a validation score --- is noisy.
Since \citet{jiang2025aideaidrivenexplorationspace} framed the problem as tree search over code, the dominant design has been the \textit{harness}: an outer program that queries an LLM for code, executes it, and orchestrates the exploration and combination of candidate solutions.
Motivated by stagnation over long horizons and by the limits of a single LLM call, harnesses have grown steadily more elaborate, adding search trees, populations of programs, memory hierarchies, and teams of specialized agents.

Much of this machinery rests on two premises that deserve reconsideration.
First, a single chat completion cannot run the code it writes, inspect the data, or react to an error, so the harness has to do it on its behalf.
Modern LLMs, however, are increasingly post-trained for tool use and can operate inside \textit{coding agent} environments such as OpenCode~\citep{opencode}, where they read and write files, execute code, and debug within one self-contained session.
Second, MLE is treated as a long-horizon problem that calls for dedicated memory and context management.
Measured in wall-clock time, up to and beyond 24h, it is.
Measured in tokens, it is less clear, now that million-token context windows are affordable.
This raises a simple question: which parts of the harness still earn their keep?

\begin{figure}[!t]
\centering
\includegraphics[width=\linewidth]{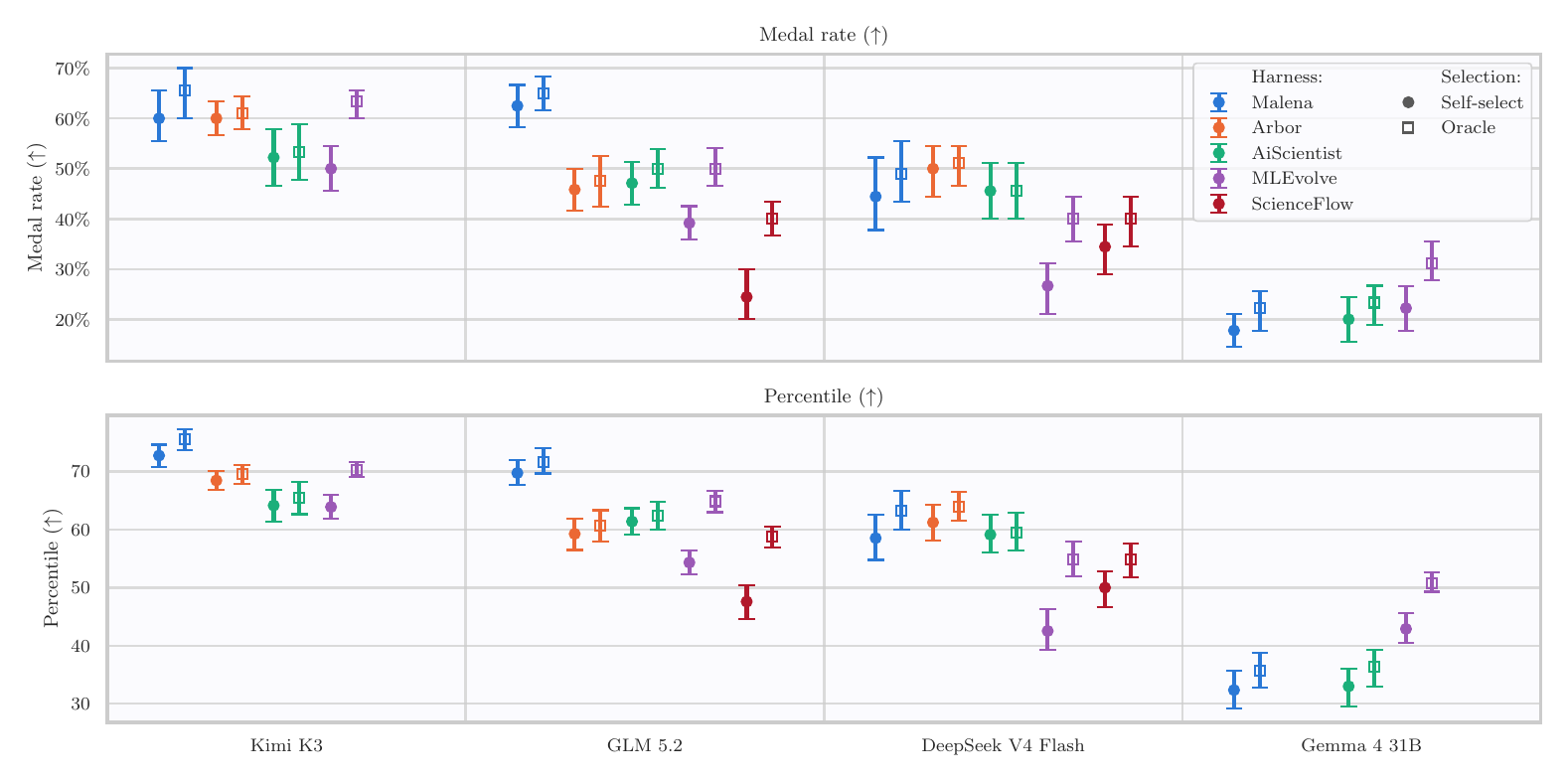}
\caption{\textbf{Performance across backbones and harnesses.}
Any-medal rate ($\uparrow$, top) and mean percentile ($\uparrow$, bottom), self-select
(filled) vs.\ oracle (open), macro-averaged over the full \protect\tasksplit{30}
set at 24h time budget. Brackets are 95\% bootstrap CIs. Backbones are grouped along
a single shared x-axis; color identifies the harness.}
\label{fig:headline-bars}
\end{figure}

The current literature is poorly placed to answer it.
Harness performance has risen alongside backbone capability, so gains are hard to attribute to harness design.
On MLE-bench~\citep{chan2025mlebenchevaluatingmachinelearning}, the standard testbed, published results are almost invariably reported under different backbones and hardware, seed counts rarely exceed three, and headline margins can be smaller than the run-to-run spread~\citep{hambardzumyan2026aira2overcomingbottlenecksai,zhang2026reasoninggradientscalingmle,yang2026frontisma1trainingai4aimodel}.
This is a matter of cost rather than neglect: a single 24h evaluation of three seeds over the 75 tasks already consumes thousands of GPU hours, so faithful re-runs of existing baselines are rare.

Evidence from neighbouring domains suggests the answer may be ``very little''.
Across 35 releases of software-engineering harnesses, \citet{sghaier2026dontblamelargelanguage} find no statistically significant improvement over a fixed backbone on SWE-bench~\citep{jimenez2024swebenchlanguagemodelsresolve}; a single well-prompted agent can match multi-agent workflows~\citep{xu2026rethinkingvaluemultiagentworkflow}; multi-agent failures stem from coordination rather than capability~\citep{cemri2025multiagentllmsystemsfail,orogat2026understandingmultiagentllmframeworks}; and orchestration can introduce substantial overhead and coordination failures~\citep{orogat2026understandingmultiagentllmframeworks}.
Coding agents have accordingly become a standard, strong baseline in agentic evaluation~\citep{sghaier2026dontblamelargelanguage,wang2026rethinkingevaluationharnessevolution,li2026benchmarktesttimescalinggeneral}.
In MLE, however, this practice has arrived only recently~\citep{jin2026generalistautonomousresearchhypothesistree,yang2026frontisma1trainingai4aimodel,chen2026autonomouslonghorizonengineeringml,zhao2026scienceflowlonghorizonagentml}, and the coding agent comparisons reported so far are sparsely documented.

We evaluate on a harder and broader task set than Lite and, holding backbone, hardware and time budget fixed, climb a ladder of harness interventions re-implemented in a single codebase (Section~\ref{sec:systematic}): from a chat-based iteration to multi-agent coding agent orchestration. %
The end point of this ladder is \malena, the \textbf{ma}chine \textbf{l}earning \textbf{en}gineer \textbf{a}gent baseline: a single well-prompted coding agent session with a minimal set of tools.
Giving the model a coding agent environment is by far the largest effect we measure; beyond it, no intervention significantly improves performance.
We then test whether this holds for systems built specifically for MLE (Section~\ref{sec:confirmation}): under matched conditions, \malena matches or outperforms four open-source state-of-the-art harnesses at every frontier backbone we test (Figure~\ref{fig:headline-bars}), on MLE-bench as well as on NatureBench~\citep{wang2026naturebenchcodingagentsmatch}.
Finally, a trace analysis (Section~\ref{sec:trace}) suggests why: the agent carries out the search these harnesses hard-code on its own, balancing rare techniques against performance-oriented refinement while reusing the artifacts it has already produced.

Our contributions are:
\begin{itemize}
	\item A controlled, budget- and backbone-matched ablation of the intervention families common to MLE harnesses --- execution environment, search, autonomy, and multi-agent orchestration --- showing that the coding agent environment is the dominant factor and that no further intervention yields a statistically significant gain (Section~\ref{sec:systematic}).
	\item A matched comparison against four open-source MLE harnesses on MLE-bench and NatureBench across multiple backbones, in which \malena is not outperformed at any frontier backbone (Section~\ref{sec:baselines-mlebench}).
	Our results suggest that coding-agent post-training substitutes for the scaffolding that harnesses hand-design: weaker backbones still benefit from hand-crafted workflow priors, but as agentic capability grows, a plain coding agent loop closes the gap, echoing the bitter lesson~\citep{sutton2019bitterlesson}.
	\item A trace analysis showing that a capable coding agent performs adaptive search without harness-imposed structure (Section~\ref{sec:trace}).
\end{itemize}

\section{Related work}

\citet{jiang2025aideaidrivenexplorationspace} target autonomous MLE tasks with tree search over single-file programs.
Subsequent systems follow a common template: an outer harness prompts an LLM for code, and handles execution, scoring and exploration~\citep{yang2025rdagentllmagentframeworkautonomous,nam2025mlestarmachinelearningengineering,fang2025mlzeromultiagentendtoendmachine,du2025automlgennavigatingfinegrainedoptimization,kulibaba2026kompeteaiacceleratedautonomousmultiagent,li2026fmagent,yuan2025archpilotproxyguidedmultiagentapproach,qu2026coralautonomousmultiagentevolution,le2026imlexecutableproblemgroundedbroadly,nadafian2026kapsoknowledgegroundedframeworkautonomous,choi2026idlespecexploitingidletime,meng2026scientistonehumanlevelautonomousresearch,fu2026solutioncentricsearchadaptiveinquiry}.
The components these harnesses usually introduce cluster into a small number of claims about what LLMs cannot be trusted to do unaided.

Several systems devote their principal contribution to compressing or partitioning accumulated state~\citep{zhu2026ultralonghorizonagenticsciencecognitive,zhao2026scienceflowlonghorizonagentml,chen2026autonomouslonghorizonengineeringml,jin2026generalistautonomousresearchhypothesistree}.
Most remaining interventions shape the search topology, either through validation-score-driven Monte-Carlo tree search or bandit rules~\citep{jiang2025aideaidrivenexplorationspace,du2026mlevolveselfevolvingframeworkautomated,toledo2025airesearchagentsmachine,zhang2026reasoninggradientscalingmle}, or through agent-driven hypothesis search and multi-agent orchestration~\citep{chen2026autonomouslonghorizonengineeringml,zhao2026scienceflowlonghorizonagentml,jin2026generalistautonomousresearchhypothesistree,chen2026marsmodularagentreflective,qiang2026matryoshkaagentunfoldingsubagents}.
Even in systems that grant agents more autonomy, the harness still largely dictates how the search should be conducted, limiting adaptability.

A parallel line of work instead modifies the backbone itself, training it for MLE with supervised fine-tuning and reinforcement learning~\citep{yang2026frontisma1trainingai4aimodel,cai2026acegrpoadaptivecurriculumenhanced,zhou2026syntheticsandboxtrainingmachine,qiang2026matryoshkaagentunfoldingsubagents,zhang2026learningideatemachinelearning}, supported by dedicated training and evaluation infrastructure~\citep{qiang2025mledojointeractiveenvironmentsempowering,zou2026fmlbenchbenchmarkingmachinelearning,lyu2026mlsbenchholisticrigorousassessment}.
However, such systems are hard to scale, given the cost of collecting 24h traces and training on such trajectories~\citep{zhou2026syntheticsandboxtrainingmachine,cai2026acegrpoadaptivecurriculumenhanced}.
In this study, we focus on the interventions present in hand-crafted harnesses.

\section{Experimental settings}
\label{sec:settings}
We build our harness on the OpenCode v1.15.6 coding agent framework~\citep{opencode}, and our primary benchmark is MLE-bench~\citep{chan2025mlebenchevaluatingmachinelearning}, with task selection described in Appendix~\ref{app:data}.
When reporting performance of a method that produces several submissions, we distinguish between \emph{self-selection}, where the reported submission is the one with the best validation score reported by the agent, and \emph{oracle} selection, where it is the one with the best hidden test score.
Unless otherwise specified, we report self-selected performance. %
Percentile indicates the percentage of human leaderboard entries that score worse than the agent.
We report 95\% Confidence Intervals (CIs) on macro-task averages, holding the set of tasks fixed and bootstrapping on the seeds available for each task.
Unless otherwise stated, we report results with the GLM 5.2 backbone; Appendix~\ref{app:models} lists all models used.
Across all experiments, agents' resources stay fixed at 1$\times$A100 80GB, 12 CPU cores and 144GB of RAM; backbones are served locally with vLLM~\citep{kwon2023efficient}.

\section{What interventions actually matter?}
\label{sec:systematic}

To study which components of MLE harnesses matter, we run a series of systematic ablations that attribute performance to individual interventions.
We identify four broad families of such interventions and study their effect. %
Each one isolates a class of intervention employed by contemporary harnesses, re-implemented inside a single codebase so that only the intervention varies.
For each intervention, we instantiate the simplest faithful version of the published component rather than any one system's variant, so that any effect is attributable to the class of intervention rather than to an implementation detail.
End-to-end fidelity to specific published systems is deferred to Section~\ref{sec:confirmation}, where we run those systems as released.

Across all intervention families studied in this work, the coding agent environment is the highest-leverage intervention an MLE system can employ.
We show empirically that additional machinery provides no statistically significant advantage as the coding agent is given more autonomy.
Furthermore, in Section~\ref{sec:results} we argue that every common additional harness intervention can plausibly be dismissed as redundant given a coding agent environment, and further hypothesize that the problems such interventions are usually designed to solve become null under the common MLE benchmark setting with modern LLMs.

\subsection{Coding agent environment}
\label{sec:layer1}

Many MLE systems call LLMs through a \emph{chat}-like interface, where the LLM receives a single prompt and can only generate text, with no direct interaction with the execution environment.
Building a successful MLE system from this primitive requires sophisticated machinery, often including dedicated draft/debug/improve agents, careful context management, retry loops, and hand-crafted data-crawler scripts that give the agent enough information about the data to solve the task.
A coding agent environment, by contrast, exposes file manipulation and execution primitives directly to the LLM, making it possible to greatly simplify such harnesses.
We measure the impact of the coding agent environment by directly comparing the effectiveness of both interfaces.
We ablate two different \textit{iterations}, which serve as atomic units of work:

\begin{description}
	\item[\chat] The LLM is called through a chat API with a single problem prompt.
	The agent is handed the task description and an overview of the data structure and file contents.
	It must produce a complete solution script as text in a single response.
	If the script fails to execute, the agent is queried again with the error stack trace for debugging.
	\item[\oneshot] A single coding agent session is spawned.
	The agent starts in a working directory containing the data and is free to use filesystem and code-execution primitives to explore, write, execute and debug.
	In addition to the ML pipeline, the agent is tasked to document its solution in a markdown file.
\end{description}
Both iterations produce a single submission file within a six-hour time limit, and their prompts share a common template, deviating only in environment-specific instructions.
The submission is later graded and the hidden test scores are never communicated back to the agents.
Implementation details are given in Appendix~\ref{app:iterations}.

Figure~\ref{fig:coding-agent} shows the impact of the coding agent environment on MLE tasks.
We observe that the performance of frontier LLMs --- GLM 5.2 and Kimi K3 --- dramatically improves when they get direct access to the execution environment.
Interestingly, even weaker models such as Gemma 4 benefit, suggesting that tool-use fluency matters more than raw model size in this setting.
This is further supported by the larger improvement of the DeepSeek V4 models compared with their Preview versions.
We study the effect of the time budget on \oneshot iterations in Appendix~\ref{app:timeouts}.

\begin{figure}[!t]
\centering
\includegraphics[width=\linewidth]{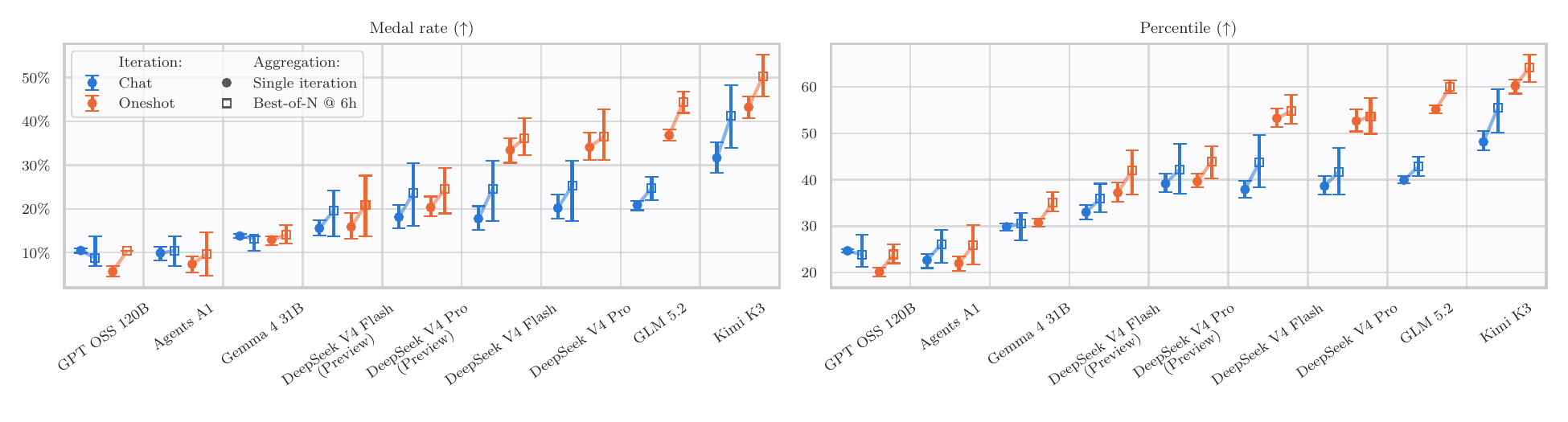}
\caption{\textbf{Impact of a coding agent environment.}
Average percentile and medal rate obtained by \chat and \oneshot iterations under different LLM backbones in the \protect\tasksplit{29} task set.
Circles represent single-iteration performance, while squares indicate Best-of-N performance at six-hour budgets.}
\label{fig:coding-agent}
\end{figure}

\subsection{Search primitives}
\label{sec:layer2}

MLE systems usually explore the solution space with tree or graph search over a set of ML pipelines.
Previous work~\citep{toledo2025airesearchagentsmachine} has explored the impact of different search algorithms for chat-based AIDE-like systems and found that careful co-design of search policy and instructions improves performance.
We test whether such findings extend to coding agent systems by chaining \oneshot iterations to refine their solutions at a matched 24h budget under different tree search algorithms (details in Appendix~\ref{app:search-details}):
\begin{description}
    \item[\Chain] always picks the most recent node as parent and represents pure iterative improvement. \item[\Greedy] picks the node with the best validation score as parent and represents pure exploitation. \item[\UCB] applies an upper-confidence-bound rule over visits and validation scores and represents the classical explore-exploit trade-off, using standard rank-normalized scores.
    \item[\Bootstrap] starts in an empty tree with no information from previous iterations and represents diverse, completely independent exploration.
\end{description}
\afterpage{\begin{figure}[!t]
\centering
\includegraphics[width=\linewidth]{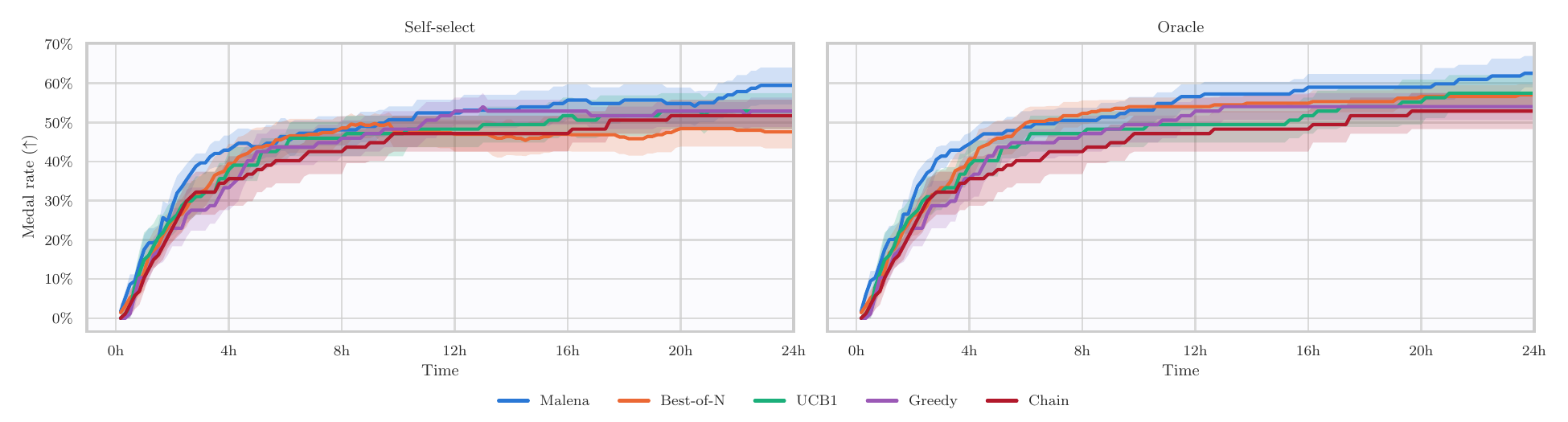}
\caption{\textbf{Performance evolution at 24h budget.}
Trajectories are extracted by selecting the best submission available at different times, aggregated in the \protect\tasksplit{29} task set.
Left: self-select medal rate.
Right: oracle-best medal rate.
\agentic achieves the highest medal rate.
\Chain, \Greedy and \UCB perform similarly.}
\label{fig:evo24h}
\end{figure}}

Figure~\ref{fig:evo24h} shows that the search strategy has a very small effect on performance, with largely overlapping CIs.
Interestingly, \Bootstrap starts strongly but slows down later on, which may indicate a ceiling beyond which refinement is needed to improve on certain tasks.
In addition to per-method marginal CIs, we compare each pair of search methods using paired-by-task 95\% CIs on their performance difference.
Appendix~\ref{app:statistics} describes our statistical protocol, and Appendix~\ref{app:pairwise-search} gives detailed results.

Under the paired-by-task analysis, no strategy provides a statistically significant advantage in either percentile or medal rate.
The pair with the largest gap is \Bootstrap and \UCB, with \UCB ahead by an average of 2.71 percentile points (pp), 95\% CI $(-1.44, +6.60)$.
We conclude that, given documentation of previous solutions and tools that facilitate data and execution exploration, the agent can already weigh different approaches and implicitly adapt its search based on previous results, as we illustrate in Section~\ref{sec:trace}.

Finally, we note that \Bootstrap's self-selected performance falls 6.832pp short of its oracle (ceiling).
This corresponds to the largest selection gap among all explored search strategies.
Part of this gap comes from noise and is unavoidable~\citep{hambardzumyan2026aira2overcomingbottlenecksai}, but strategies to reduce it should be a priority when designing MLE harnesses.
Appendix~\ref{app:hce} explores a few mitigation strategies in our controlled environment.
We leave an in-depth exploration of more strategies for future work.

\subsection{Autonomy}
\label{sec:layer3}

In the previous ablations, the search is orchestrated manually by the harness.
In this section, we remove such orchestration altogether and arrive at an even simpler baseline, which we label \agentic.
The \agentic iteration is also built on the coding agent infrastructure but, instead of being prompted to produce a single submission, runs a single session spanning the entire time budget, only briefly prompted to continue working whenever a turn ends.
This lets the agent drive exploration, exploitation and its own knowledge base autonomously.
We build a minimal harness around it, providing a small set of tools that may help the agent stay grounded as the session grows:\footnote{We ablate the importance of such tools in Appendix~\ref{app:malena-infra}, and provide further details of them in Appendix~\ref{app:infra-details}.}
\begin{itemize}
	\item \texttt{submission} tool: Adds a submission file to the registry.
	No test or leaderboard scores are communicated to the agent.
	This tool is not needed in \chat and \oneshot iterations, whose single submission is added to the registry by the harness.
	\item[\texttt{S}]  \texttt{system} tool: Returns the current hardware resources, utilization, and remaining time.
	Note that the available resources and time budget are also present in the prompt of all iterations (including \chat and \oneshot).
	\item[\texttt{J}] \texttt{jobs} tools: Allow the agent to run multiple bash commands in the background, inspect their stdout separately, wait for their completion, and cancel ongoing jobs.
\end{itemize}

As seen in Figure~\ref{fig:evo24h}, the \agentic iteration reaches the highest average medal rate.
Paired-by-task analysis indicates a significant medal rate advantage over all manual search algorithms, with 95\% CI lower bounds between 0.4\% and 5.6\% depending on the method.
Despite having also better average percentile, when compared with \UCB and \Greedy it becomes statistically indistinguishable.
Regardless, \agentic exhibits the lowest validation gap of all iterations at 1.876pp, presumably as reusing the exact validation pipeline between attempts allows to maintain higher consistency between internal validation scores than tree-search methods.
We further explore the effect of \malena under different time budgets in Appendix~\ref{app:malena-scaling} and the effect of backbone models in  Appendix~\ref{app:malena-models}.
One possible drawback of this single-session strategy arises in the long-horizon regime, where the context may overflow the LLM's window.
To test this, Appendix~\ref{app:context-management} artificially caps \malena's context window. %

\subsection{Multi-agent orchestration}
\label{sec:layer5}

MLE harnesses often guide exploration using several, potentially parallel, agents.
Such agents are typically organized either hierarchically or as a network of peers.
We attempt to isolate the effectiveness of such procedures by introducing three coordination interventions:
\begin{itemize}
	\item[\texttt{D}] Delegation.
	Enables background subagents via the native \texttt{task} tool and encourages their use through a prompt addition.
	This represents the hierarchical planner--executor structure.
	\item[\texttt{P3}] Parallelism.
	When enabled, the harness spawns three independent \malena iterations on the same physical machine.
	While this allows more submissions and ideas to be tested, it also risks hardware contention.
	Parallelism represents the network of peers.
	\item[\texttt{B}]  \texttt{broadcast} tool.
	Enabled jointly with parallelism.
	Allows agents to broadcast messages to their peers; messages are queued and delivered to each recipient when its current turn ends.
\end{itemize}

We present the results in Table~\ref{tab:agentic-arms}.
The base \agentic iteration performs no significantly worse than any intervention tested.
While the parallel addition provides a higher ceiling in oracle-selected percentile score, the selection gap of independent \agentic iterations increases under such setting so much that the final self-select submission is of similar quality.
This suggests that the multi-agent orchestration introduced in previous work is less critical for modern frontier LLMs operating in coding agent environments.
How to properly coordinate strong agents towards a common goal remains an open question, and we leave a deeper examination for future work.

\begin{table}[H]
\centering
\captionsetup{singlelinecheck=false}
\input{tables/table_agentic.tex}
\caption{\textbf{Performance of \agentic under different ablations on the \protect\tasksplit{14} split at 24h budget.}
The 95\% CI for each metric is reported in brackets.
\texttt{+} indicates interventions on top of \texttt{base}, \texttt{-} indicates intervention removals.}
\label{tab:agentic-arms}
\end{table}

\section{Generalization to production harnesses}
\label{sec:confirmation}
\label{sec:results}

We test whether the conclusions of Section~\ref{sec:systematic} generalize to a wider selection of open-source state-of-the-art harnesses.
We compare our base \agentic single long-session loop with MLEvolve~\citep{du2026mlevolveselfevolvingframeworkautomated}, AiScientist~\citep{chen2026autonomouslonghorizonengineeringml}, Arbor~\citep{jin2026generalistautonomousresearchhypothesistree}, and ScienceFlow~\citep{zhao2026scienceflowlonghorizonagentml};
full descriptions of each baseline, and the full per-backbone MLE-bench comparison table, are in Appendix~\ref{app:baselines}.
The selected methods span a wide range of harness design choices along the axes we ablated. %
MLEvolve represents the older paradigm, in which the minimal unit of work is a
single completion; the others are more recent and emphasize
orchestrating multiple agentic sessions with tools.
\agentic itself is a single such session, so the comparison is not
chat-vs-agent but rather a minimal single-session harness against harnesses that add
scaffolding \emph{on top of} agentic sessions or completions.
Crucially, all methods are evaluated under the same compute resources, time budget and backbone models.\footnote{Every harness runs at its backbone's maximum reasoning effort, except MLEvolve on DeepSeek V4 Flash, which runs at medium effort; see Appendix~\ref{app:mlevolve-dsv4-reasoning} for details.}
In addition to MLE-bench, we report results on NatureBench~\citep{wang2026naturebenchcodingagentsmatch}, a benchmark of open
scientific-research tasks that is newer and less saturated by published
baselines than MLE-bench; Appendix~\ref{app:naturebench} describes it in full.

\subsection{Results}
\label{sec:baselines-mlebench}
\begin{wraptable}[12]{r}{0.58\linewidth}

\centering
\vspace{-1em}
\caption{\textbf{NatureBench results.} Surpassed-SOTA rate ($\uparrow$, \% of tasks where the
oracle-best run surpasses the published SOTA), macro-averaged over tasks. A run
with no valid submission counts as a failure. Brackets are 95\% CIs.}
\label{tab:naturebench-glm}
\input{tables/table_naturebench_from_csv.tex}
\end{wraptable}

\begin{figure}[t]
\centering
\includegraphics[width=0.48\linewidth]{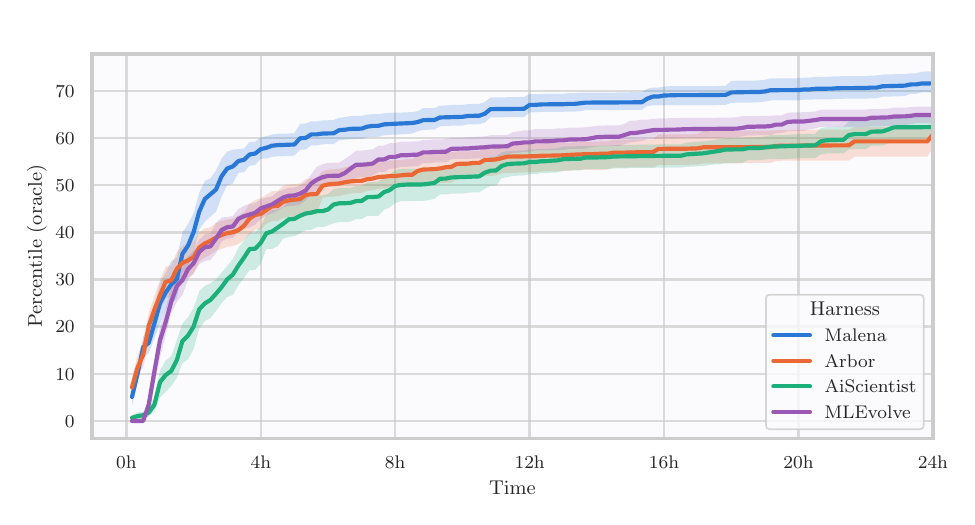}
\hfill
\includegraphics[width=0.48\linewidth]{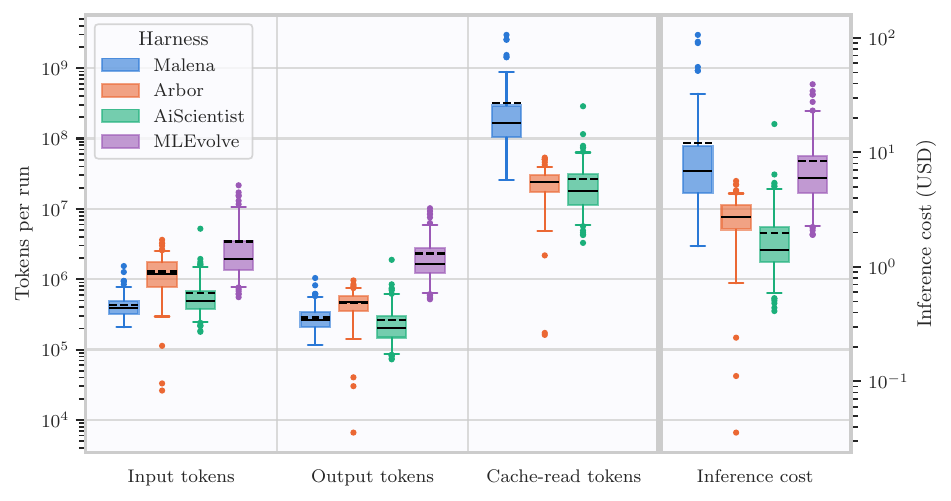}
\caption{\textbf{Left:} oracle best-so-far percentile vs.\ wall-clock time, macro-averaged over
\protect\tasksplit{30} tasks, for different harnesses.
\textbf{Right:} per-run distribution over all seeds of cumulative input,
output and modelled cache-read tokens by the end of each run, plus a fourth group
giving the resulting per-run USD cost (right-hand axis).
Each box's solid black line indicates median; dashed line indicates mean.
MLEvolve's cache is always exactly 0, since its tree-search nodes are independent LLM calls.
All results shown are at the GLM 5.2 backbone.}
\label{fig:trajectory-average}
\end{figure}

Table~\ref{tab:naturebench-glm}, Figure~\ref{fig:headline-bars}, and Figure~\ref{fig:trajectory-average} summarize the headline results: across 17 harness--backbone pairs on the 30 MLE-bench tasks and 6 pairs on the 40 NatureBench tasks (full specification in Appendix~\ref{app:baselines}), \agentic matches or beats every harness on every tested pair. Only at the smaller Gemma 4 31B backbone does the picture change, with MLEvolve ahead by mean estimate, though its medal rate confidence intervals still overlap \agentic's. A detailed paired-by-task comparison is in Appendix~\ref{app:detailed-mlebench}.

The left panel of Figure~\ref{fig:trajectory-average} shows all four methods following a
broadly similar quality-vs-wall-clock-time profile: quality rises quickly in the first few
hours, plateaus by around the 5--10 hour mark, and further gains afterwards arrive mostly as
occasional step jumps --- when a new best submission lands --- rather than as smooth continued
improvement.
AiScientist generally lags behind, except in the final $\approx$4 hours, where it quickly catches up, and MLEvolve is the slowest to plateau.

The right panel of Figure~\ref{fig:trajectory-average} shows the inference cost of each method.
Although total token volumes are comparable across agentic harnesses, cost is driven
almost entirely by cache-read tokens. At GLM 5.2, \agentic's modeled
USD cost (\$12.12) is $\approx$6.2$\times$ AiScientist's (\$1.95), even though its input and
output token volumes are comparable to Arbor's and AiScientist's --- only its cache-read
usage is roughly an order of magnitude higher. This follows directly from \agentic
running as a single ever-growing session that resends its whole prior context every turn
(the same-session cache model still prices that resent context, even though it credits
most of it as a cache hit); Arbor and AiScientist instead reset into fresh, shorter-lived
threads per branch or submission, so their modeled cache-read volume never grows as large.
Nonetheless, overall cost remains dominated by hardware.\footnote{For our hardware configuration, 24h of A100 80GB compute can total between \$24 and \$85, depending on the provider.}

\subsection{Trace Analysis}
\label{sec:trace}

\begin{wrapfigure}[16]{r}{0.45\linewidth}
\centering
\includegraphics[width=\linewidth]{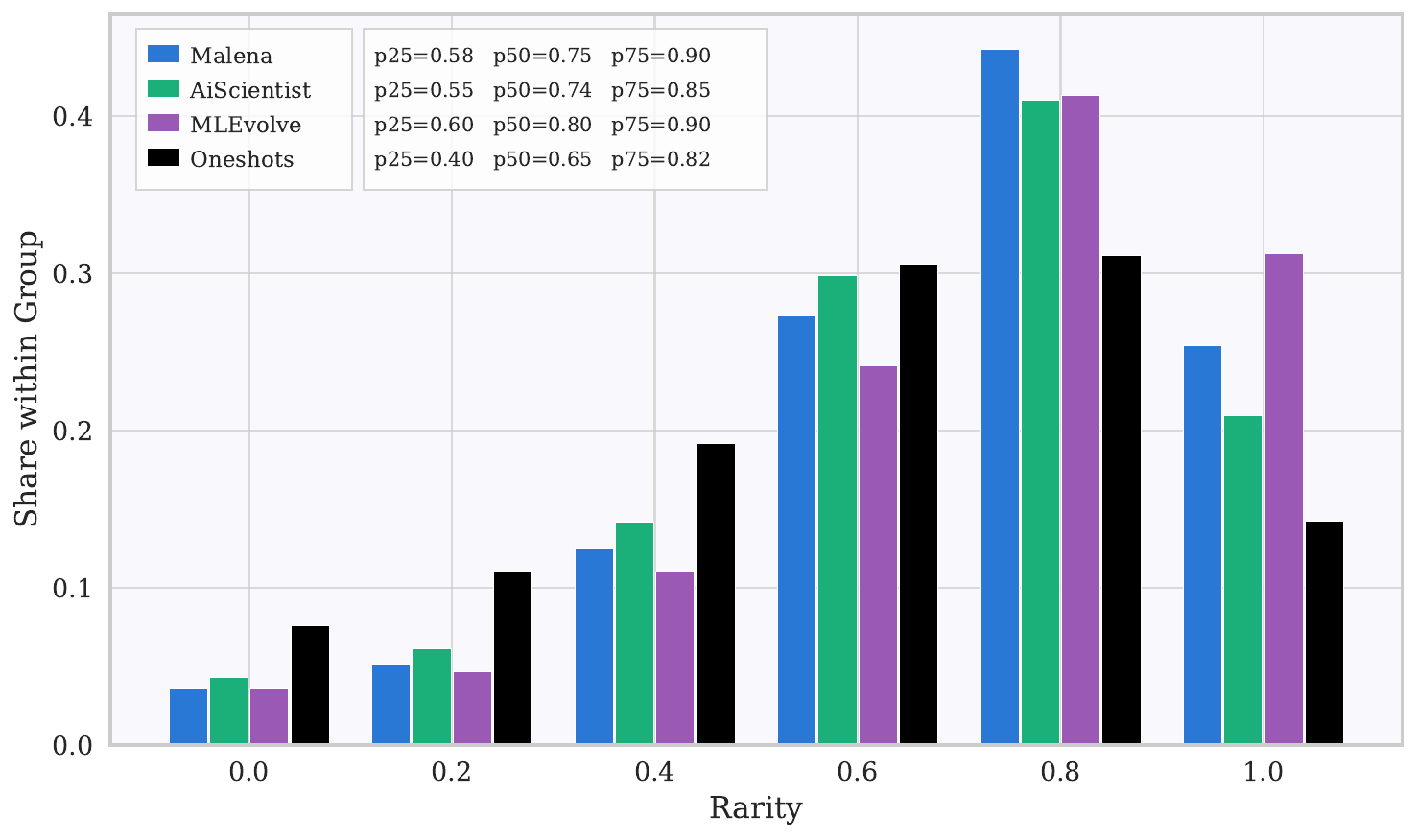}
\caption{\textbf{Per-technique rarity by method}. Rarity of a technique is computed once per task over every method/backbone group's runs combined (see Appendix~\ref{app:technique-rarity}).}
\label{fig:technique-rarity-heatmap}
\end{wrapfigure}
We now examine whether \agentic exhibits the failure mode that motivates prior designs, and how severely. The failure mode these designs are most explicitly built around is an inability to reliably \textit{search} the space of ML solutions, so we focus our trace analysis on search, studying in depth the traces of Kimi K3 and GLM 5.2 under MLEvolve, AiScientist, \agentic and \oneshot iterations.

Prior designs typically impose an explicit search structure --- a tree over candidate ideas (Arbor, MLEvolve), a population of competing programs, or a stagnation-triggered switch to a fresh idea (ScienceFlow) --- built around a discrete unit of work or checkpoint that is either refined or abandoned. \agentic's workflow with GLM 5.2 and Kimi K3 differs qualitatively: it produces a single general pipeline spanning a task's natural stages (e.g., the two-stage embed-then-retrieve pipeline on \texttt{whale}; see Appendix~\ref{app:whale-example}), then spends the rest of the run on what amounts to hyperparameter search, but over code patches rather than parameter values. This lets it reuse checkpoints and utility code (data loading, preprocessing, validation splits) across iterations, improving efficiency and narrowing the generalization gap.

\begin{figure}[t]
\centering
\includegraphics[width=\linewidth]{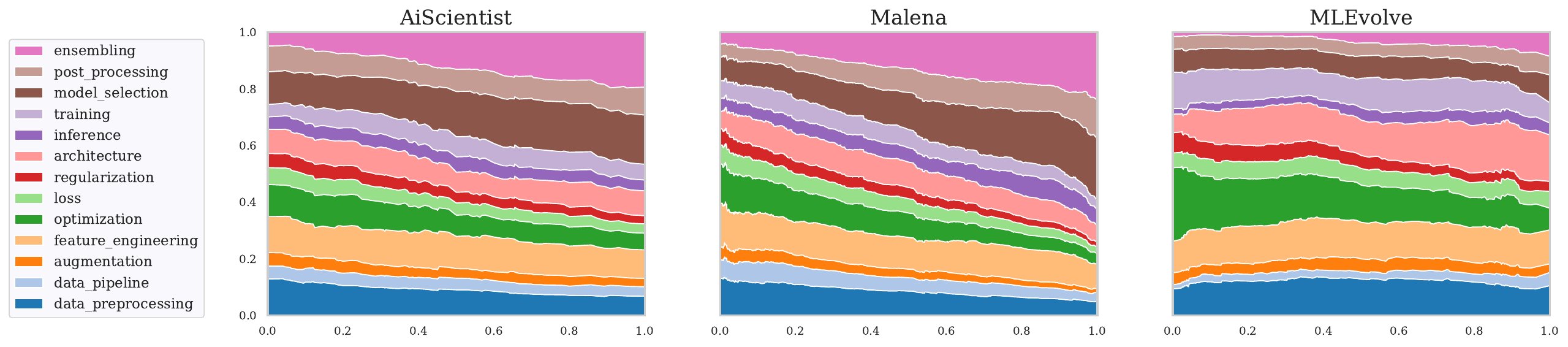}
\caption{\textbf{Category composition of techniques used so far, over the course of a run}, one column per method. The x-axis is checkpoint fraction through the run ($\in [0, 1]$); the y-axis is the decayed share of technique-category tags at that point (stacked, summing to 1). Each technique contributes to a checkpoint's composition with exponential decay by recency, giving a balanced view of the solution's techniques that still favours ones added recently. Per-task and alternate half-life versions are in Appendix~\ref{app:category-share-decay-robustness}.}
\label{fig:technique-category-share-decay}
\end{figure}

Going further, we annotate each trajectory's checkpoints with the ML techniques used (see Appendix~\ref{app:checkpoint-methodology} for the full labeling methodology) to see how each method searches and what it searches over. Two findings stand out.
First, methods with LLM-driven search (\agentic and AiScientist) start out balanced across the stages of a standard ML pipeline --- data preprocessing, feature engineering, architecture, training, and beyond --- with AiScientist slightly skewed towards model selection early on. Both become more performance-oriented towards the end of the run, concentrating on ensembling and, at times, techniques borrowed from the Kaggle community such as pseudo-labeling (Appendix~\ref{app:pseudo-label-example} gives a worked example of \agentic independently rediscovering this technique under two different backbones). This late-run shift is more pronounced for \agentic than for AiScientist, with the shares of ensembling, model selection, and post-processing all increasing until the very end of the run. MLEvolve, in contrast, settles into a roughly stable technique distribution after about the first 10\% of the run (Figure~\ref{fig:technique-category-share-decay}).
Second, this skew towards performance does not confine \agentic to well-known recipes: Figure~\ref{fig:technique-rarity-heatmap} shows that \agentic and MLEvolve are consistently the top two methods in technique rarity (see Appendix~\ref{app:technique-rarity} for how rarity is computed), i.e., the techniques they use tend to appear in fewer of the other groups' runs. \agentic therefore keeps proposing comparatively rare techniques while becoming more performance-oriented late in the run.

Together, these results suggest that a sufficiently capable coding-agent backbone can drive
this kind of search largely on its own: \agentic balances exploring genuinely rare techniques
against a growing performance focus as the run progresses, all while efficiently reusing the
checkpoints, utility code, and other artifacts it has already produced rather than starting
over, with no hand-built harness needed to elicit this behaviour.

\section{Conclusion}
\label{sec:conclusion}

Systems for autonomous machine learning engineering have grown steadily more elaborate:
search trees over candidate solutions, populations of competing programs, teams of
specialized agents. That machinery was designed when the underlying unit of work was a
single model completion, which cannot run the code it writes, remember what it already
tried, or react to an error. Coding-focused post-training has since changed the primitive
itself: a model highly capable of autonomous work plans, executes and debugs inside one
long-running session, without an outer program telling it what to do next. This paper asks
which of the accumulated machinery still earns its keep once the primitive can do these
things itself, and answers it by ablating that machinery layer by layer at a fixed model,
budget and task set.

Almost all of the gain comes from the runtime and LLM backbone, and almost none from what is built on top of
them. Giving the model a shell and a filesystem instead of a chat interface is the single
largest effect we measure. Once that is in place, and as the coding agent is given more autonomy, the techniques accumulated by the test-time-compute literature over the last few years barely move the result.
A model highly capable of autonomous work already does much of what these techniques were designed to supply from
outside, so supplying it again buys little. A single long-running agent session with minimal
machinery on top turns out to be a strong baseline, and none of the standard components we
added improved on it.

We then confirm the same picture from the outside, against state-of-the-art systems built specifically for
this task. Holding the model, hardware, and budget fixed,
none of the published harnesses we tested significantly outperforms the plain agent session
with frontier backbones.
With GLM 5.2, \malena earns a medal on 62.5\% of the competitions, compared with 47.1\% for the best external harness tested.
Across all backbones, our statistical analysis either favours \malena over the other harnesses or finds no significant medal-rate difference between them.

The practical reading is that effort spent elaborating harnesses around a model that is
already strong at working autonomously has poor returns. The leverage is in the model and in
the runtime it is given, not in the scaffolding built around them.

\paragraph{Limitations} Most of our results rely on MLE-bench, which has known task-preparation
issues and, being built from public Kaggle competitions, is a plausible contamination target;
we mitigate this by fixing broken tasks, directly checking for contamination, and additionally
evaluating on NatureBench. A second limitation is statistical power: some results still have
confidence intervals wide enough to preclude strong claims about effect size, though we believe
the weaker \emph{no-worse-than} claims are well supported.
A third is scope: most harness interventions we test operate within a single worker, and we leave a more in-depth exploration of sophisticated inter-agent coordination protocols to future work.
A fourth is possible asymmetry in tuning effort between \agentic and the external harnesses it is compared against; Appendix~\ref{app:baseline-tuning-effort} documents the backbone-specific bug fixes and configuration ablations we applied to each external harness to mitigate this.
Full discussion in Appendix~\ref{app:discussion}.

\subsection*{Reproducibility statement}

Full experimental settings are described in Section~\ref{sec:settings} and in
Appendix~\ref{app:implementation}: the coding agent framework and tool set exposed to the
model in each iteration (Appendix~\ref{app:infra-details}), the search strategies compared
(Appendix~\ref{app:search-details}), and the exact LLM backbones and hosting configuration
used, with their HuggingFace identifiers (Appendix~\ref{app:models}). The task sets used for
both benchmarks are listed in full in Appendix~\ref{app:data} (MLE-bench, including which
tasks we fixed to remove data or ground-truth-leakage issues, and why) and
Appendix~\ref{app:naturebench} (NatureBench); the exact code diffs we applied to fix the
affected MLE-bench preparation scripts are given in Appendix~\ref{app:task-fixes}. Our
statistical methodology --- how macro-task averages, bootstrap confidence intervals, and
paired-by-task comparisons are computed --- is described in Appendix~\ref{app:statistics}.

\subsection*{Acknowledgments}

We thank Simin Fan for help during the earlier stages of this project, and
Samy Bengio for useful feedback during its later stages.

\bibliography{iclr2027_conference}
\bibliographystyle{iclr2027_conference}

\newpage

\appendix

\input{appendices.tex}

\applefootnote{\textcolor{textgray}{\sffamily Apple and the Apple logo are trademarks of Apple Inc., registered in the U.S. and other countries and regions.}}

\end{document}

%% file: math_commands.tex
\usepackage{amsmath,amsfonts,bm}

\def\eqref#1{equation~\ref{#1}}

\def\1{\bm{1}}

\DeclareMathAlphabet{\mathsfit}{\encodingdefault}{\sfdefault}{m}{sl}
\SetMathAlphabet{\mathsfit}{bold}{\encodingdefault}{\sfdefault}{bx}{n}

%% file: tables/table_agentic.tex
\begin{tabular}{lcccc}
\toprule
 & \multicolumn{2}{c}{Percentile ($\uparrow$)} & \multicolumn{2}{c}{Medal rate (\%, $\uparrow$)} \\
\cmidrule(lr){2-3} \cmidrule(lr){4-5}
 & Self-select & Oracle & Self-select & Oracle \\
\midrule
\texttt{base} & 66.51 [62.46, 69.97] & 69.29 [65.82, 72.13] & \textbf{55.7} [48.2, 62.9] & 60.4 [53.2, 66.8] \\
\texttt{+D} & 63.62 [59.27, 68.26] & 67.40 [63.22, 71.98] & 45.2 [38.1, 54.8] & 47.0 [38.1, 56.0] \\
\texttt{+P3} & \textbf{67.77} [64.58, 70.62] & \textbf{71.79} [69.34, 73.83] & 52.4 [42.8, 61.9] & \textbf{60.7} [53.6, 67.3] \\
\texttt{+B+P3} & 58.91 [53.50, 64.05] & 68.57 [65.77, 71.14] & 33.3 [23.8, 40.5] & 53.0 [45.8, 60.1] \\
\bottomrule
\end{tabular}

%% file: tables/table_naturebench_from_csv.tex
\begin{tabular}{lccc}
\toprule
& \multicolumn{3}{c}{Surpassed SOTA (\%) $\uparrow$} \\
Backbone & Malena & AiScientist & MLEvolve \\
\midrule
GLM-5.2 & \shortstack{21.7\\{\scriptsize [15.0, 27.5]}} & \shortstack{17.5\\{\scriptsize [12.5, 22.5]}} & \shortstack{10.8\\{\scriptsize [7.5, 15.0]}} \\
Kimi-K3 & \shortstack{26.7\\{\scriptsize [20.0, 32.5]}} & \shortstack{27.1\\{\scriptsize [20.0, 35.0]}} & \shortstack{11.7\\{\scriptsize [7.5, 15.0]}} \\
\bottomrule
\end{tabular}

%% file: appendices.tex
\section{Experimental setting details}

\subsection{MLE-bench tasks used}
\label{app:data}
Our primary dataset used in this paper is MLE-bench~\citep{chan2025mlebenchevaluatingmachinelearning}, which consists of 75 tasks.
In this paper we focus on a subset of 30 tasks.
We largely follow~\citet{hambardzumyan2026aira2overcomingbottlenecksai}, which used the 30 tasks reported in the GPT-5 system card~\citep{singh2026openaigpt5card} for their experiments.
We, however, substitute the task \texttt{bms-molecular-translation} in favor of \texttt{alaska2-image-steganalysis}, due to its more permissive license.
Our final split consists of 5 Lite, 21 Medium and 4 High tasks.
All tasks used in the paper are shown in Table~\ref{tab:tasks}, along with the split they belong to and the shorthand name by which they may be referenced in this paper.

Results presented in Section~\ref{sec:systematic} are reported on the original MLE-bench without any intervention, and in the 29 task split (without either \texttt{bms} or \texttt{alaska2}).
This split is referred to in this work as the \tasksplit{29} task split.

We note that a subset of our tasks has been previously reported containing issues in the MLE-bench preparation scripts.
Such issues may include missing data needed to correctly train ML pipelines, data present that leaks the ground truth predictions, etc.
We have identified four competitions that could either lead to impossible tasks or are susceptible to cheating.
In order to maintain a fair comparison set between our baselines and other tested public harnesses, we fixed such issues and report their performance for all results in this paper when comparing against existing baselines.
Such tasks are marked with \textsuperscript{\textdaggerdbl} in Table~\ref{tab:tasks} and the split is labeled as the \tasksplit{30} set in this paper.
Appendix~\ref{app:task-fixes} lists the exact code diffs applied to the MLE-bench preparation scripts, together with the underlying bugs they address.

The selection criterion for our 14-task and 10-task challenging splits presented in Section~\ref{sec:layer5} and Appendix~\ref{app:context-management} was based on removing the floor -- tasks solved by Gemma4 31B -- and removing ceiling -- tasks not solved by Kimi K3.
These selections are shown in Table~\ref{tab:tasks} as tasks with * marks and \textsuperscript{\textdagger} marks, for \tasksplit{14} and \tasksplit{10}, respectively.
As their name suggests, this subset is running with the fixed task version, when applicable.

\begin{table}[h]
\centering
\caption{\textbf{Full MLE-bench task suite used in the paper.}
Competitions marked with \textsuperscript{\textbardbl} are not part of the \protect\tasksplit{29} split.
Tasks with * indicate the \protect\tasksplit{14} task subset, and \textsuperscript{\textdagger} indicates \protect\tasksplit{10}.
Tasks with \textsuperscript{\textdaggerdbl} were identified to have breaking issues, and results aggregated on them in \texttt{fixed} subsets come from our fixed version of such tasks.}
\label{tab:tasks}
\input{tables/mlebench-tasks.tex}
\end{table}

\subsection{Fixes applied to MLE-bench tasks}
\label{app:task-fixes}

Below we list the fixes for \texttt{champs}, \texttt{hubmap}, \texttt{smartphone}, and \texttt{multi-modal}: the bug in each, and the change we made to its MLE-bench \texttt{prepare.py}.
All four issues are documented by the MLE-bench maintainers on GitHub, in the ``Known Issues'' section of the official repository's README (\url{https://github.com/openai/mle-bench#known-issues}): \texttt{champs} in pull request \#70 (\url{https://github.com/openai/mle-bench/pull/70}), \texttt{multi-modal} in issue \#77 (\url{https://github.com/openai/mle-bench/issues/77}), \texttt{smartphone} in issue \#93 (\url{https://github.com/openai/mle-bench/issues/93}), and \texttt{hubmap} directly in the README.
The maintainers deferred fixing them to avoid invalidating existing leaderboard submissions, so the upstream preparation scripts still contain these bugs; the diffs below are our own fixes.
The diffs are shortened for space (\texttt{...} marks omitted context, comments are trimmed, and long lines are re-wrapped).
In all four cases the train/test splits and private answers are identical to the original MLE-bench preparation; only the files given to the agent change.

\input{task-fixes/champs.tex}

\input{task-fixes/hubmap.tex}

\input{task-fixes/smartphone.tex}

\input{task-fixes/multi-modal.tex}

\subsection{NatureBench tasks used}
\label{app:naturebench}

NatureBench~\citep{wang2026naturebenchcodingagentsmatch} asks an agent to try to match or
exceed the \emph{published} state-of-the-art result of a recent Nature-family paper on
that paper's own held-out evaluation protocol. Unlike \tasksplit{30}'s Kaggle
competitions, there is no external leaderboard to compute a percentile against: each task
ships a dedicated evaluation service that the agent may query online, unlimited times,
via an \texttt{/evaluate} endpoint that returns the candidate's score. We follow
NatureBench's own protocol and report a task as surpassed when its normalized gain score
$g > 0.1$. We report the beat-SOTA rate --- the
fraction of tasks on which the best score achieved anywhere in the run (oracle) surpasses
the published SOTA --- macro-averaged over a licensed task cohort at an 8h time budget, for both a GLM 5.2 and a Kimi K3 backbone.

Table~\ref{tab:nature-tasks} lists the 40 NatureBench tasks used for the results in Section~\ref{sec:results}.

\begin{table}[h]
\centering
\caption{\textbf{NatureBench task cohort used in the paper.} Task IDs are the paper's
Nature-family DOI suffix.}
\label{tab:nature-tasks}
\input{tables/table_nature_tasks.tex}
\end{table}

\subsection{LLM backbones used}
\label{app:models}

Unless otherwise stated, we report results using GLM 5.2 backbone LLM.
All backbones are served locally with vLLM~\citep{kwon2023efficient} on 8$\times$B200 or 8$\times$A100 80GB nodes for smaller models.
Unless otherwise specified, all models run with reasoning enabled at their maximum available configuration, and at their canonical maximum sequence length.
Input and output costs, when reported, are the weighted effective per-token prices listed
for each backbone on \url{https://openrouter.ai/} --- a marketplace that aggregates
multiple inference providers per model and reports a usage-weighted effective price across
them, rather than any single provider's list price --- snapshotted around September 2026.
Cache-read cost is modeled as $0.1\times$ the input cost: none of the harnesses compared
here report real cache-hit token counts directly (see the per-method telemetry notes in
Section~\ref{sec:baselines-mlebench} and Appendix~\ref{app:kimi-trajectory-tokens}), so this
is a fixed discount applied uniformly rather than a per-backbone reported cache price.
The following are all the LLMs reported in this paper, along with their HuggingFace
identifier; backbones we compute a dollar cost for elsewhere in the paper are annotated
with their September-2026 OpenRouter price:

\begin{itemize}
	\item GPT OSS 120B: \texttt{openai/gpt-oss-120b}~\citep{openai2025gptoss120bgptoss20bmodel}.
	\item Agents A1: \texttt{InternScience/Agents-A1}~\citep{bai2026scalinghorizonparametersreaching}.
	\item Gemma 4 31B: \texttt{google/gemma-4-31B-it}~\citep{gemmateam2026gemma4} (\$0.1641 / M input, \$0.5146 / M output).
	\item DeepSeek V4 Flash (Preview): \texttt{deepseek-ai/DeepSeek-V4-Flash-0731}~\citep{deepseekai2026deepseekv4}.
	\item DeepSeek V4 Flash: \texttt{deepseek-ai/DeepSeek-V4-Flash}~\citep{deepseekai2026deepseekv4}.
	\item DeepSeek V4 Pro (Preview): \texttt{nvidia/DeepSeek-V4-Pro-NVFP4}~\citep{deepseekai2026deepseekv4}.
	\item DeepSeek V4 Pro: \texttt{nvidia/DeepSeek-V4-Pro-0813-NVFP4}~\citep{deepseekai2026deepseekv4}.
	\item GLM 5.2: \texttt{nvidia/GLM-5.2-NVFP4}~\citep{glm5team2026glm5vibecodingagentic} (\$0.3477 / M input, \$3.123 / M output).
	\item Kimi K3: \texttt{moonshotai/Kimi-K3}~\citep{kimiteam2026kimik3openfrontier} (\$0.5714 / M input, \$13.99 / M output).
\end{itemize}

\subsection{Statistical protocol}
\label{app:statistics}

Throughout the paper, our main performance reported is based on the macro-task average.
For an experiment with $T$ different tasks, each task $t$ containing $N_t$ seeds, for metric $\mathrm{metric}$, we report:
\[\frac1T \sum_{t=1}^T \frac{1}{N_t} \sum_{r=1}^{N_t} \mathrm{metric}(\mathrm{run}_{t,r}).\]
Our confidence intervals are estimated via bootstrap on the fixed set of $T$ tasks, sampling for each task from the $N_t$ available seeds before computing the macro average, and taking the quantiles of the simulated distribution.
For each plot and table presented, the comparison is always based on the same task set.
In the tables presented in this paper, we mark as bold numbers the best results obtained across arms, and when present, with underlined and italics the second and third best, respectively.

For our $A$ vs $B$ comparisons, we compute the paired-by-task difference to isolate positive or negative effects on the ablation:
\[\frac1T \sum_{t=1}^T \left(\frac{1}{N(A)_t} \sum_{r=1}^{N(A)_t} \mathrm{metric}(\mathrm{run}(A)_{t,r}) - \frac{1}{N(B)_t} \sum_{r=1}^{N(B)_t} \mathrm{metric}(\mathrm{run}(B)_{t,r})\right).\]
To determine statistical significance, we compute the 95\% CIs of the paired-by-task difference (again, on the fixed set of $T$ tasks, bootstrapping from $N(A)_t$ and $N(B)_t$).
If $0$ does not lie in the CI --- which we mark with * in our paired-by-task tables throughout the paper --- we can establish a statistically significant difference in the metric in the ablation.
When $0$ lies in the CI, the arms become statistically indistinguishable at our power level based on the amount of seeds we ran in our experiments.
With this method, between-task difficulty cancels, which results in narrower intervals than regular marginal CIs.

\section{Implementation details}
\label{app:implementation}

In this section, we provide more details about our core iterations ablated, along with their optional tool and infrastructure addons below.

\subsection{Iterations}
\label{app:iterations}

All three iterations are compiled from a single prompt-template family.
All iterations share a base prompt with the agent persona, the scientific-approach questions, the resource and time budget line and a few general recommendations.
A specialized task description and instruction block is then chosen for each iteration and aligns with the goal and environment available for each method.

In the task description, the \oneshot and \agentic iterations are handed only the data paths (raw data directory, description file and sample submission file) since they can explore the data themselves through their tools.
On the other hand, \chat receives the task description verbatim and an inline overview of the data computed by the harness: a directory tree of the data folder, per-file previews (per-column statistics for CSVs --- dtype, cardinality/value counts, mean, range and NaN counts --- auto-generated schemas for JSON files, and the first lines of text files, subject to a global character budget), and the first rows of the sample submission.
All iterations execute inside a git worktree checked out at their parent node's commit, and each produces a node whose recorded validation score tuple is what the search rules of Section~\ref{sec:layer2} (Appendix~\ref{app:search-details}) consume.

\begin{enumerate}
	\item \textbf{\chat}. A single chat-completion call with no tool access.
	The model receives the base prompt with the inline data overview and is instructed to reply with a complete, self-contained Python script that implements the full pipeline (data $\to$ model $\to$ training $\to$ inference) and writes \texttt{submission.csv}.
	The response text is saved verbatim to \texttt{main.py} after stripping any enclosing code fences, and executed once with the remaining time budget.

	If the run succeeds and the submission validates against the expected format, the agent is queried again with the output of the command to extract the validation score, and the iteration ends.
	Otherwise the harness re-prompts the model --- appending to the same conversation --- with a failure report containing the failure reason (non-zero exit, missing or invalid submission), the truncated stdout/stderr, a hardware-utilization summary, and a footer that disambiguates system kills (GPU-memory watchdog, RAM watchdog) from ordinary crashes.
	The model must answer with the full corrected script.
	This loop repeats until a valid submission is produced.

	\item \textbf{\oneshot}. One coding-agent session, one conversation turn, under the default tool environment of Appendix~\ref{app:infra-details}.
	The prompt instructs the agent to treat the worktree as a scratch directory to confirm its design with cheap verification runs (fast data-inspection scripts, then minimal \emph{test runs}), and to converge to exactly one \emph{production run}.
	After one production run succeeds, no further heavy training is permitted, and failures may only be diagnosed with cheap runs, never with a second production attempt.

	Four deliverables are required: \texttt{main.py} (the pipeline), \texttt{submission.csv}, \texttt{design.md} (a self-contained write-up of the design, results and rationale, meant to serve as the node's documentation for later iterations), and \texttt{score.txt} (a single floating-point validation score on the task's metric and direction).
	These deliverables are expressed as \emph{guarantees}: when the agent ends its turn, the harness statically verifies the file set and re-prompts the agent with the list of missing or malformed guarantees (e.g.\ an unparsable score file) until all hold.

	When the parent node already carries a scripted solution (a previous \texttt{main.py} and \texttt{design.md}), the prompt switches to an \emph{improve} variant: the agent is instructed to study the documented design, to decide between refining and pivoting using the recorded execution cost and results, and to overwrite the same files (any other scripts are discarded).
	This is the mechanism by which the search strategies of Section~\ref{sec:layer2} share information between nodes.

	The implementation also supports appending optional instructions to the prompt as described in Appendix~\ref{app:infra-details}.
	This addition is only added when the corresponding intervention is enabled.

	\item \textbf{\agentic}. The same coding-agent environment as \oneshot, but the session is allowed to span the iteration's entire time budget in place of a single production run, and the one-production-run and documentation requirements are lifted.
	The instructions grant the agent freedom to design any number of pipelines and produce any number of submissions, dividing its budget as it sees fit.
	The harness drives the session as a turn loop: whenever the agent ends a turn before the budget expires, it is re-prompted with a short follow-up message that reports the remaining time.

	Submissions are generally tracked using the submission plugin.
	The agent calls \texttt{submissions\_register} itself (optionally consulting \texttt{submissions\_list}), reporting each submission together with its validation score and free-form metadata, with the prompt making explicit that only registered submissions count and that each registration is atomic and immediate.
	When the submission plugin is disabled, the agent simply saves submission files under a \texttt{submissions/} directory and the harness salvages and registers every file found there at wrap-up (without a validation score).
	Unless explicitly stated, all experiments reported use the submission plugin with validation score signals.
\end{enumerate}

\subsection{Infrastructure}
\label{app:infra-details}

The discussion in this section only applies to coding agent iterations (\oneshot and \agentic).

Our core coding agent loop harnesses over OpenCode.
It preserves all default tools, except we explicitly disabled \texttt{doom\_loop}, \texttt{plan\_exit}, and \texttt{question}.
Moreover, we overwrite the native \texttt{bash} tool with our own implementation (more information in the following overview).
The most important and used tools for the coding agent are:
\begin{itemize}
	\item \texttt{read}: Reads the contents of a file.
	\item \texttt{write}: Writes the contents of a file.
	\item \texttt{edit}: Edits the contents of a file.
	\item \texttt{bash}: Executes a command.
	Note that we disabled the native \texttt{bash} implementation, and used our own version.
	Both versions remain highly similar in usage, description and output format, except for the fact that we manually control the maximum command execution timeout (set to 24h in all experiments unless otherwise stated), general resource statistics are shown along with the output of the command upon completion (command duration and average and max CPU, GPU, RAM and VRAM usage by the command), along with a background watchdog service that automatically kills running commands shortly before the system runs out of memory, to avoid hard pod ejection by our compute provider.
	We only allow our agent to execute commands from a short whitelist, including \texttt{pwd}, \texttt{ls*}, \texttt{python*}, \texttt{nvidia-smi} and more.
	\item \texttt{task}: Allows the coding agent to launch multiple subagents with different prompts.
	The tool blocks until all subagents finish their work.
	\item \texttt{todowrite}: Allows the agent to maintain a task list of goals during its session.
\end{itemize}

On top of this, we additionally ablated the following tools.
The default \agentic configuration has \texttt{submission}, \texttt{system}, and \texttt{jobs} tools enabled.
The other coding agent iterations run with them disabled, unless otherwise specified for any particular ablation.
\begin{itemize}
	\item \texttt{submission\_register} tool: Adds a submission file to the registry.
	No test or leaderboard scores are communicated to the agent.
	Additionally, the agent may use \texttt{submission\_list} to list the previous submissions it has produced during the session.
	Whenever this tool is enabled, we also add a short text in the session's prompt to guide the agent regarding its usage.
	\item[\texttt{S}] \texttt{system} tool: Returns the current hardware resources, utilization, and time remaining.
	\item[\texttt{J}] \texttt{jobs} tools: Allow running bash commands in the background, waiting for them to complete and cancelling ongoing jobs.
	Whenever this tool is enabled, we also set a maximum time limit of 10m to all \texttt{bash} calls to encourage its usage, and add a short usage guide in the session's prompt about how to use it.
	This suite is composed of the following tools.
	\texttt{job\_submit}: non-blocking, executes a command in the background and returns its general information (job ID, stdout path, etc.), \texttt{job\_wait}: waits for a job to finish or, if specified, until a certain amount of time passes, \texttt{job\_queue}: lists the running jobs, and \texttt{job\_info}: a more detailed view of a particular job, including hardware usage, stdout location, and more.
	\item[\texttt{s}] \texttt{skill}: Loads a skill matching the input's identifier.
	The agent is given the name and a one-line description of each available skill up front, and only loads the full body of a skill (via this tool) when it decides to consult it.
	We equip the agent with eight skills encoding common ML engineering strategies for iterating on a competition:
	\textit{submission-blueprint} (build and verify the end-to-end submission pipeline first, using a trivial predictor, before training a real model);
	\textit{quick-strong-baseline} (get the highest score for the least training and implementation effort, leaning on pretrained models and off-the-shelf libraries);
	\textit{cost-saving-techniques} (a toolbox of levers --- frozen backbones, low resolution, few epochs, data subsampling --- for getting a faithful signal cheaply, meant to be folded into the other skills);
	\textit{simplification-probe} (shrink the task, either by holding one axis at full difficulty while collapsing the rest, or by reformulating it as a smaller proxy task, to isolate where a score gap comes from);
	\textit{ablation-probe} (design a controlled comparison that isolates one component's contribution, holding everything else fixed);
	\textit{runtime-estimate} (measure a few scaled-down configurations and extrapolate wall-clock and memory cost before committing to a full-scale run);
	\textit{stagnation-pivot} (detect when recent attempts within the same paradigm have stopped improving and force the next attempt to pivot the model, loss, input representation, or problem decomposition rather than tweak a hyperparameter); and
	\textit{cross-iteration-synthesis} (step back and read across several past attempts together to compose ideas, extract best practices, standardize the validation protocol, or resolve contradictions).
	\item[\texttt{H}]  \texttt{huggingface\_hub\_repo\_search}: The official HuggingFace MCP.
	Enables the agent to search for models in the HuggingFace Hub.
	Whenever this tool is enabled, we also add a short text in the session's prompt to guide the agent regarding its usage.
	\item[\texttt{B}] \texttt{broadcast}: Enabled jointly with parallelism.
	Allows agents to broadcast messages to their peers, which are then queued and displayed after a turn end.
\end{itemize}

Additionally, we also ablate the following set of wider interventions, that each may include prompt changes or other environment differences for the agent:
\begin{itemize}
	\item[\texttt{D}] Delegation encouragement.
	This intervention enables background subagents via the native \texttt{task} tool, and encourages their use via a prompt addition.
	\item[\texttt{N}] Time nudges: Scheduled at 25\%, 50\%, 75\% and 90\%, the system automatically queues a brief message to the session and informs the agent about elapsed time.
	\item[\texttt{R}] Resource nudge: Whenever the CPU or GPU utilization falls under 50\%, the system queues a brief message to the agent to encourage more efficient use of resources.
	There is a 5 minute waiting time between nudges to avoid overcrowding the session with messages.
	\item[\texttt{V}] Strict validation guidelines: We inject to the prompt a set of strict validation guidelines the agent must adhere to, with the hopes to minimize data leakage.
	\item[\texttt{I}] Minimal prompt: The initial prompt of the agent is simplified to contain only the crucial information, removing general guidelines to follow and examples.
\end{itemize}

In the entire paper, unless otherwise specified, \oneshot iteration is evaluated without any intervention, and \agentic makes use of the \texttt{submission}, \texttt{system} and \texttt{jobs} tools only.

\subsection{Search strategies}
\label{app:search-details}

The \Chain and \Greedy strategies are straightforward and are implemented as described in Section~\ref{sec:layer2}.
The \UCB strategy follows a deterministic selection rule, applying the UCB1 index~\citep{Auer2002-st} to the pool of existing nodes.
Unlike the tree-descent formulation of UCT~\citep{Kocsis2006-uj} adopted by prior MLE systems~\citep{du2026mlevolveselfevolvingframeworkautomated,chen2026marsmodularagentreflective}, which recursively selects among the children of the current node, we score every node in the tree and take a single global supremum.

In the following, assume the competition metric is higher-is-better (the lower-is-better case is analogous, replacing each normalized score by its complement).
Let $T$ be the current tree, $V(T)$ its nodes, and $\mathrm{s}(n)$ the validation score recorded on node $n$.
Then, $\hat{s}(n) = \max_{n' \in \mathrm{subtree}(n)} \mathrm{s}(n')$, so that a node's value reflects what its descendants eventually achieved.
This maximum backup replaces the running mean fitness used in canonical UCT implementations~\citep{toledo2025airesearchagentsmachine,du2026mlevolveselfevolvingframeworkautomated}, which is appropriate here because candidates are deterministic artifacts to be kept rather than stochastic returns to be averaged.
\UCB then computes
\[R(n) = \frac{\sigma(\hat{s}(n))}{|V(T)| - 1} + c \sqrt{\frac{\ln\left(|V(T)| + 1\right)}{|C(n)| + 1}},\]
where $\sigma(x)$ is the $0$-indexed rank of $x$ among all available scores, $C(n)$ are the children of $n$,  and $c > 0$ is the exploration-exploitation hyperparameter (fixed at $c = 0.75$ for all runs).

Ranking rather than using the raw metric is what makes a single $c$ transferable across competitions: MLE-bench metrics are arbitrarily scaled and of mixed direction, and a rank transform is invariant to any order-preserving reparameterization of the metric while remaining robust to outlying runs, in the manner of the rank-based fitness shaping used by evolution strategies~\citep{Hansen2001-rl,JMLR:v15:wierstra14a}.
Prior MLE harnesses instead min-max normalize the metric~\citep{chen2026marsmodularagentreflective}, substitute a bounded hand-designed reward~\citep{du2026mlevolveselfevolvingframeworkautomated}, or leave the metric raw and retune the exploration constant per setting~\citep{toledo2025airesearchagentsmachine}.
Keeping the first term in $[0,1]$ also lets us adopt the exploration constant that the UCT analysis prescribes for bounded rewards, $c \approx 1/\sqrt{2}$, without per-task tuning~\citep{Kocsis2006-uj}.

Our strategy for \Bootstrap methods consists of running independent single \oneshot iterations, and then composing a set of bootstrapped-trajectories $B_j =\{T_1^j, \ldots, T_N^j\}$ where each $T_i^j$ is a composition of multiple independent \oneshot iterations of the requested time limit budget.
Each trajectory is packed greedily so that the summed \emph{effective} time of its iterations does not exceed the target budget, and an iteration that would overflow the trajectory is discarded rather than carried into the next one.
The budget is therefore matched on the time the iterations actually consume, not on their nominal per-iteration time limit.
We can then extract all metrics for each trajectory $T_i^j$ by concatenating the submissions of all iterations in the trajectory.
Finally, we sample multiple independent bootstrapped-trajectory sets $B_1, \ldots, B_M$ and report the average-of-averages across this set.

\section{Additional \chat, \oneshot and \agentic experiments}

\subsection{\oneshot time limit effect}
\label{app:timeouts}

\begin{figure}[ht]
\centering
\includegraphics[width=\linewidth]{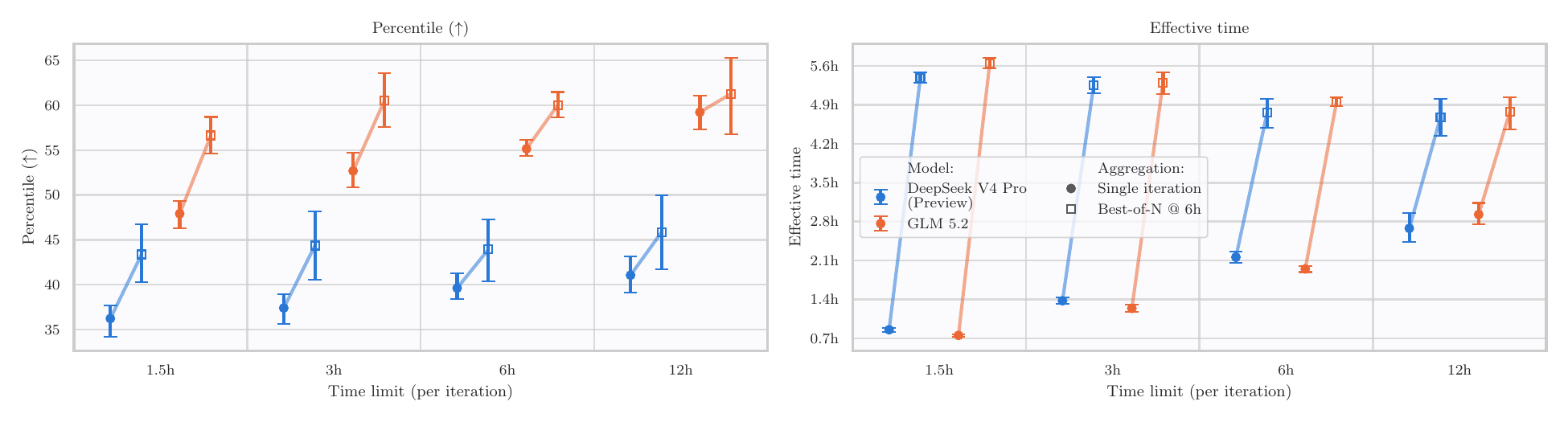}
\caption{\textbf{Impact of per-iteration time limits.}
Average percentile and effective-time obtained by \oneshot iterations under different per-iteration time budgets in the \protect\tasksplit{29} set.
Circled points represent single-iteration performance, while boxes indicate \Bootstrap performance at six hour budgets.
Error bars represent 95\% CIs across our fixed 29-task set.
The effective time measures the time taken for the run to finish.
\Bootstrap aggregates may remain below the target of 6h in the case of failing iterations that would exceed the maximum budget allocated.}
\label{fig:timeouts}
\end{figure}

We ablate the effect of the per-iteration time limit in our default \oneshot experimental setting for DeepSeek V4 Pro (Preview) and GLM 5.2 models.
Results are in Figure~\ref{fig:timeouts}.
We observe, as expected, a monotonic improvement of per-iteration performance as the time budget increases.
Nonetheless, the actual performance of the aggregated \Bootstrap composites quickly saturates at the three hour mark.
In our experimental setup, the six hour time limit default provides an reasonable choice to balance the performance and cost tradeoff.
Note that, even when prompted to use the highest possible time budget of 12h, the agents still have a much lower average effective time usage, which is what allows us to aggregate \Bootstrap at the 6h budget.

\subsection{Pairwise \oneshot search strategy and \malena agent comparison}
\label{app:pairwise-search}

\begin{table}[h]
\centering
\caption{\textbf{Pairwise paired-by-task comparison.}
Results reported on the 24h budget, using the \protect\tasksplit{29} task split.
In an A vs B comparison, positive percentile differences indicate better average performance of A.
Percentile difference reported along with 95\% CIs.}
\label{tab:pairwise-search}
\input{tables/pairwise-search.tex}
\end{table}

\begin{table}[h]
\centering
\caption{\textbf{Selection gap for different methods.}
Results reported on the 24h budget, with the \protect\tasksplit{29} split, along with 95\% CIs.}
\label{tab:selection-gap}
\input{tables/selection-gap.tex}
\end{table}

In Section~\ref{sec:layer2} and~\ref{sec:layer3}, we provided a head-to-head comparison in the paired-by-task regime between all search methods tested, and the \agentic agent.
In Table~\ref{tab:pairwise-search} we give a more detailed examination of such comparisons, and Table~\ref{tab:selection-gap} gives a more detailed view on the selection gap between these methods.

\subsection{Hidden validation gap}
\label{app:hce}

In Section~\ref{sec:systematic} we observed that the gap of the performance between self-select (based on validation score) and oracle-best (based on actual test scores) becomes a problem, particularly for strategies that share less information between their iterations.
In this section, we analyze three different interventions that can help mitigate such issue.
Our experiment setting is on the default \oneshot iteration configuration (six hour per-iteration time limit, GLM 5.2), and the baseline behavior control is the \Bootstrap aggregation at 12h budget setting.
We compare:
\begin{itemize}
	\item[\texttt{V}] Strict validation guidelines: We inject to the prompt a set of strict validation guidelines the agent must adhere to, with the hopes to minimize data leakage.
	\item[\texttt{BI}] Baseline iteration: At the beginning of the run, a Baseline iteration is spawned.
	The Baseline iteration aims to provide a simple, fast-to-run ML pipeline baseline to solve the task, while ensuring the validation split strategy follows a strict set of guidelines.
	Once this iteration ends, and until the 12h time limit, new \oneshot iterations use the Baseline's workspace as starting point, while still remaining independent from each other.
	The goal for this intervention is to provide a common validation strategy for the iterations, and minimize the inconsistencies that come from differently designed sets.
	\item[\texttt{HV}] Hidden validation: At the beginning of the iteration, the harness prepares a deterministic validation-train split from the training data, using MLE-bench's data preparation scripts.
	The agent does not have access to the target predictions of the hidden validation split.
	This strategy has the drawback of reducing the total amount of training data available to train the model in the first place.
	Note that, as described, this strategy cannot be deployed in practice, due to the hard requirement of the deterministic splitter which would require human intervention, and is therefore reported as a ceiling of what pre-splitting strategies can achieve.
\end{itemize}

\begin{table}[ht]
\centering
\caption{\textbf{Selection gap mitigation strategies.}
Performance reported with 95\% CIs. \protect\tasksplit{29} split at 12h budget.}
\label{tab:hce}
\input{tables/hce.tex}
\end{table}

Table~\ref{tab:hce} shows the results of this experiment.
We see that the base independent \Bootstrap \oneshot configuration provides the best percentile ceiling (oracle), while the introduction of the strict validation guidelines in the prompt alone allows for a slightly better self-select performance, at the cost of also lowering the oracle ceiling.
Finally, the introduction of the initial baseline iteration provides the smallest gap of all interventions ablated, at the cost of a significantly worse performance ceiling.
A plausible mechanism for these differences is the comparability of the validation scores themselves, rather than the number of submissions available to choose from: a single session produces all of its candidates under one validation split, so their reported scores are mutually comparable, whereas independent \Bootstrap iterations each design their own split and the harness is then asked to rank scores that were never on a common scale.
The \texttt{BI} intervention is consistent with this reading, as it is the one that supplies a shared validation workspace.
We report this as a hypothesis and not as a result: it predicts that the gap should grow with the number of distinct validation methodologies in the pool rather than with the number of submissions, which we do not test here.
While these results provide a quick overview about possible validation gap mitigation strategies, we acknowledge that at our current statistical power level, we cannot make strong conclusions about these results, and we encourage a more in-depth future work in this direction, including on the mechanism above.

\subsection{\agentic budget scaling}
\label{app:malena-scaling}

We study how \agentic scales under different time budgets.
Our experimental setting consists of a DeepSeek V4 Pro (Preview) backbone, with a \textit{bare} \agentic iteration, meaning with the \texttt{submission}, \texttt{system}, and \texttt{jobs} tools disabled.
This serves as the minimal harness over a coding agent.
In our implementation, we use the \texttt{submission} tool to register the validation scores of each submission, and without it, we have no self-select mechanism.
For this reason, in this section we report only the oracle ceiling score which, despite not being a production-ready configuration, still gives a realistic range of the temporal behavior of the agent, especially since the \agentic iteration shows the lowest validation gap in our tested iterations (see Appendix~\ref{app:pairwise-search}).

\begin{figure}[ht]
\centering
\includegraphics[width=0.75\linewidth]{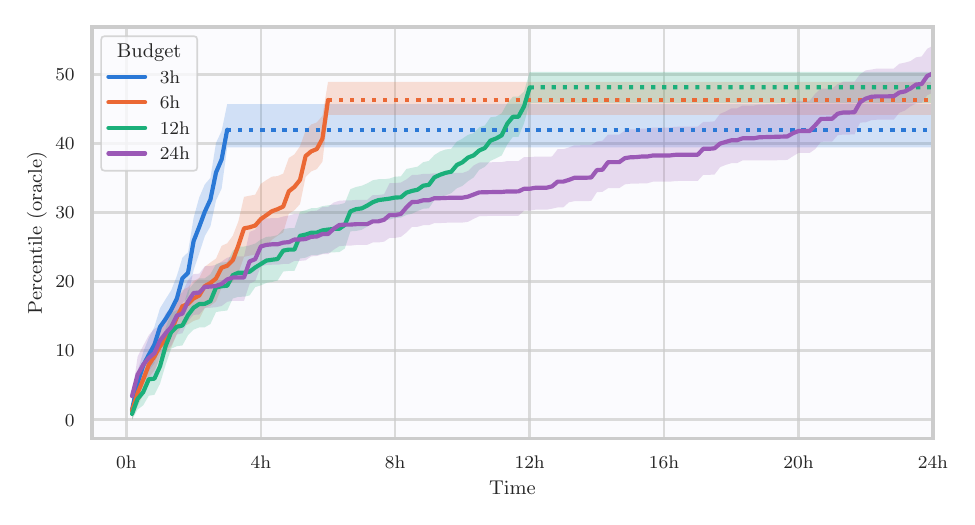}
\caption{\textbf{Behaviour of \agentic under different time budgets.}
Average oracle percentile obtained by \agentic on the \protect\tasksplit{29} split with a DeepSeek V4 Pro (Preview) backbone model.
Trajectories are extracted from the full run by scoring the submissions available at each time.
Dotted lines are shown after a timeout target is hit, and maintain the performance achieved.}
\label{fig:malena-budget}
\end{figure}

Figure~\ref{fig:malena-budget} shows the results of \agentic under different time limits.
We observe that the performance steadily increases as more budget is available.
Interestingly, despite having an almost identical prompt where the only difference is the time limit mentioned itself, and the automatic time left reminders mentioned at the end of the turn,\footnote{Not to be confused with the Time nudge (\texttt{N}) intervention --- which is also disabled in this experiment --- which schedules automatic messages periodically regardless of the progress in the session turn.} the performance of the agent at identical timeframes varies significantly depending on the total time budget allocated.
This suggests that the agent does adapt its exploration vs exploitation strategy depending on the time budget.

\subsection{\agentic with different backbones}
\label{app:malena-models}

\begin{figure}[ht]
\centering
\includegraphics[width=\linewidth]{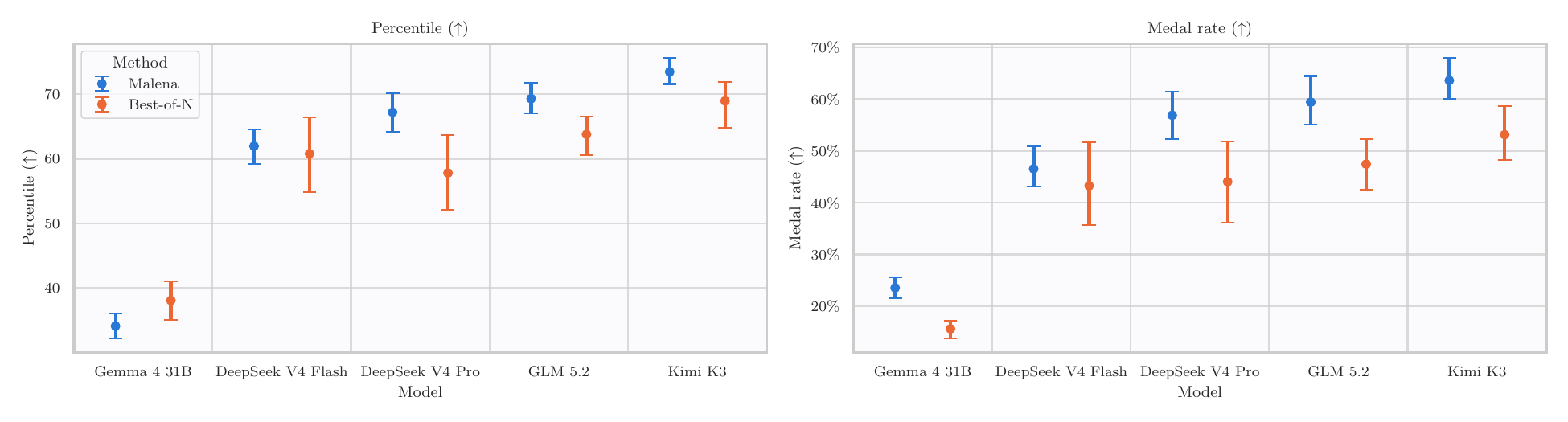}
\caption{\textbf{\agentic model ablation.}
Average percentile and medal rate obtained by the \agentic iteration and \Bootstrap \oneshot iterations at the 24h time budget in the \protect\tasksplit{29} set.
Error bars represent 95\% CIs across our fixed 29-task set.}
\label{fig:malena-models}
\end{figure}

We test the performance of the \agentic iteration with different backbone models, and compare it with the \Bootstrap \oneshot composite at the 24h time budget target.
Our selection of backbones is obtained from the results observed in Section~\ref{sec:layer1}, and we selected two frontier LLMs (GLM 5.2 and Kimi K3), both final releases of the DeepSeek V4 family, and Gemma 4 31B --- a smaller model that still shows clear benefits from the agentic environment.
We plot our results in Figure~\ref{fig:malena-models}, and detailed paired-by-task comparison in Table~\ref{tab:malenamodels}.
While the frontier models show a clear benefit from the long single-session iteration, the percentile of the weaker model favors \Bootstrap in comparison --- although the medal rate remains consistently better in \agentic iterations.
This may suggest that larger models can better leverage the autonomy in the single-session, or that the smaller context window of 256k in the weaker model becomes a bottleneck at this time scale, as some important details may be lost in the summarization during the context window compaction that is triggered when the context limit is reached.
We discuss this possibility further in Appendix~\ref{app:context-management}.

\begin{table}[h]
\centering
\caption{\textbf{Pairwise paired-by-task comparison of different backbones.}
Reported is the performance difference between \agentic (A) vs \Bootstrap (B); higher numbers indicate stronger average \agentic performance.
Results reported using the \protect\tasksplit{29} task split, with 95\% CIs.}
\label{tab:malenamodels}
\input{tables/malenamodels.tex}
\end{table}

\subsection{\agentic infrastructure ablation}
\label{app:malena-infra}

In this section, we ablate further interventions of the \agentic iteration itself.
We first ablate the importance of all interventions of the \agentic harness presented in Section~\ref{sec:layer3}, and we introduce a fourth intervention:
\begin{itemize}
	\item[\texttt{N}] Time nudges: Scheduled at 25\%, 50\%, 75\% and 90\%, the system automatically queues a brief message to the session and informs the agent about elapsed time.
	Note that for our default budget of 24h, the first time nudge is queued at 6h, which corresponds to the hard time limit of each \oneshot iteration.
\end{itemize}
Results are shown in Table~\ref{tab:malena-infra}, top row set.
We observe that for such long 24h sessions, the performance of the agent is negatively impacted when removing the \texttt{system} tool, suggesting that the agent struggles to track time correctly without it.
On the other hand, the addition of the time nudges results in a performance loss, perhaps due to a small distraction effect of unexpected messages. %
Finally, the removal of the \texttt{jobs} results in marginally better performance.

Furthermore, we incorporate on top of the time nudges (\texttt{N}) intervention a wider set of general-purpose infrastructure enhancements, with the objective of better understanding \textit{where} could the \agentic iteration be aided by the external  harness.
Namely, we test: a resource nudge that reminds the agent when hardware utilization is low (\texttt{R}), the HuggingFace MCP that allows the agent to look for pretrained models (\texttt{H}), a set of strict validation guidelines in the prompt (\texttt{V}) and a set of eight skills covering common ML engineering strategies for iterating on a competition, from establishing baselines to diagnosing stagnation (\texttt{s}).
More information about these interventions in Appendix~\ref{app:infra-details}.
In Table~\ref{tab:malena-infra}, bottom row set, we show the results of this experiment.
None of the interventions provide significant performance improvements at our current statistical power.
We conclude that \agentic as presented here serves as a capable enough baseline and further interventions either hurt performance, or do not produce significant improvements.
This is in contrast to \oneshot iterations, which may benefit from more complex harnesses as explored in Appendix~\ref{app:oneshot-infra}.

Finally, Table~\ref{tab:malena-infra-pairwise} shows the pairwise comparison of all our ablations, including the ones presented in Section~\ref{sec:layer5}.

\begin{table}[h]
\centering
\caption{\textbf{\agentic infrastructure ablation.}
Performance reported with 95\% CIs. Top set: \protect\tasksplit{29} split at 24h. Bottom set: \protect\tasksplit{14} split at 12h.}
\label{tab:malena-infra}
\input{tables/malena-infra.tex}
\end{table}

\begin{table}[h]
\centering
\caption{\textbf{\agentic pairwise paired-by-task comparison.}
In an A vs B comparison, positive percentile differences indicate better average performance of A.
Differences reported along with 95\% CIs.
Top set: 24h, \protect\tasksplit{29} split.
Middle set: 24h, \protect\tasksplit{14} split.
Bottom set: 12h, \protect\tasksplit{14} split.}
\label{tab:malena-infra-pairwise}
\input{tables/table_agentic_pairwise.tex}
\end{table}

\subsection{\agentic under a constrained context budget}
\label{app:context-management}

We additionally test how \agentic performs under an extremely short context budget with frequent summarization, on the full \tasksplit{10} task subset, comparing against the standing, unconstrained \agentic baseline on the same tasks.
We compare the behavior on GLM 5.2 running at its native 1M context limit, with an artificially constrained version that at whenever the session hits the 64k context limit, it triggers OpenCode's native session summarization compaction routine to reduce the context window.

\begin{table}[h]
\centering
\caption{Avg.\ compactions before first medal.}
\label{tab:compactions-before-medal}
\scriptsize
\begin{tabular}{lcccccccccc}
\toprule
Task & alaska2 & cassava & champs & freesound & hubmap & imet & kuzushiji & multi-modal & nfl & petfinder \\
\midrule
Avg & -- & 1.2 & -- & 1.5 & -- & 2.0 & 4.2 & 12.5 & 2.5 & 1.7 \\
\bottomrule
\end{tabular}
\end{table}

\textbf{Compaction fires reliably under the shrunk budget.} Every shrunk-context run triggers OpenCode's built-in compaction at least once, at an average rate of 9.5 compactions over the 24h budget (it ranges from 3 to 63 compactions; over the $n=40$ runs we observe 381 compaction events total). The context size right before compaction averages 65.4k tokens, tightly hugging the 64k threshold; the reset immediately after averages 15.6k tokens, i.e.\ each compaction discards roughly 76\% of the accumulated context. Checking submission timing against compaction timing shows the run does not simply ``beat'' context pressure by locking in its medal before compaction kicks in: Table~\ref{tab:compactions-before-medal} counts, per task, how many compactions occur before a run first reaches any medal. Across the 20/40 runs that achieve a medal, the model keeps making progress after 3.25 compactions on average (range 0--19) before its first medal, so repeated compaction cycles do not by themselves prevent the run from reaching a medal.

\textbf{Performance drops under the shrunk budget.} Table~\ref{tab:context-management} compares self-selected and oracle (best-of-pool) any-medal rate and mean percentile, with 95\% CIs, between the full-context and shrunk-context \agentic arms, on the full \tasksplit{10} task subset ($n=40$ runs per arm, 10 tasks $\times$ 4 seeds). Performance drops under the shrunk budget at both selection strategies, though the CIs overlap.

\begin{table}[!t]
\centering
\caption{\textbf{\agentic under a shrunk context/output budget.} Any-medal rate ($\uparrow$) and mean percentile ($\uparrow$), with 95\% CIs, self-selected and oracle (best-of-pool), on the full \protect\tasksplit{10} task subset (10 tasks $\times$ 4 seeds, $n=40$ per arm).}
\label{tab:context-management}
\begin{tabular}{lcccc}
\toprule
& \multicolumn{2}{c}{Self-selected} & \multicolumn{2}{c}{Oracle} \\
Group & Any-medal & Mean percentile & Any-medal & Mean percentile \\
\midrule
Malena 1M & 57.5\% [42.2, 71.5] & 77.1 [68.0, 86.2] & 65.0\% [49.5, 77.9] & 78.8 [69.8, 87.8] \\
Malena 64k & 50.0\% [35.2, 64.8] & 73.5 [65.3, 81.7] & 50.0\% [35.2, 64.8] & 74.5 [66.3, 82.7] \\
\bottomrule
\end{tabular}
\end{table}

We read this drop as evidence that \agentic is sensitive to operating near a hard context ceiling, though the point estimates remain within overlapping 95\% CIs of the full-context arm.

\subsection{\oneshot infrastructure ablation}
\label{app:oneshot-infra}

We extend the experiments reported in Appendix~\ref{app:malena-infra} and ablate different interventions in the \oneshot iteration.
More information about these interventions in Appendix~\ref{app:infra-details}.
We report the average performance per iteration on GLM 5.2 and DeepSeek V4 Pro (Preview) backbones in Table~\ref{tab:oneshot-infra}.
While most interventions provide a modest point-estimate improvement and only \texttt{H} resolving as statistically significant for GLM 5.2, and \texttt{I} for DeepSeek V4 Pro (Preview), only when running all of them together we can observe a larger performance improvement, as shown in Table~\ref{tab:oneshot-infra-pairwise}.

\begin{table}[h]
\centering
\caption{\textbf{\oneshot infrastructure ablation.}
Average performance observed per \oneshot iteration reported with 95\% CIs in the \protect\tasksplit{29} split.
Top set: GLM 5.2 backbone.
Bottom set: DeepSeek V4 Pro (Preview) backbone.}
\label{tab:oneshot-infra}
\input{tables/oneshot-infra.tex}
\end{table}

\begin{table}[h]
\centering
\caption{\textbf{Pairwise paired-by-task comparison.}
Average per-iteration difference observed.
Results reported using the \protect\tasksplit{29} task split.
In an A vs B comparison, positive percentile differences indicate better average performance of A.
Percentile difference reported along with 95\% CIs.
Top set: GLM 5.2 backbone.
Bottom set: DeepSeek V4 Pro (Preview) backbone.}
\label{tab:oneshot-infra-pairwise}
\input{tables/oneshot-infra-pairwise.tex}
\end{table}

\section{Additional experiments with external harnesses}
\label{app:baselines}

\subsection{Detailed external harness description}

\textbf{MLEvolve.}~\citep{du2026mlevolveselfevolvingframeworkautomated} MLEvolve maintains a \emph{tree} of candidate solutions, each node a
complete training script with a single parent, and expands it by UCT: nodes carry visit
counts and a backpropagated reward, and selection descends by the usual
upper-confidence-bound rule with an exploration constant that decays over the run. The
backpropagated reward is not the task metric but a small hand-designed signal --- a bonus
for beating the incumbent, a penalty for a node that fails to run. Past the halfway point
of the budget the selector blends UCT with rank-weighted sampling from a global,
greedily-ranked top-$K$, so exploration gives way to exploitation on a schedule rather than
through the search rule itself. Six operators produce nodes: \texttt{draft},
\texttt{improve}, \texttt{evolution} (improve using the branch's own trajectory),
\texttt{debug}, and two cross-branch operators, \texttt{fusion} and \texttt{fusion\_draft},
which synthesize a candidate from the best nodes of several branches. Crossover therefore
flows through the \emph{prompt} rather than the topology: a fusion node still has exactly
one parent, with the other candidates injected as reference text, so the structure is a tree
with cross-branch information flow and not a genealogical graph. Three candidates execute
concurrently on one GPU. It commits, greedily and with no end-of-run re-ranking, whichever
submission had the best self-reported validation metric at the time it was produced --- a
number an LLM extracts from the candidate's own stdout. It also keeps a hybrid BM25+FAISS
retrieval memory over past attempts within a run, indexing each successful node by its plan
and a code summary (and each debug node by the error it was repairing) and querying it when
drafting, improving and debugging; we enable it in all runs reported here.

\textbf{AiScientist.}~\citep{chen2026autonomouslonghorizonengineeringml} AiScientist has no search structure: it is a single conversation
stepping through tool calls, and what to do next is whatever tool the model names. State
persists in one mutable git repository plus two append-only markdown logs, one for
implementation and one for experiments; whenever the agent implements or runs an experiment,
the harness greps the most recent block of the other log and prepends it. This
``file-as-bus'' is the whole of its memory --- there is no retrieval index, and the only
other summarization fires on context overflow. Its operators are \texttt{analyze\_data},
\texttt{prioritize\_tasks}, \texttt{implement} (whose modes cover full/fix/explore/refine/ensemble, so debugging is a mode rather than a separate operator), \texttt{run\_experiment},
\texttt{spawn\_subagent} and \texttt{submit}, alongside shell primitives. Work is strictly
sequential; subagents run blocking, and the only concurrency is whatever background training
the model chooses to launch inside its sandbox. Most consequentially for
Section~\ref{sec:baselines-mlebench}, the final choice among candidates is made by the \emph{agent} rather
than by harness code: no code ranks candidates, but the agent is required, as a step before
submitting, to compare its candidates against the registry and its experiment log and copy
the best to the canonical submission path. It does so substantively --- where it records the
decision, it cites the cross-validation or hold-out protocol it designed, per-candidate
scores under that protocol, and a rationale comparing them (for instance choosing a blended
ensemble at OOF $F_1$ $0.944$ over its two components at $0.942$ and $0.938$). Its
self-selected number is therefore a genuine self-selection on its own validation signal, as
for the other methods; the difference is that the mechanism is prompt-driven rather than
enforced, and the structured record of it is present in only a minority of runs.

\textbf{Arbor.}~\citep{jin2026generalistautonomousresearchhypothesistree} A research agent built around a Coordinator/Executor loop over a
persistent ``Idea Tree'': the Coordinator maintains the tree of candidate ideas and their
outcomes, the Executor implements and evaluates a chosen idea in an isolated git worktree
per experiment, and the loop expands whichever branch of the tree looks most promising.
Two run modes exist upstream: a native CLI over a swappable backbone LLM, and a
keyless coding-agent skill-suite mode where the host coding agent's own model does the
reasoning; this paper uses the native-CLI mode.

\textbf{ScienceFlow.}~\citep{zhao2026scienceflowlonghorizonagentml} A config/manifest-driven long-horizon ML-research agent
(\texttt{scienceflow.cli}, verbs \texttt{repl}/\texttt{parallel}/\texttt{monitor}): a run
is specified as a YAML task manifest and executed by one or more parallel exploratory
workers under a chosen concurrency level, orchestrating both the agent's own reasoning
calls and the ML training/evaluation code the workers produce.

\subsection{Detailed MLE-bench results}
\label{app:detailed-mlebench}

In Table~\ref{tab:agentic-vs-baselines-24h-mlebench}, we provide more details on the headline results presented in Figure~\ref{fig:headline-bars} of \malena, along with the different external harnesses we ran, on the \tasksplit{30} split in MLE-bench.
Additionally, Table~\ref{tab:paired-robustness} shows the paired-by-task relative performance for \malena against all other methods.
We find that \malena outperforms every presented baseline at the GLM 5.2 backbone across both performance metrics; obtains strictly better percentile scores for Kimi K3, and either beats or comes up as statistically indistinguishable at the DeepSeek V4 Flash.
The only backbone where \malena presents a significant regression against other baselines is with Gemma 4 31B --- the weaker of all models tested --- where MLEvolve outperforms according to the percentile metric, while maintaining an equivalent medal rate at our current statistical power.

\begin{table}[h]
\centering
\caption{\textbf{MLE-bench performance with different harnesses and backbones.}
Any-medal rate ($\uparrow$) and mean percentile ($\uparrow$),
self-selected vs.\ oracle (best-in-pool), macro-averaged over the full \protect\tasksplit{30}
set (mean over seeds within a task, then mean over tasks) at 24h time budget. A run with
no valid submission counts as the worst possible outcome (no medal, worst percentile)
rather than being excluded from its row's average. Brackets are 95\%
bootstrap CIs (seed-resampling, task set fixed). Sections are grouped by backbone model;
\textbf{bold}/\underline{underline}/\textit{italic} rank the 1st/2nd/3rd best value per
column within a section, among rows on that section's
shared task set.}
\label{tab:agentic-vs-baselines-24h-mlebench}
\input{tables/headline_mlebench.tex}
\end{table}

\begin{table}[h]
\centering
\caption{Paired \agentic vs.\ baseline comparison across all models,
macro-averaged per task, over the \protect\tasksplit{30} task set.
Higher percentile numbers indicate stronger average performance for \agentic.}
\label{tab:paired-robustness}
\input{tables/baselines_pairwise.tex}
\end{table}

Two data caveats apply across methods. First, a run with no valid submission is
imputed at the worst possible percentile rather than dropped (Section~\ref{sec:baselines-mlebench}'s
convention); this creates exact ties between two very different failure modes and gives a
handful of large-magnitude gaps outsized leverage on every method's point estimate or test
statistic. Second, seed counts per task can be uneven, which the per-task-mean-based
methods treat as equally reliable regardless of count. %

\subsection{Tuning effort invested in baseline harnesses}
\label{app:baseline-tuning-effort}

Because \agentic is our own system, a natural concern is that we invested asymmetrically
more integration and tuning effort into it than into the external harnesses it is compared
against. This appendix reports that effort, spanning backbone-compatibility bug fixes and
targeted configuration ablations on MLEvolve, ScienceFlow, and Arbor. In each case, this
work either (i) fixed a genuine defect --- adopted into every canonical batch reported
elsewhere in the paper; (ii) found no reliable improvement, in which case we kept the
harness's original canonical configuration and report the negative result here rather than
omitting it; or (iii) found a genuine backbone-specific configuration effect and adopted the
better setting as canonical --- the one such case is MLEvolve's reasoning effort on
DeepSeek V4 Flash (Appendix~\ref{app:mlevolve-dsv4-reasoning}).

\paragraph{MLEvolve: reasoning effort on DeepSeek V4 Flash.}
\label{app:mlevolve-dsv4-reasoning}
For every other backbone reported above, MLEvolve's canonical batches use that backbone's
own \texttt{reasoning\_effort=max} (or equivalent) setting, matching every other method's
convention. On DeepSeek V4 Flash, we instead adopt \texttt{reasoning\_effort=medium} as
MLEvolve's canonical setting, the only backbone where it departs from \texttt{max}. The
reason is a length effect specific to this backbone: at \texttt{max}, DeepSeek V4 Flash
emits substantially longer reasoning traces than it does at \texttt{medium}. Because
MLEvolve's tree search issues many independent, short LLM calls rather than growing a single
session, per-call reasoning length translates almost directly into wall-clock cost per search
step, so under \texttt{max}'s much longer traces MLEvolve completes only a fraction of the
search steps it completes at \texttt{medium} within a fixed 24h budget.

To quantify this we compare \texttt{max} and \texttt{medium} on an identical 10-task subset
(same tasks, same seed range, same 24h timeout for both arms; Table~\ref{tab:mlevolve-dsv4-reasoning-effort}).
Cutting reasoning effort to \texttt{medium} reduces the median reasoning length per LLM call
by roughly 5$\times$ (37{,}124 to 7{,}069 characters), which more than triples the mean number
of completed search-tree nodes per run (18.5 to 62.8). Per-node quality is essentially
unchanged between the two settings --- the buggy-node rate is within noise (61\% vs.\ 62\%)
--- so the additional search volume at \texttt{medium} is not, on average, lower-quality than
\texttt{max}'s, just far more numerous. That additional search volume alone is enough to move
outcomes substantially: any-medal rate rises from 0\% to 25.0\% (95\% CI [10.0, 30.0]) and mean
percentile from 42.6 (95\% CI [36.0, 48.5]) to 53.1 (95\% CI [44.9, 59.8]).

\begin{table}[h]
\centering
\caption{\textbf{MLEvolve on DeepSeek V4 Flash, \texttt{max} vs.\ \texttt{medium}
reasoning effort}, restricted to an identical 10-task subset and 24h timeout for both arms.
Any-medal rate and none-rate are self-selected outcomes; mean percentile is oracle
(best-in-pool). 95\% CIs are bootstrapped over per-task run lists (2000 resamples), matching
the convention used for the paper's headline tables. Reasoning length is a per-LLM-call
median pooled across all calls in each arm's runs; node counts (mean/run) and the buggy-node
rate are from each run's full search-tree journal.}
\label{tab:mlevolve-dsv4-reasoning-effort}
\input{tables/table_mlevolve_dsv4_reasoning_effort.tex}
\end{table}

\paragraph{Independent tool-calling failures on GLM 5.2.} Before either harness's canonical
GLM 5.2 batch could run, we found and fixed two unrelated bugs that both manifested as
tool-calling failures specific to this backbone. MLEvolve's structured-output calls request a
named (forced) \texttt{tool\_choice}; GLM 5.2 served behind our vLLM deployment enters
grammar-constrained decoding under a forced choice, and its FSM has no transition for the
reasoning tokens the model wants to emit before its answer, so every such call failed with an
internal server error. We fixed this by defaulting MLEvolve's structured-output calls to
\texttt{tool\_choice=auto} (configurable per stage, so backends that do handle a forced choice
reliably are unaffected). Independently, ScienceFlow's LLM client always included a
\texttt{tools}/\texttt{tool\_choice} field in its request payload, even for calls with no tools
to offer (e.g., its metric-output-interpreter step); the same GLM 5.2 deployment rejects an
explicit empty \texttt{tools: []} array outright, which we fixed by omitting the field entirely
rather than sending an empty array. These are two independent root causes, each surfaced by
validating against a live endpoint before launching the canonical batch, not two symptoms of
the same underlying fix.

\paragraph{ScienceFlow: worker count and resource-admission policy.} ScienceFlow's canonical
configuration runs multiple parallel exploratory workers sharing a single GPU (2 workers for
GLM 5.2; a single worker gated by an LLM-based resource-admission heuristic for DeepSeek V4
Flash). Suspecting that this concurrency could itself be a source of resource congestion, we
tested two independent mitigations, one per backbone.

On GLM 5.2, we compared 1 worker against the canonical 2 workers across the full
\tasksplit{30} split (Table~\ref{tab:scienceflow-1w-vs-2w-full-gold30}). Any-medal
rate and mean percentile are statistically indistinguishable between the two worker counts;
the 1-worker point estimate is nominally higher on any-medal rate (43.9\% vs.\ 40.0\%). This
comparison uses the full \tasksplit{30} split, following an inconclusive result on an earlier
11-task subset; CIs at this sample size cannot rule out a modest true effect. We kept the
canonical 2-worker configuration.

\begin{table}[h]
\centering
\caption{\textbf{ScienceFlow (GLM 5.2), num\_workers=1 vs.\ the canonical num\_workers=2},
full \protect\tasksplit{30} coverage, with no data-variant confound
between the two. Both any-medal and mean percentile are oracle (best-of-pool) across every
candidate ScienceFlow harvests per run, flipped so higher is better. 95\%
CIs are bootstrapped over per-task run lists (2000 resamples).}
\label{tab:scienceflow-1w-vs-2w-full-gold30}
\input{tables/table_scienceflow_1w_vs_2w_full_gold30.tex}
\end{table}

On DeepSeek V4 Flash, we replaced ScienceFlow's canonical \texttt{resource\_smart\_llm}
admission mode --- an LLM-based admission heuristic plus a main-agent advisory signal that
decides when a worker may start a new GPU job --- with the static rule-based
\texttt{resource\_smart\_policy}, on the same canonical 10-task ablation subset used by
Appendix~\ref{app:mlevolve-dsv4-reasoning}'s MLEvolve ablation, to check whether the added
complexity of the canonical policy was necessary (Table~\ref{tab:scienceflow-resource-ablation}). It
was not: any-medal rate and mean percentile are both somewhat lower under
\texttt{resource\_smart\_policy}, and the gap is within the noise we see elsewhere at this sample size. We kept
\texttt{resource\_smart\_llm} as canonical.

\begin{table}[h]
\centering
\caption{\textbf{ScienceFlow (DeepSeek V4 Flash), resource-admission ablation}, identical
10-task subset and 24h timeout as Appendix~\ref{app:mlevolve-dsv4-reasoning}'s MLEvolve
ablation. Both any-medal and mean percentile are oracle (best-of-pool) across every candidate
ScienceFlow harvests per run, flipped so higher is better, matching every
other percentile column in this paper. 95\% CIs are bootstrapped over per-task run lists (2000
resamples), matching the convention used for the paper's headline tables
(Section~\ref{sec:baselines-mlebench}).}
\label{tab:scienceflow-resource-ablation}
\input{tables/table_scienceflow_resource_ablation.tex}
\end{table}

\paragraph{Arbor: executor timeout and convergence early-stopping (GLM 5.2).} Arbor's canonical
configuration for both GLM 5.2 and DeepSeek V4 Flash uses \texttt{reasoning\_effort=max}, a
1M-token context window, 3 parallel executors, and a 4-hour timeout per individual experiment
inside its 24h wall-clock budget; the executor timeout and the \texttt{stop\_after=8}
convergence early-stopping threshold ablated below are not values we chose --- both are
ported unmodified from the \texttt{mle\_kaggle} competition-optimization plugin bundled in
Arbor's own public GitHub repository, so these ablations test Arbor's own maintainers'
defaults rather than a configuration we picked ourselves. We ran two targeted ablations
against this configuration. First, we tripled the executor timeout to
12 hours; the medal rate is
unchanged (Table~\ref{tab:arbor-executor-convergence-ablation}). Second, we tried relaxing
Arbor's convergence-based early stopping (default \texttt{stop\_after=8} consecutive
non-improving idea-tree expansions) to \texttt{stop\_after=999}, effectively disabling it so a
run keeps exploring until its wall-clock budget runs out rather than stopping early: medal rate
is again unchanged within noise
(Table~\ref{tab:arbor-executor-convergence-ablation}).

\begin{table}[h]
\centering
\caption{\textbf{Arbor (GLM 5.2), executor-timeout and convergence ablations} against the
canonical 4-seed baseline, matched per task. Any-medal is
self-selected, best-of-pool across every scored idea-tree node. 95\% CIs are bootstrapped
over per-task run lists (2000 resamples), matching the convention used for the paper's
headline tables.}
\label{tab:arbor-executor-convergence-ablation}
\input{tables/table_arbor_executor_convergence_ablation.tex}
\end{table}

\subsection{Trajectory and token usage with Kimi K3}
\label{app:kimi-trajectory-tokens}

\begin{figure}[h]
\centering
\includegraphics[width=0.48\linewidth]{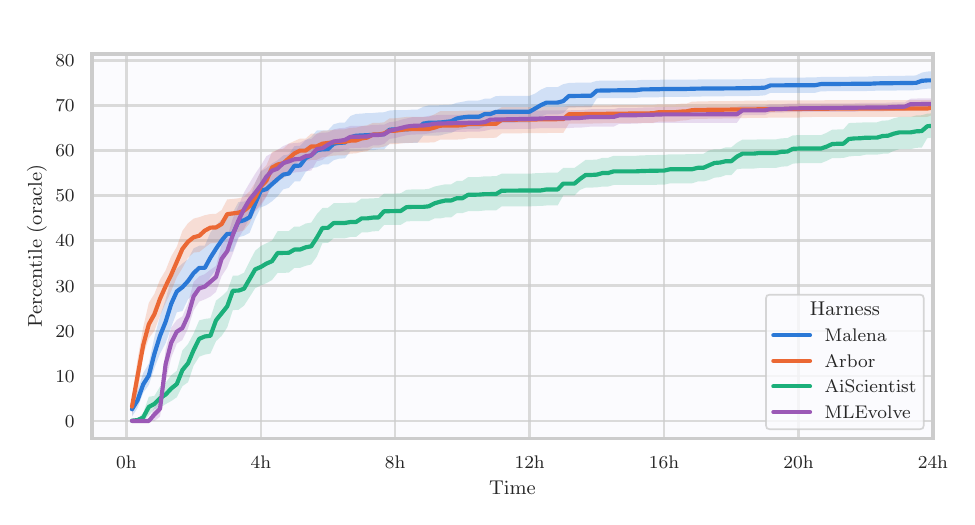}
\hfill
\includegraphics[width=0.48\linewidth]{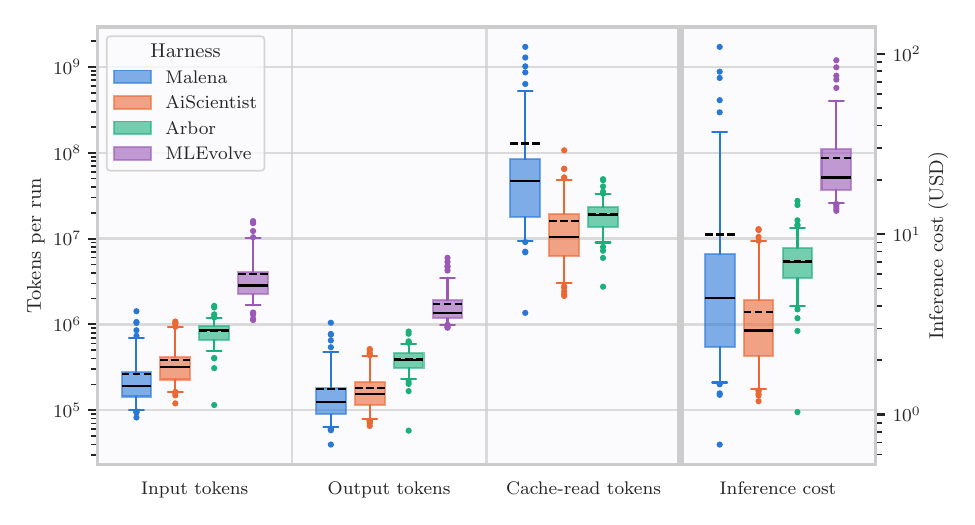}
\caption{\textbf{Left:} oracle best-so-far vs.\ wall-clock time at Kimi K3 for \agentic,
Arbor, AiScientist, and MLEvolve. \textbf{Right:} per-run distribution of cumulative input,
output(+reasoning), and modeled cache-read tokens, plus a USD-cost group at Kimi K3
pricing, for \agentic, Arbor, AiScientist, and MLEvolve. MLEvolve's cache group is omitted
(always exactly 0, same reason as the GLM 5.2 figure).}
\label{fig:trajectory-tokens-kimi}
\end{figure}

Figure~\ref{fig:trajectory-average}'s two panels are GLM 5.2 only. Figure~\ref{fig:trajectory-tokens-kimi} reproduces
both at Kimi K3 using the same \protect\tasksplit{30} task set.

The overall shape matches Figure~\ref{fig:trajectory-average}: quality rises quickly in the
first few hours and plateaus thereafter, with AiScientist again the slowest to converge. On
cost, \agentic's cache-read volume again dominates its token footprint the same way it does
at GLM 5.2, but Kimi K3's much higher output pricing (\$13.99/M vs.\ GLM 5.2's \$3.123/M)
makes MLEvolve's uncached, output-heavy tree search the most expensive of the four shown
here.

\subsection{Quality by task difficulty}
\label{app:clusters}

\begin{figure}[h]
\centering
\includegraphics[width=\linewidth]{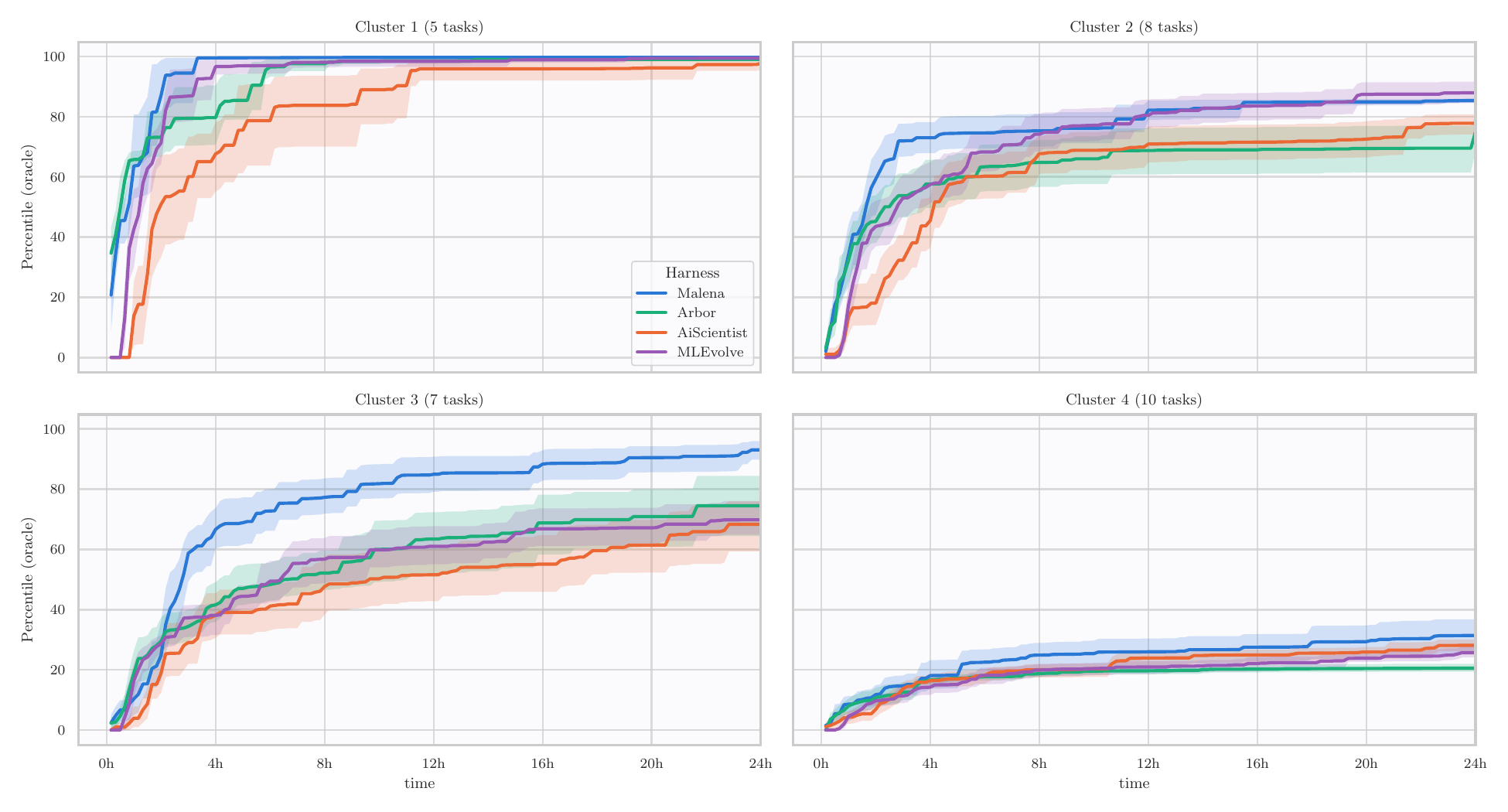}
\caption{Oracle best-so-far vs.\ wall-clock time, faceted by task cluster. \agentic's lead is
largest on cluster 3, smaller but present on clusters 2 and 4, and narrowest on cluster 1,
where every method approaches ceiling performance by 24h.}
\label{fig:trajectory-clusters}
\end{figure}

In order to better understand the strengths of \agentic against other baselines, and better investigate its improved performance, we partition the \tasksplit{30} split into four different clusters, according to perceived difficulty among all harnesses:
\begin{enumerate}
	\item \texttt{nomad2018}, \texttt{plant}, \texttt{spooky}, \texttt{stanford}, \texttt{us}
	\item \texttt{aptos2019}, \texttt{freesound}, \texttt{h\&m}, \texttt{hotel}, \texttt{hubmap}, \texttt{mlsp}, \texttt{tweet}, \texttt{whale}
	\item \texttt{billion-word}, \texttt{cassava}, \texttt{imet}, \texttt{kuzushiji}, \texttt{multi-modal}, \texttt{nfl}, \texttt{petfinder}
	\item \texttt{alaska2}, \texttt{champs}, \texttt{hms}, \texttt{jigsaw}, \texttt{new}, \texttt{osic}, \texttt{smartphone}, \texttt{tensorflow2}, \texttt{uw}, \texttt{ventilator}
\end{enumerate}
We plot the individual oracle performance among these clusters in Figure~\ref{fig:trajectory-clusters}.
On the easiest cluster (1), all methods approach ceiling performance by 24h, though AiScientist lags behind for much of the run before catching up.
On cluster 2, \agentic and MLEvolve maintain a sustained edge over Arbor and AiScientist through 24h.
\agentic performs significantly better than all other baselines on cluster 3, its clearest lead of the four clusters, suggesting a sweet spot between tasks that are inherently difficult for current AI agents and tasks that are comparatively easy.
On the hardest cluster (4), \agentic keeps a smaller but consistent edge over the other three, which track closely together.

\subsection{NatureBench: evaluation protocol}
\label{app:naturebench-protocol}

For NatureBench, we follow the benchmark's own online evaluation protocol as published
(Appendix~\ref{app:naturebench}): rather than a single held-out grader run once at the end
of a run, as in MLE-bench, each task exposes a live \texttt{/evaluate} HTTP endpoint that an
agent may query an unbounded number of times over the course of a run, with the endpoint
automatically tracking the best score seen across all calls. We give every harness we
compare here --- \agentic, AiScientist, and MLEvolve --- unrestricted access to this
endpoint, matching NatureBench's protocol as published.

We extend the per-run time budget to 8h, double NatureBench's own published default of 4h,
to give the slower of our search-based baselines (MLEvolve, whose tree search advances
through many short, sequential steps rather than one long session) a more comfortable
margin within which to complete meaningful search.

Because the harness's own protocol reports the best score achieved anywhere in the run ---
an oracle score, as described in Appendix~\ref{app:naturebench} --- rather than a
self-selected final submission, this metric implicitly rewards however many independent
attempts a harness manages to get graded within the time budget, on top of whatever quality
gains its search process makes. Table~\ref{tab:nature-evaluate-calls} reports the number of
\texttt{/evaluate} calls per run for each harness, across our full NatureBench run cohort.
\agentic and AiScientist call the endpoint at a broadly similar rate
(median 16 and 18 calls per run, respectively); MLEvolve calls it substantially less often
(median 8 calls per run). Since a benefit under an oracle-best-of-run metric scales with the
number of independent scoring opportunities exercised, this asymmetry may put MLEvolve at a
mechanical disadvantage on this particular metric, independent of any difference in the
quality of the strategies it discovers.

\begin{table}[h]
\centering
\caption{\textbf{Number of \texttt{/evaluate} calls per run on NatureBench.} 25th/50th/75th
percentiles across our full NatureBench run cohort, pooling both the GLM 5.2 and Kimi K3
backbones.}
\label{tab:nature-evaluate-calls}
\begin{tabular}{lccc}
\toprule
& \agentic & AiScientist & MLEvolve \\
\midrule
p25 & 8 & 7 & 2 \\
p50 (median) & 16 & 18 & 8 \\
p75 & 36 & 96 & 18 \\
\bottomrule
\end{tabular}
\end{table}

\section{Trace Analysis}
\label{app:checkpoint-methodology}

This appendix gives the full methodology behind Section~\ref{sec:trace}'s checkpoint
labeling and the technique-overlap analysis it enables, including the labeling
schema, the offline deduplication/clustering procedure, and technique-name
normalization we omitted from the main text for space. Labeling was run on a fixed set of
10 original MLE-bench tasks (\texttt{aptos2019}, \texttt{billion-word-imputation},
\texttt{cassava}, \texttt{champs}, \texttt{freesound}, \texttt{h\&m}, \texttt{jigsaw},
\texttt{nfl}, \texttt{petfinder}, \texttt{ventilator}) -- not to be confused with the
\protect\tasksplit{10} challenging split used elsewhere in this paper, which is a different
set of 10 tasks. The checkpoint-mining and
labeling steps below (through canonicalization) are the current pipeline behind
Figure~\ref{fig:technique-category-share-decay} and
Figure~\ref{fig:technique-rarity-heatmap}. Unless stated otherwise, the
labeling/judge model throughout is GLM 5.2.

\subsection{Checkpoint mining}
\label{app:checkpoint-mining}

\textbf{Checkpoints.} We unify run traces across methods to a common checkpoint
format: a snapshot of the run's code repository at specific points in time,
method-dependent since each harness structures its own trajectory differently
-- for \agentic, immediately before a long job submission, a long bash tool
call, or a submission-registration tool call; for MLEvolve, before executing a
candidate; for AiScientist, before a long bash tool call; for Arbor, before a
job-submit call. \oneshot and Completions have no intermediate trajectory to
snapshot and are labeled once, on the final state of the repository. Runs with
multiple checkpoints are labeled sequentially along this chain, each job
seeded with the entries already logged for its nearest earlier checkpoint so
it only has to add, remove, or revise entries rather than re-deriving the
whole technique set from scratch.

\textbf{Labeling schema (pass 1: technique extraction).} The judge (GLM 5.2) reads a
checkpoint's code and logs zero or more entries, each with: a free-form
\texttt{snake\_case} name (no reuse-search against prior entries -- naming
convergence is handled later, offline); 1--3 \texttt{categories}, ordered
most-to-least applicable, drawn from a fixed 13-way taxonomy (\texttt{data\_preprocessing}, \texttt{data\_pipeline}, \texttt{feature\_engineering},
\texttt{augmentation}, \texttt{architecture}, \texttt{loss},
\texttt{regularization}, \texttt{optimization}, \texttt{training},
\texttt{ensembling}, \texttt{model\_selection}, \texttt{post\_processing},
\texttt{inference}, plus an \texttt{other} escape hatch (used for only
0.02\% of all logged category-tag annotations across our labeled
checkpoints), which we exclude from the category-share figures throughout
this section); and three separate
text fields instead of one blob -- \texttt{mechanism} (task-agnostic, roughly
40 tokens, describing only WHAT IS COMPUTED, stripped of domain nouns and
hyperparameters, the only field ever embedded for similarity/dedup),
\texttt{instantiation} (task-specific detail, hyperparameters, and domain
nouns, kept for search but never embedded), and \texttt{evidence\_sig} (a 1--3
line canonical code excerpt for human/reviewer sanity-checking). Splitting
mechanism from instantiation this way removes a specific noise source found
in an earlier embedding-similarity audit, where chunk embeddings picked up
code-context co-occurrence (e.g.\ dropout/mlp/layer\_norm clustering together
purely because they appear near each other in a forward pass) rather than
technique identity.

\textbf{Deduplication and canonicalization (offline, per task).} All identity
resolution -- deciding whether two independently-logged entries describe the
same underlying technique -- happens offline, never at labeling time, and
entirely within one task's own runs (never pooled across tasks, so no
task's vocabulary is built from or reviewed against another task's runs).
Pass-1 entries are embedded on their \texttt{mechanism} text using \texttt{Qwen3-Embedding-8B}
and grouped by complete-linkage hierarchical clustering at a fixed cosine-similarity
threshold (0.70, picked by sweeping method $\times$ threshold against a
hand-audited corpus for cluster-size and same-run-duplicate violations). An
automated GLM 5.2 review then runs in two phases: a split-review phase flags
multi-member clusters with no category shared by every member and asks GLM 5.2
whether any member should be extracted into its own cluster; a merge-review
phase takes every singleton cluster and asks the judge whether it is really the
same technique as one of its nearest neighbors by mechanism-embedding
similarity. A final canonicalization pass has GLM 5.2 synthesize one canonical
name and mechanism description per cluster that a reader would recognize as
covering every member's variant, rather than mechanically picking one
member's text. Pass 2 then re-labels every run a second time, this time
against the finalized, canonical cluster set: the judge calls
\texttt{log\_known\_technique(cluster\_id, ...)} directly when a checkpoint's
technique matches an existing cluster, and free-form entries otherwise, so a
pass-2 match is never subject to a separate, error-prone post-hoc
nearest-centroid assignment step.

\subsection{Main-text plot methodology}
\label{app:trace-plot-methodology}

This subsection gives the exact construction behind the two main-text plots built on top
of the canonical technique clusters above: the category-share-decay curves
(Figure~\ref{fig:technique-category-share-decay}) and the per-technique rarity distribution
(Figure~\ref{fig:technique-rarity-heatmap}).

\textbf{Category-share decay (half-life).} Figure~\ref{fig:technique-category-share-decay}
tracks a method's technique-category composition as a function of \emph{checkpoint
fraction} $x \in [0, 1]$ (progress through a run, pooled over every checkpoint of every run
of that method). Every technique discovered at fraction $f \le x$ contributes weight
$0.5^{(x-f)/h}$ to $x$'s composition, for half-life $h$: a technique discovered exactly at
$x$ has full weight 1, one discovered $h$ earlier has weight 0.5, and older ones decay
smoothly toward (but never reach) 0. Weighted category-tag counts are normalized to shares
at each $x$. $h \to \infty$ recovers a plain cumulative count of every technique discovered
so far; $h \to 0$ recovers an immediate, local snapshot of only the most recently discovered
techniques. The main text uses $h = 0.25$; Appendix~\ref{app:category-share-decay-robustness}
below checks smaller half-lives and per-task (non-pooled) curves.

\textbf{Technique rarity.}
\label{app:technique-rarity}
Figure~\ref{fig:technique-rarity-heatmap} reports, per method,
the distribution of how rare the canonical technique clusters that method's runs exhibit
are. Rarity is computed independently \emph{per task}, not globally: for a fixed task, pool
every run of every (method, backbone) group on that task, and for each cluster $c$ let $n_c$
be how many of those pooled runs exhibit it; $c$'s rarity is $1 - n_c / n_\text{total}$, the
fraction of that task's pooled runs that did \emph{not} use it. Pooling within the task
across every group (rather than, say, comparing a group only to one fixed baseline) means
rarity reflects how many \emph{runs} overall exhibit a technique, not how many distinct
\emph{groups} do -- a technique used by only a handful of runs stays rare even when those runs
happen to span more than one group (Appendix~\ref{app:pseudo-label-example} gives a
rarity-0.9 example shared across exactly two groups).
Figure~\ref{fig:technique-rarity-heatmap} plots, per method, the resulting distribution of
these per-technique rarity values (pooled across backbones and 10 tasks) into
six windows centered at rarity $0, 0.2, \ldots, 1.0$ (half-width $0.1$).

\subsection{Robustness of the category-share-decay curves}
\label{app:category-share-decay-robustness}

\begin{figure}[h]
\centering
\includegraphics[width=\linewidth]{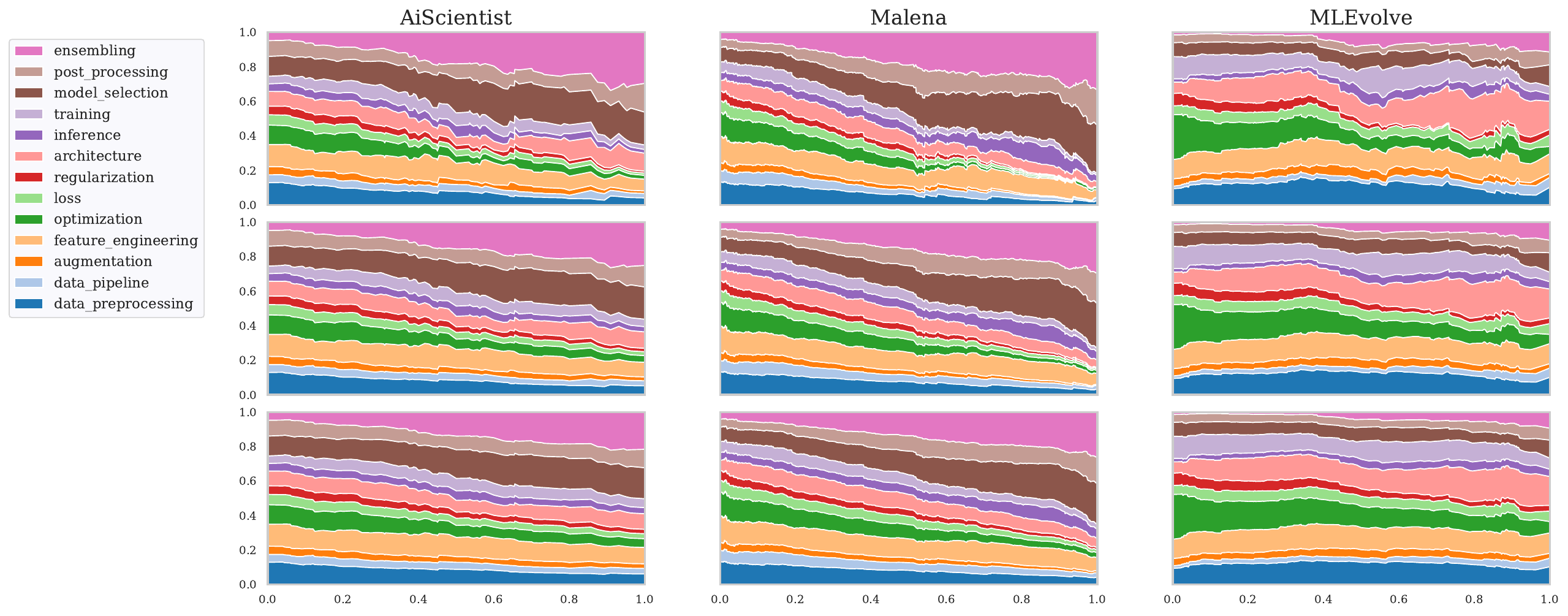}
\caption{Pooled category-share-decay curve (all labeled tasks) at half-lives 0.1, 0.15, and 0.2 (rows, top to bottom) -- smaller half-life is a more local, noisier snapshot of recently-introduced techniques. As in Figure~\ref{fig:technique-category-share-decay}, the x-axis is checkpoint fraction through the run ($\in [0, 1]$) and the y-axis is the decayed share of technique-category tags at that point.}
\label{fig:category-share-decay-pooled-all-hl}
\end{figure}

\begin{figure}[h]
\centering
\includegraphics[width=\linewidth]{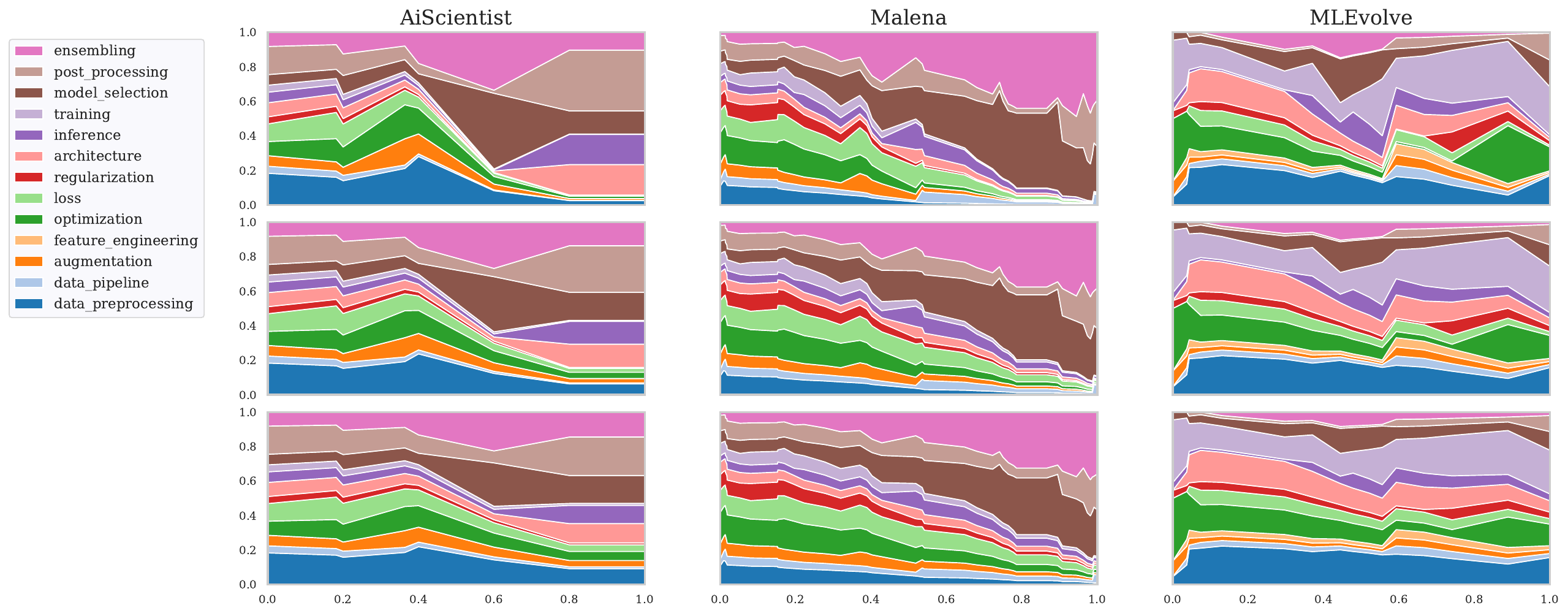}
\caption{Category-share-decay curve on \texttt{aptos2019} alone, at half-lives 0.1, 0.15, and 0.2 (rows, top to bottom). As in Figure~\ref{fig:technique-category-share-decay}, the x-axis is checkpoint fraction through the run ($\in [0, 1]$) and the y-axis is the decayed share of technique-category tags at that point.}
\label{fig:category-share-decay-aptos}
\end{figure}

\begin{figure}[h]
\centering
\includegraphics[width=\linewidth]{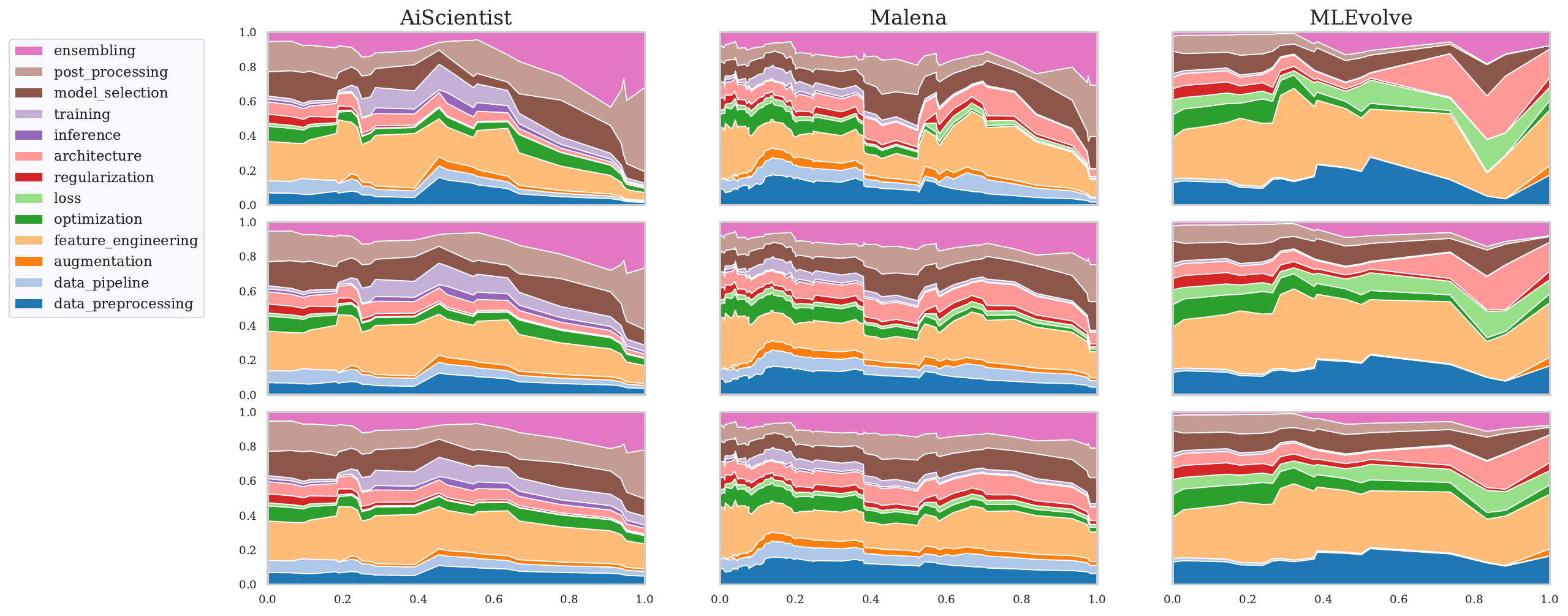}
\caption{Category-share-decay curve on \texttt{nfl} alone, at half-lives 0.1, 0.15, and 0.2 (rows, top to bottom). As in Figure~\ref{fig:technique-category-share-decay}, the x-axis is checkpoint fraction through the run ($\in [0, 1]$) and the y-axis is the decayed share of technique-category tags at that point.}
\label{fig:category-share-decay-nfl}
\end{figure}

Figure~\ref{fig:technique-category-share-decay} in the main text pools all
labeled tasks at a single half-life (0.25). Figure~\ref{fig:category-share-decay-pooled-all-hl}
shows the same pooled construction at three half-lives closer to the main
text's value (0.1, 0.15, 0.2), and
Figures~\ref{fig:category-share-decay-aptos} and~\ref{fig:category-share-decay-nfl}
show the identical three-half-life construction on two individual tasks
(\texttt{aptos2019} and \texttt{nfl})
instead of pooled, to check that the shape of Malena's late-run shift toward
ensembling/model\_selection is not an artifact of the particular half-life or
of pooling across tasks.

\subsection{Per-task technique-category composition across methods}
\label{app:category-profiles-by-task}

Figure~\ref{fig:technique-category-share-decay} and
Figure~\ref{fig:technique-rarity-heatmap} both look at a method's technique
choices pooled across all 10 tasks -- respectively how its category mix
shifts over the course of a run, and how rare its techniques are relative to
every other group. A complementary question is whether methods reach for a
different mix of technique \emph{categories} than each other on the
\emph{same} task, i.e.\ whether task identity or method identity is the
bigger driver of what a method tries. Here we look directly at the
technique-category composition of each method's repertoire on each task
independently.

For a given (task, method) pair, we take the set of every distinct canonical
technique cluster (Appendix~\ref{app:checkpoint-mining} above) any of that
method's backbone/seed runs on that task exhibit, and tally, for each
category tag in the taxonomy above, how many of those clusters carry it (a
cluster tagged with 2 categories contributes to both -- categories are not
split fractionally, so the resulting bar shows a share of category-\emph{tags},
not of techniques). This is a presence-based snapshot (did the method ever
try a category-tagged technique at all over the course of the run), not
usage-frequency-weighted like Figure~\ref{fig:technique-category-share-decay}'s
decay curves -- deliberately simpler, since here we only need one static
composition profile per (task, method) rather than a time series.

\begin{figure}[h]
\centering
\includegraphics[width=\linewidth]{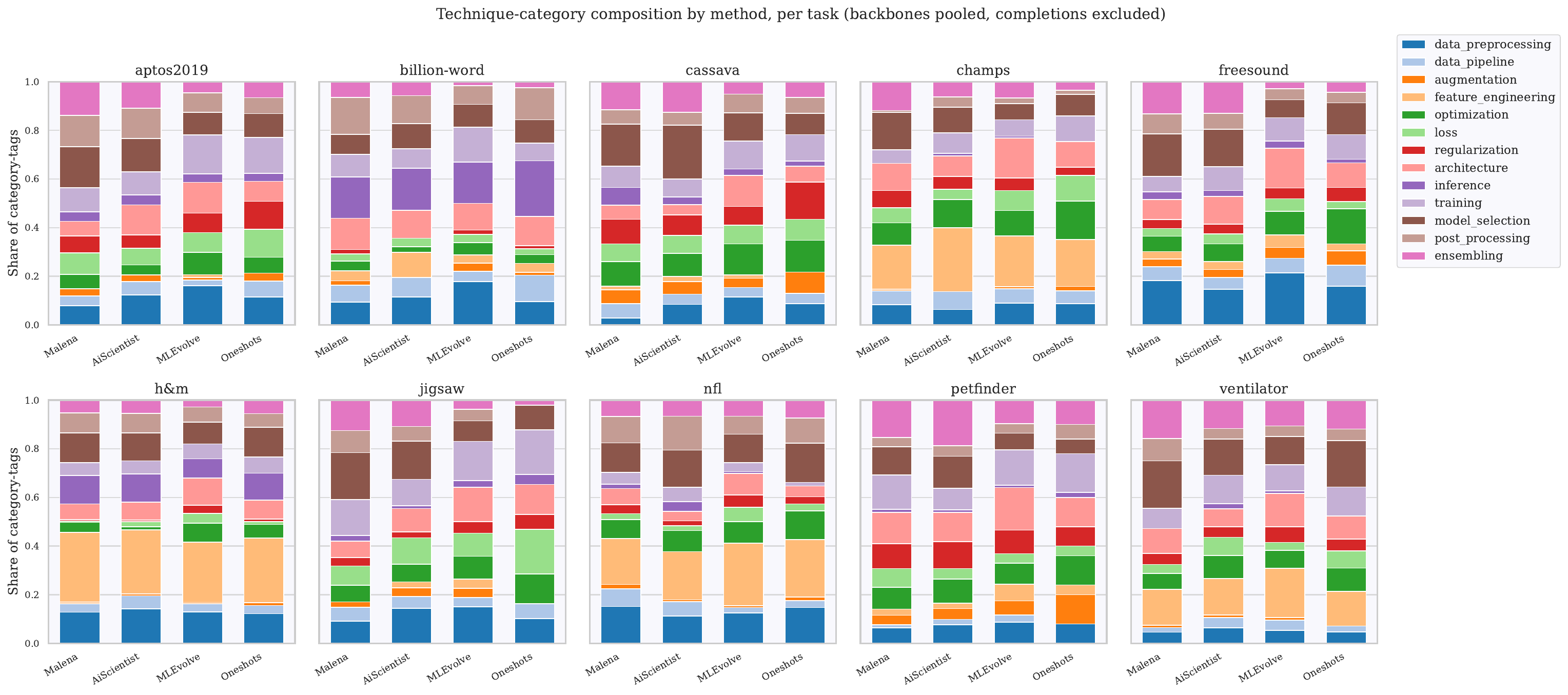}
\caption{Technique-category composition of each method's technique
repertoire, per task (backbones pooled, \texttt{completions} excluded). The
x-axis is method (\agentic, AiScientist, MLEvolve, Oneshots); the y-axis is
share of category-tags. Bar
segments are shares of category-tags (a technique tagged with multiple
categories counts toward each), not shares of techniques.}
\label{fig:category-profiles-by-task}
\end{figure}

Figure~\ref{fig:category-profiles-by-task} shows the resulting composition
for all 10 tasks. Two things stand out. First, task identity is a strong
driver on its own: \texttt{feature\_engineering} dominates every method's
profile on \texttt{champs},
\texttt{h\&m}, and
\texttt{nfl}, while \texttt{ensembling} dominates on
\texttt{petfinder} and, for Malena and AiScientist
specifically, on \texttt{aptos2019},
\texttt{cassava}, and
\texttt{jigsaw} as well. Second,
on top of that shared task-driven baseline, Malena and AiScientist's
profiles track each other closely on most tasks (e.g.\
\texttt{aptos2019},
\texttt{cassava},
\texttt{freesound},
\texttt{nfl}), while MLEvolve is the most frequent
outlier -- visibly heavier on \texttt{data\_preprocessing} and lighter on
\texttt{ensembling} on several of those same tasks, consistent with its
pooled under-representation of ensembling growth in
Figure~\ref{fig:technique-category-share-decay}.
The task \texttt{jigsaw} shows the largest
cross-method spread: Oneshots' profile there is missing most of the
\texttt{ensembling} share the other three methods have, replaced by a much
larger \texttt{loss}/\texttt{optimization} block.

\subsection{Illustrative session excerpts}
\label{app:trace-excerpts}

The excerpts below are verbatim (only trimmed for length, marked with \texttt{...}) pulls
from real session transcripts, given as illustrative examples of the reasoning/tool-call
structure behind the checkpoint-labeling claims in Section~\ref{sec:trace}. \texttt{[bash]}
lines are literal shell commands the agent issued; \texttt{<reasoning>} blocks are the
agent's own prose immediately before or after them.

\textbf{Example 1: \agentic, GLM 5.2, \texttt{tweet}.}
\begin{quote}
\small
\begin{verbatim}
<reasoning>
...Let me train deberta with 4 epochs using pseudo_test_v2, tag debv3pl2...
</reasoning>
[bash] python src/train.py --model microsoft/deberta-v3-large --tag debv3pl2
       --lr 1e-5 --epochs 4 --extra_train work/pseudo_test_v2.csv
-> [fold 0 epoch 3] val_jaccard=0.7211 ... best val_jaccard per fold
<reasoning>
Iterated pseudo-labels + 4 epochs: OOF = 0.7212 (vs 0.7199 for v1 pseudo + 3
epochs). +0.0013 improvement. Fold scores: fold 0: 0.7223 -> 0.7211 (slightly
down); fold 1: 0.7123 -> 0.7139 (up); fold 2: 0.7237 -> 0.7249 (up); fold 3:
0.7169 -> 0.7189 (up); fold 4: 0.7245 -> 0.7271 (up). Overall OOF 0.7199 ->
0.7212. Let me compute S1 and per-sentiment for debv3pl2.
</reasoning>
[bash] python src/infer.py --model microsoft/deberta-v3-large --tag debv3pl2
       --out submission_v5_debv3pl2.csv
[submissions_register] Registered submission... Total number of submissions:
5. Best validation score so far: 0.7212.
\end{verbatim}
\end{quote}

\textbf{Example 2: MLEvolve, GLM 5.2, \texttt{aptos2019}.}
\begin{quote}
\small
\begin{verbatim}
<reasoning>
The evolution trajectory reveals severe overfitting (train loss drops below
0.10 while validation loss plateaus > 0.60), indicating that the highly
capable ConvNeXt base model easily memorizes this small dataset. To break
this stagnation, we must apply a Tier 2 shift focusing on robust
regularization rather than just tweaking losses or backbones. By
introducing Mixup data augmentation, stochastic depth (Drop Path), and
feature-level dropout, we force the model to learn smoother, generalized
decision boundaries between severity grades, effectively combatting
single-landscape overfitting.
</reasoning>
\end{verbatim}
\end{quote}

\textbf{Example 3: AiScientist, GLM 5.2, \texttt{aptos2019}.}
\begin{quote}
\small
\begin{verbatim}
<reasoning>
Now I have a clear picture. Let me summarize the state and plan refinements:

Current state:
- 8 models, 5-fold CV, ensemble OOF QWK = 0.9290
- Best single models: cnexts_384 (0.9236), efnb5_512 (0.9231)
- Models highly correlated (0.96-0.99) -- limited diversity gain
- Huge raw->tuned QWK gap (0.09-0.13) -- continuous scores poorly calibrated
- Training is noisy (raw_qwk bounces across epochs)
- Training takes ~130 min/model

Refinement plan (low-risk, high-impact, no broad rewrite):
1. Ensemble weighting refinement (no retraining): rank-averaging,
   z-score normalization, greedy forward selection -- cheap, could
   push 0.9290 -> 0.93+
2. Threshold tuning quality -- better optimization
3. Training stability (one refined retrain): SWA, be... [truncated in log]
</reasoning>
[bash] cd /home/code && python3 -c "
import pandas as pd, numpy as np
from utils import OptimizedRounder, qwk
from scipy.optimize import minimize

oof = pd.read_csv('artifacts/ensemble_kfold_oof.csv')
model_cols = ['score_efnb5_456','sco... [truncated in log]
-> [tool_result] result_len=1233 (stdout not recoverable in this log format)
\end{verbatim}
\end{quote}
This is the final record of the subagent's session (30 total records): after
inspecting all 8 trained models' summaries/histories over 9 prior turns
(git log, config/dataset/model/utils reads, per-model kfold summaries,
training histories, GPU/torch versions), it explicitly frames the decision
as ``no broad rewrite'' and chooses an ensemble-reweighting analysis over
\texttt{artifacts/ensemble\_kfold\_oof.csv} rather than editing any training
code (14 bash + 6 \texttt{read\_file\_chunk} calls, 0 \texttt{edit\_file}
calls across the whole subagent session).

\textbf{Example 4: \agentic, GLM 5.2, \texttt{whale} --
the two-stage pipeline referenced in Section~\ref{sec:trace}.}
\label{app:whale-example}
Within its first 17 checkpoints, \agentic settles on a two-stage pipeline and
never changes its architecture again over the remaining 63 (80 total): a
metric-learning embedder, followed by nearest-neighbor retrieval with a
learned novelty detector.
\begin{quote}
\small
\begin{verbatim}
<reasoning> (checkpoint 000)
...Compute embeddings for all training images and test images... find nearest
training images... Insert new_whale based on threshold... Use validation
set to find optimal threshold.
</reasoning>
\end{verbatim}
\end{quote}
Stage 1 (\texttt{src/train.py}) trains a CNN/ViT backbone with an ArcFace
loss to produce normalized whale-fluke embeddings. Stage 2
(\texttt{src/retrieve.py}, \texttt{src/centroid\_submit.py},
\texttt{src/detector.py}) computes cosine-similarity nearest neighbors
against the training embeddings, ranks top-5 candidates via per-class
centroids, and uses a small logistic-regression classifier over
similarity-gap features (top-1/2/3 similarity, gaps, top-5 mean/std) to
flag \texttt{new\_whale} queries -- individuals never seen in training.
All 63 subsequent checkpoints are targeted patches to this fixed pipeline
rather than new stages: backbone swaps (EfficientNetV2 $\to$ ConvNeXt
$\to$ Swin), input-resolution bumps, novelty-threshold retuning, and
ensembling. One checkpoint (\texttt{014\_ms\_centroid\_fAB}) illustrates the
granularity of these patches:
\begin{quote}
\small
\begin{verbatim}
<reasoning> (checkpoint 014_ms_centroid_fAB)
...The gap between 0.663 and 0.722 oracle is the new-detection error. A
better new-detector could use... but let me first improve the embedding...
</reasoning>
\end{verbatim}
\end{quote}
That is, the agent is choosing which fixed sub-module of the pipeline to
improve next, at the granularity of a targeted code change, rather than
tuning a numeric hyperparameter or introducing a new pipeline stage.

\textbf{Example 5: independent convergence on soft-label self-training,
\texttt{aptos2019} (GLM 5.2 vs.\ Kimi K3).}
\label{app:pseudo-label-example}
\agentic's GLM 5.2 and Kimi K3 runs on \texttt{aptos2019}
independently converge on the same canonical technique cluster -- soft-label
self-training (pseudo-labeling using an ensemble's own predictions as soft
targets) -- without sharing a run, then diverge in how much surrounding
machinery each backbone builds around it. This cluster has rarity 0.9
(Appendix~\ref{app:technique-rarity}): across every AiScientist, MLEvolve, and
\oneshot run on this task, none use it.

The GLM 5.2 run (logged as \texttt{pseudo\_label\_self\_training}):
\begin{quote}
\small
\begin{verbatim}
Mechanism: Generate soft pseudo-labels by running a trained model on
unlabeled data, combine the pseudo-labeled samples with the original
labeled training set (using soft target distributions and a per-sample
weight for pseudo-labeled examples), and retrain the model on the
augmented dataset.

Instantiation: dataset.py PseudoDataset combines train images (hard
labels, soft-onehot with label smoothing) and pseudo-labeled test images
(soft probs from ensemble predictions). train_pl.py orchestrates per-fold
pseudo-label training. Pseudo-labels sourced from infer.py
--save_test_probs output.
\end{verbatim}
\end{quote}
The Kimi K3 run (logged independently as \texttt{self\_training\_pseudo\_labels}),
landing on the identical mechanism:
\begin{quote}
\small
\begin{verbatim}
Instantiation: train_pseudo.py loads ensemble test probability
predictions (from compute_ensemble_test_probs.py) as soft pseudo-labels
for unlabeled test images. Trains jointly on labeled train data (one-hot
hard targets) and pseudo-labeled test data (soft 5-class probs) via soft
cross-entropy... Pseudo data influence controlled by --pseudo_repeat
(dataset repetition factor)... Supports per-fold warm-start via
--init_from_dir.

Evidence:
comb = ConcatDataset([train_ds_hard, *([train_ds_pseudo] * cfg.pseudo_repeat)])
loss = -(targets * logp)
if weight is not None: loss = loss * weight[None, :].to(device)
\end{verbatim}
\end{quote}
The two runs share the core mechanism but not its surroundings. Before
being folded into the single canonical cluster above during the offline canonicalization
stage, the GLM 5.2 run builds substantially more machinery around it: a
weighted soft-cross-entropy loss (pseudo-labeled samples down-weighted 0.2
vs.\ 1.0 for labeled ones), pseudo-label \emph{sharpening} (raising soft
probabilities to a power $>1$ and renormalizing to peak the target
distribution), iterative multi-round pseudo-labeling (retraining on a
previous round's ensemble output, then keeping whichever round scored better
per model on OOF), and two rounds of brute-force combinatorial search --
first over which round/resolution each ensemble member should use (up to 324
combinations, each scored by out-of-fold threshold-optimized QWK), then over
7 candidate final ensemble configurations. The Kimi K3 run instead builds a
leaner version of the same core loop, adding per-fold warm-starting
(\texttt{-{}-init\_from\_dir}) rather than the GLM 5.2 run's iterative-round
search. Both runs independently rediscover the same underlying technique --
reflected in Figure~\ref{fig:technique-rarity-heatmap} as a rare (0.9) but
recurring point in \agentic's rarity distribution -- while differing in how
much extra engineering each backbone invests around it.

\section{Contamination check}
\label{app:contamination}

Every \tasksplit{30} task is a public Kaggle competition, and MLE-bench itself
ships a set of reference kernels/notebooks for a subset of its competitions (public
solutions collected as part of the benchmark's own construction). We check for
memorized/copied code using a vendored copy of Dolos, MLE-bench's own
source-code similarity tool, against a curated corpus of 1{,}350 distinct real solutions
spanning all 30 \tasksplit{30} tasks.

\textbf{Code similarity (Dolos).} Dolos's default pairwise \emph{similarity} score
normalizes shared token $k$-grams by the union of both files' tokens (a Jaccard-style
measure), which is a poor fit for our setting and, notably, biased in the direction of
\emph{under}-reporting contamination rather than over-reporting it: \agentic's solution
files are typically much smaller than the monolithic Kaggle notebooks in the real corpus,
so even if \agentic fully reused a small fragment of a large real file verbatim, the
shared tokens would be a small fraction of the (large) union and the similarity score
would stay low -- the metric would fail to flag exactly the kind of partial copying we
care about. We instead score every pair by \emph{containment} -- the fraction of
the \emph{smaller-file's} tokens that also appear in the other file, together with the
longest matching token run -- which asks ``how much of this file was drawn from the
other'', independent of how large the other file is, and is sensitive to exactly the
small-fragment-fully-reused case that raw similarity would miss. We compare
\agentic-vs-real containment against a real-vs-real cross-label null baseline (distinct,
independently-written real solutions, pooled across the whole 30-task corpus rather than
restricted to same-task pairs -- see below for why). Table~\ref{tab:contamination-similarity}
reports the containment distribution for both, across all 924{,}750 \agentic-vs-real pairs
and 473{,}850 real-vs-real cross-label pairs (every \agentic checkpoint / every real
solution, pooled across all 30 tasks -- i.e.\ including task-mismatched pairs, not just
same-task ones -- exhaustive): \agentic-vs-real containment sits below real-vs-real
containment at every percentile shown. A pair is flagged only if it exceeds \emph{both}
the real-vs-real containment and longest-match-run 95th percentiles;
\textbf{zero of the 924{,}750 pairs are flagged}. This is the pattern we would expect
from independently-written code, not from copied or memorized solutions. (The embedding
check below restricts to same-task pairs only, to keep the $O(n^2)$ embedding-similarity
cost tractable; the two pair counts are therefore not directly comparable.)

\textbf{Embedding similarity (bag-of-chunks).} As a second, token-independent check, we
embed both corpora with \texttt{Qwen3-Embedding-8B}~\citep{qwen3embedding}, which is a 4096-dim embedding model, splitting each file into a sliding
window of chunks (since many solution files exceed the embedding model's context window)
and representing each solution as a bag of chunk embeddings rather than a single vector.
We compare bags with three metrics: \emph{mean-pooling cosine} (cosine similarity of the
two bags' mean embeddings); \emph{Chamfer similarity} (for each chunk in one bag, its max
cosine similarity to any chunk in the other bag, averaged over the bag, symmetrized by
averaging both directions); and \emph{Chamfer with background subtraction} (the raw
Chamfer score minus a background level -- the median Chamfer score over 5{,}000 random
\agentic-vs-real chunk-bag pairs drawn from \emph{different} tasks, 0.504 -- to remove the
generic code/library-idiom similarity any two unrelated files share). We compute all three
over every same-task \agentic-vs-real pair (62{,}207 pairs) and every same-task
real-vs-real cross-label pair (33{,}465 pairs), exhaustively, across all 30 tasks.
Table~\ref{tab:contamination-similarity} reports the percentiles alongside the Dolos
numbers above. At the median (and, for mean-pooling cosine, up to the 90th percentile),
\agentic-vs-real similarity is comparable to the real-vs-real baseline rather than below
it; but at the percentiles where copied or memorized content would actually show up --
the upper tail, p95 and p99 -- \agentic-vs-real similarity is consistently \emph{below}
the real-vs-real baseline for all three metrics, i.e.\ there is no anomalously heavy right
tail. This is the same qualitative conclusion as the Dolos check, from an independent,
token-free signal.

\begin{table}[h]
\centering
\caption{Contamination-check similarity scores by percentile: \agentic-generated code
vs.\ real \protect\tasksplit{30} solutions, compared against a real-vs-real cross-label null baseline.
Top: Dolos code-similarity containment. Bottom: bag-of-chunk-embedding similarity.}
\label{tab:contamination-similarity}
\begin{tabular}{llcccc}
\toprule
metric & group & p50 & p90 & p95 & p99 \\
\midrule
Dolos containment & real-vs-real & 0.008 & 0.077 & 0.143 & 0.453 \\
Dolos containment & \agentic-vs-real & 0.002 & 0.009 & 0.013 & 0.020 \\
\midrule
mean-pooling cosine & real-vs-real & 0.735 & 0.877 & 0.909 & 0.968 \\
mean-pooling cosine & \agentic-vs-real & 0.773 & 0.880 & 0.896 & 0.919 \\
Chamfer & real-vs-real & 0.634 & 0.741 & 0.778 & 0.882 \\
Chamfer & \agentic-vs-real & 0.638 & 0.713 & 0.726 & 0.746 \\
Chamfer (bg-subtracted) & real-vs-real & 0.130 & 0.237 & 0.273 & 0.377 \\
Chamfer (bg-subtracted) & \agentic-vs-real & 0.134 & 0.209 & 0.222 & 0.242 \\
\bottomrule
\end{tabular}
\end{table}

\textbf{Behavioral check: environment interaction matters.} A separate, model-independent
argument against memorization comes from comparing iteration types rather than code
directly: if \agentic's solutions were largely memorized from pretraining, a single
completion asked to produce a full solution to a competition it has ``seen'' should score
close to the full agentic loop, since no environment feedback would be needed to recall a
memorized answer.
We can clearly observe that such thing is not the case in our corpus.
From Figure~\ref{fig:coding-agent}, we can observe that even \Bootstrap \oneshot iterations aggregated at 6h time budgets performs significantly better than the single \chat iteration.
This is the opposite of what contamination would predict, and is consistent with the agent doing real iterative
problem-solving rather than recalling known answers.

\section{Discussion}
\label{app:discussion}

\subsection{Limitations}
\label{app:limitations}

The primary limitation of this study is that most of our results rely on
MLE-bench~\citep{chan2025mlebenchevaluatingmachinelearning}, which has known issues in its
task preparation scripts and, being built from public Kaggle competitions, is a plausible
target for pretraining contamination. We mitigate this in three ways. First, we identified
and fixed the subset of \tasksplit{30} tasks with broken preparation scripts (missing data
or ground-truth leakage) rather than excluding them, so that our comparisons stay on the
full 30-task set used elsewhere in the paper; see Appendix~\ref{app:data} and
Table~\ref{tab:tasks} for which tasks were fixed. Second, we directly checked for contamination via
two independent methods, a code-similarity (Dolos) check and a bag-of-embeddings
similarity check, both against a curated corpus of real \tasksplit{30} solutions and both finding
no anomalous similarity between \agentic-generated code and real solutions (Appendix~\ref{app:contamination}).
Third, we additionally evaluate on NatureBench~\citep{wang2026naturebenchcodingagentsmatch}
(Appendix~\ref{app:naturebench}), whose tasks are recent open scientific-research questions
with no public solution corpus to memorize, though there we could only afford a smaller
number of models and fewer methods than on \tasksplit{30}.

A second limitation is statistical power. Some of our results still have confidence
intervals wide enough that we cannot make strong claims about the size of an effect; we
believe our weaker claims of \agentic being no worse than a given baseline are well
supported (Appendix~\ref{app:statistics} reports the paired-comparison methodology this
relies on), but stronger claims about exactly how much better should be read with this in
mind. This is a direct consequence of cost: running agents for the full 24-hour budget is
expensive, so we could only afford 3--4 seeds per task at 24 hours, versus 6--10 seeds per
task for our 6-hour and completion-only conditions.

A third limitation is scope on coordination between multiple agent workers. Our claim
that most harness interventions add little on top of a plain coding agent (Section~\ref{sec:systematic})
is made almost entirely with a single worker; we do run one coordination ablation
(Section~\ref{sec:layer5}) across a small set of coordination primitives on top of
multiple parallel workers, but we do not otherwise invest in coordination research
between multiple concurrent streams of work. We defer this to future work, since it is
its own interesting problem, plausibly with a different set of useful interventions than
the single-worker setting, and one that likely requires an even larger compute budget
than what we use here to study properly.

A fourth limitation is possible asymmetry in tuning effort: because \agentic is our own
system, it is natural to worry that we invested disproportionately more debugging and
configuration effort into it than into the external harnesses it is compared against,
inflating its apparent advantage. We mitigate this by treating every harness's backbone
port the same way we treat our own: each external harness went through its own
backbone-specific verification pass before its canonical batch was launched, which
surfaced and required fixing genuine incompatibility bugs in the harnesses' own code
(e.g., independent tool-calling failures in both MLEvolve and ScienceFlow specific to
GLM 5.2), and beyond bug-fixing we additionally ran targeted configuration ablations on
ScienceFlow and Arbor (worker count, resource-admission
policy, executor timeout, and early-stopping aggressiveness) to check whether their
canonical settings were leaving performance on the table.
Appendix~\ref{app:baseline-tuning-effort} reports this work in full; none of these
ablations moved outcomes by more than run-to-run noise at our sample sizes, so we kept
each harness's original canonical configuration, but we surface the negative results
here rather than reporting only the positive tuning story we have for our own system.

\subsection{Broader Outlook}
\label{app:outlook}

Beyond the direct ablation results, our observations across backbones and prior harnesses point at two related hypotheses about why scaffolding is losing its edge, which we lay out here as motivation for future work rather than claims this paper establishes.

\textbf{Post-training ate the harness.} Frontier LLMs are increasingly post-trained directly inside a coding-agent loop -- rewarded for planning, executing, and debugging across many turns of real tool use -- rather than only on single-turn completions. We hypothesize that this kind of post-training substitutes for exactly the scaffolding that a plain coding-agent environment (\oneshot) would otherwise need on top of a chat completion (\chat): search over what to try next, recovery from a failed step, deciding when a result is good enough to stop on. If so, the \chat-vs-\oneshot gap (Section~\ref{sec:layer1}) should track how much of this coding-agent post-training a model has received, more than it tracks raw model scale. We see suggestive evidence for this in the DeepSeek V4 family: the Preview releases show close to no gap between \chat and \oneshot performance, while the final releases -- built on comparable base models but with substantially more coding-agent post-training -- open up a clear gap, though still narrower than the gap we measure for GLM 5.2 (Section~\ref{sec:layer1}). If this pattern holds more broadly, it reframes a common motivation behind self-evolving-harness designs, where an agent is given the ability to rewrite its own scaffolding to better match its abilities~\citep{wang2026rethinkingevaluationharnessevolution}: rather than each model needing a harness custom-fit to it at inference time, post-training may already be doing this fitting once, in advance, for every downstream task the model is later pointed at.

\textbf{Coding agents as the bitter lesson for test-time inference.} A second, related reading of our results is architectural rather than about training data: for models capable enough at autonomous tool use, adding hand-designed structure on top of a plain coding-agent session -- explicit search trees, planner/executor splits, multi-worker coordination -- has a poor and often negative return (Section~\ref{sec:layer2}--\ref{sec:layer5}), while simply running more (and more capable) coding-agent post-training inside that same plain loop could potentially perform better. This is the same shape of argument Sutton makes about general-purpose search and learning methods outscaling hand-engineered domain structure as compute grows~\citep{sutton2019bitterlesson}, applied to test-time inference specifically: a basic coding-agent harness plus a stronger backbone tends to close the gap that a specialized harness had opened at a weaker backbone, rather than the specialized harness's advantage compounding. We see the same qualitative pattern in our own results: weaker backbones still benefit measurably from hand-designed workflow priors like MLEvolve's tree search, but as backbone agentic capability increases, \agentic's plain single-session harness closes that gap and matches or exceeds MLEvolve without any of its search machinery (Table~\ref{tab:agentic-vs-baselines-24h-mlebench}, Appendix~\ref{app:detailed-mlebench}). Taken together with the first hypothesis, the practical implication is the same one drawn in Section~\ref{sec:conclusion}: the harness is not where the durable advantage lives.

%% file: tables/mlebench-tasks.tex
\begin{tabular}{llc}
\toprule
 & Full competition ID & Split \\
\midrule
alaska2*\textsuperscript{\textdagger}\textsuperscript{\textbardbl} & alaska2-image-steganalysis & Medium \\
aptos2019 & aptos2019-blindness-detection & Lite \\
billion-word* & billion-word-imputation & Medium \\
cassava*\textsuperscript{\textdagger} & cassava-leaf-disease-classification & Medium \\
champs*\textsuperscript{\textdagger}\textsuperscript{\textdaggerdbl} & champs-scalar-coupling & Medium \\
freesound*\textsuperscript{\textdagger} & freesound-audio-tagging-2019 & Medium \\
h\&m & h-and-m-personalized-fashion-recommendations & Medium \\
hms & hms-harmful-brain-activity-classification & High \\
hotel & hotel-id-2021-fgvc8 & Medium \\
hubmap*\textsuperscript{\textdagger}\textsuperscript{\textdaggerdbl} & hubmap-kidney-segmentation & Medium \\
imet*\textsuperscript{\textdagger} & imet-2020-fgvc7 & Medium \\
jigsaw* & jigsaw-unintended-bias-in-toxicity-classification & Medium \\
kuzushiji*\textsuperscript{\textdagger} & kuzushiji-recognition & Medium \\
mlsp & mlsp-2013-birds & Lite \\
multi-modal*\textsuperscript{\textdagger}\textsuperscript{\textdaggerdbl} & multi-modal-gesture-recognition & Medium \\
new* & new-york-city-taxi-fare-prediction & Lite \\
nfl*\textsuperscript{\textdagger} & nfl-player-contact-detection & High \\
nomad2018 & nomad2018-predict-transparent-conductors & Lite \\
osic* & osic-pulmonary-fibrosis-progression & Medium \\
petfinder*\textsuperscript{\textdagger} & petfinder-pawpularity-score & Medium \\
plant & plant-pathology-2021-fgvc8 & Medium \\
smartphone\textsuperscript{\textdaggerdbl} & smartphone-decimeter-2022 & High \\
spooky & spooky-author-identification & Lite \\
stanford & stanford-covid-vaccine & High \\
tensorflow2 & tensorflow2-question-answering & Medium \\
tweet & tweet-sentiment-extraction & Medium \\
us & us-patent-phrase-to-phrase-matching & Medium \\
uw & uw-madison-gi-tract-image-segmentation & Medium \\
ventilator & ventilator-pressure-prediction & Medium \\
whale & whale-categorization-playground & Medium \\
\bottomrule
\end{tabular}

%% file: task-fixes/champs.tex
\textbf{\texttt{champs-scalar-coupling}: missing test geometry.}
The original script kept \texttt{structures.csv} and the per-molecule \texttt{.xyz} files only for training molecules.
On Kaggle, the 3D structures are provided for both train and test (only the five quantum-chemistry ``additional data'' files are train-only).
We keep structures for every train and test molecule.
\begin{quote}
\scriptsize
\begin{verbatim}
--- a/mlebench/competitions/champs-scalar-coupling/prepare.py
+++ b/mlebench/competitions/champs-scalar-coupling/prepare.py
+    all_molecule_names = (set(new_train["molecule_name"])
+                          | set(new_test["molecule_name"]))
     structures = read_csv(raw / "structures.csv")
-    structures = structures[
-        structures["molecule_name"].isin(new_train["molecule_name"])]
+    structures = structures[
+        structures["molecule_name"].isin(all_molecule_names)]
 ...
-    data_csvs = {
-        "structures": structures,
+    assert set(structures["molecule_name"]) == all_molecule_names
+    train_only_csvs = {
         "dipole_moments": dipole_moments,
         ...
 ...
     for molecule_name in tqdm(
-        new_train["molecule_name"].unique(), desc="Copying ..."
+        sorted(all_molecule_names), desc="Copying ..."
     ):
\end{verbatim}
\end{quote}

%% file: task-fixes/hubmap.tex
\textbf{\texttt{hubmap-kidney-segmentation}: leaked test annotations.}
The original script copied \texttt{\{image\_id\}.json} into \texttt{public/test/}.
This file is the unencoded glomerulus polygon annotation, i.e.\ the same information as the RLE-encoded target.
Kaggle's real test set only ships the \texttt{-anatomical-structure.json} sidecar.
We drop the annotation from the test copy and assert that only the sidecar remains.
\begin{quote}
\scriptsize
\begin{verbatim}
--- a/mlebench/competitions/hubmap-kidney-segmentation/prepare.py
+++ b/mlebench/competitions/hubmap-kidney-segmentation/prepare.py
     for image_id in tqdm(new_test["id"], desc="Copying test images"):
         shutil.copy(raw / "train" / f"{image_id}.tiff",
                     public / "test" / f"{image_id}.tiff")
-        shutil.copy(raw / "train" / f"{image_id}.json",
-                    public / "test" / f"{image_id}.json")
         shutil.copy(
             raw / "train" / f"{image_id}-anatomical-structure.json",
             public / "test" / f"{image_id}-anatomical-structure.json",
 ...
+    assert len(list((public / "test").glob("*.json"))) == len(new_test)
\end{verbatim}
\end{quote}

%% file: task-fixes/smartphone.tex
\textbf{\texttt{smartphone-decimeter-2022}: leaked reference trajectory.}
The original script copied each test drive's full raw directory into \texttt{public/test/} and removed only \texttt{ground\_truth.csv}.
The latitude/longitude in \texttt{ground\_truth.csv} is linearly interpolated from the reference NovAtel SPAN receiver's NMEA log, and that log stayed in the public test folder.
We delete every SPAN-reference \texttt{.nmea} file and keep the per-phone onboard-GNSS \texttt{.nmea} logs, which Kaggle legitimately provides.
\begin{quote}
\scriptsize
\begin{verbatim}
--- a/mlebench/competitions/smartphone-decimeter-2022/prepare.py
+++ b/mlebench/competitions/smartphone-decimeter-2022/prepare.py
     for fpath in (public / "test").rglob("ground_truth.csv"):
         fpath.unlink()  # don't include ground truth in public test data
+    span_nmea_files = [
+        fpath for fpath in (public / "test").rglob("*.nmea")
+        if "span" in fpath.name.lower()
+    ]
+    assert len(span_nmea_files) > 0
+    for fpath in span_nmea_files:
+        fpath.unlink()  # reference trajectory ground truth is derived from
 ...
+    assert not any("span" in fpath.name.lower()
+                   for fpath in (public / "test").rglob("*.nmea"))
\end{verbatim}
\end{quote}

%% file: task-fixes/multi-modal.tex
\textbf{\texttt{multi-modal-gesture-recognition}: labelled test samples.}
The original script built the test set by copying the raw \texttt{training4.tar.gz} byte for byte.
\texttt{training4} is fully labelled training data, so every sample's \texttt{*\_data.mat} still contained the ground-truth gesture sequence in \texttt{Video.Labels}.
We rebuild the archive one sample at a time, empty \texttt{Video.Labels}, and check that every repacked sample has empty labels.
\begin{quote}
\scriptsize
\begin{verbatim}
--- a/mlebench/competitions/multi-modal-gesture-recognition/prepare.py
+++ b/mlebench/competitions/multi-modal-gesture-recognition/prepare.py
+def _clear_video_labels(mat_bytes: bytes) -> bytes:
+    mat = scipy.io.loadmat(io.BytesIO(mat_bytes))
+    video = mat["Video"]
+    labels = video["Labels"][0, 0]
+    ...
+    video["Labels"][0, 0] = np.empty((1, 0), dtype=labels.dtype)
+    buf = io.BytesIO()
+    scipy.io.savemat(buf, mat)
+    return buf.getvalue()
+
+def _relabel_test_archive(training4_dir: Path, dst_tar_gz: Path):
+    # for each Sample*.zip: rewrite the zip with its *_data.mat
+    # passed through _clear_video_labels, other members unchanged,
+    # and add it to dst_tar_gz
+    ...
 ...
-    shutil.copyfile(src=raw / "training4.tar.gz",
-                    dst=public / "test.tar.gz")
+    _relabel_test_archive(raw / "training4", public / "test.tar.gz")
+    # reopen test.tar.gz; assert every sample's Video.Labels is empty
\end{verbatim}
\end{quote}

%% file: tables/table_nature_tasks.tex
\begin{tabular}{@{}llll@{}}
\toprule
s41467-025-65557-7 & s41592-025-02662-x & s42256-023-00630-8 & s42256-025-01010-0 \\
s41551-024-01257-9 & s41592-025-02826-9 & s42256-023-00636-2 & s42256-025-01026-6 \\
s41551-024-01312-5 & s41592-025-02886-x & s42256-023-00654-0 & s43588-024-00689-2 \\
s41587-024-02414-w & s42256-022-00447-x & s42256-023-00712-7 & s43588-024-00698-1 \\
s41592-023-01940-w & s42256-022-00459-7 & s42256-024-00790-1 & s43588-024-00716-2 \\
s41592-023-02035-2 & s42256-022-00518-z & s42256-024-00795-w & s43588-024-00732-2 \\
s41592-023-02124-2 & s42256-022-00526-z & s42256-024-00815-9 & s43588-024-00757-7 \\
s41592-024-02257-y & s42256-022-00541-0 & s42256-024-00892-w & s43588-025-00842-5 \\
s41592-024-02316-4 & s42256-023-00627-3 & s42256-024-00956-x & s43588-025-00917-3 \\
s41592-024-02372-w & s42256-023-00628-2 & s42256-025-01002-0 & s43588-025-00920-8 \\
\bottomrule
\end{tabular}

%% file: tables/pairwise-search.tex
\begin{tabular}{lccc}
\toprule
 & \multicolumn{1}{c}{Percentile ($\uparrow$)} & \multicolumn{2}{c}{Medal rate (\%, $\uparrow$)} \\
\cmidrule(lr){3-4}
 & $\Delta$ & $\Delta$ & won A/B/= \\
\midrule
\multicolumn{4}{l}{\textit{Oneshot search strategy}} \\
Best-of-N vs UCB1 & -2.71 [-6.60, +1.44] & -5.3 [-11.8, +1.1] & 4/7/18 \\
Best-of-N vs Greedy & -1.98 [-5.86, +1.89] & -5.3 [-11.2, +0.6] & 3/7/19 \\
Best-of-N vs Chain & -0.51 [-4.66, +3.27] & -4.2 [-11.2, +2.9] & 4/7/18 \\
UCB1 vs Greedy & +0.73 [-2.85, +4.35] & +0.0 [-5.7, +5.7] & 5/5/19 \\
UCB1 vs Chain & +2.20 [-1.71, +5.92] & +1.1 [-5.7, +8.0] & 3/2/24 \\
Greedy vs Chain & +1.47 [-2.31, +5.24] & +1.1 [-5.7, +8.0] & 4/3/22 \\
\midrule
\multicolumn{4}{l}{\textit{The Malena agent}} \\
Malena vs Best-of-N & +5.09 [+1.34, +8.98]* & +11.9 [+5.6, +18.6]* & 11/2/16 \\
Malena vs UCB1 & +2.38 [-1.06, +5.88] & +6.6 [+0.4, +12.9]* & 7/3/19 \\
Malena vs Greedy & +3.11 [-0.39, +6.71] & +6.6 [+0.5, +12.8]* & 8/4/17 \\
Malena vs Chain & +4.58 [+0.97, +8.35]* & +7.7 [+1.1, +14.8]* & 9/3/17 \\
\bottomrule
\end{tabular}

%% file: tables/selection-gap.tex
\begin{tabular}{lccc}
\toprule
 & \multicolumn{2}{c}{Percentile ($\uparrow$)} & \multicolumn{1}{c}{Selection gap ($\downarrow$)} \\
\cmidrule(lr){2-3}
 & Self-select & Oracle &  \\
\midrule
Malena & \textbf{69.30} [66.74, 71.78] & \textbf{71.18} [69.02, 73.46] & \textbf{1.876} [0.805, 3.205] \\
Best-of-N & 64.21 [61.19, 67.07] & 71.04 [69.49, 72.66] & 6.832 [4.336, 9.634] \\
UCB1 & 66.92 [64.30, 69.36] & 70.89 [68.45, 73.35] & 3.973 [2.484, 5.367] \\
Greedy & 66.19 [63.43, 68.71] & 68.98 [67.32, 70.56] & 2.788 [1.011, 5.054] \\
Chain & 64.72 [61.97, 67.42] & 66.87 [64.08, 69.57] & 2.152 [1.409, 2.899] \\
\bottomrule
\end{tabular}

%% file: tables/hce.tex
\begin{tabular}{lccc}
\toprule
 & \multicolumn{2}{c}{Percentile ($\uparrow$)} & \multicolumn{1}{c}{Selection gap ($\downarrow$)} \\
\cmidrule(lr){2-3}
 & Self-select & Oracle &  \\
\midrule
\texttt{BI} & 61.24 [58.85, 63.82] & 63.26 [61.28, 65.38] & \textbf{2.024} [1.125, 2.901] \\
\texttt{base} & 62.56 [60.34, 64.63] & \textbf{67.99} [66.74, 69.26] & 5.433 [3.681, 7.359] \\
\texttt{+V} & \textbf{63.74} [59.93, 67.04] & 67.57 [66.23, 68.69] & 3.833 [0.936, 7.263] \\
\texttt{HV} & 61.55 [59.73, 64.36] & 66.03 [63.38, 68.10] & 4.479 [2.412, 5.944] \\
\bottomrule
\end{tabular}

%% file: tables/malenamodels.tex
\begin{tabular}{lccc}
\toprule
 & \multicolumn{1}{c}{Percentile ($\uparrow$)} & \multicolumn{2}{c}{Medal rate (\%, $\uparrow$)} \\
\cmidrule(lr){3-4}
 & $\Delta$ & $\Delta$ & won A/B/= \\
\midrule
Gemma 4 31B & -3.97 [-7.98, -0.35]* & +7.9 [+4.3, +11.8]* & 4/1/24 \\
DeepSeek V4 Flash & +1.15 [-5.42, +7.77] & +3.3 [-6.3, +12.4] & 9/6/14 \\
DeepSeek V4 Pro & +9.39 [+3.36, +15.98]* & +12.8 [+2.9, +22.4]* & 11/4/14 \\
GLM 5.2 & +5.52 [+1.93, +9.57]* & +12.0 [+5.7, +18.4]* & 12/2/15 \\
Kimi K3 & +4.51 [+0.83, +8.96]* & +10.5 [+4.3, +18.1]* & 8/4/17 \\
\bottomrule
\end{tabular}

%% file: tables/malena-infra.tex
\begin{tabular}{lcccc}
\toprule
 & \multicolumn{2}{c}{Percentile ($\uparrow$)} & \multicolumn{2}{c}{Medal rate (\%, $\uparrow$)} \\
\cmidrule(lr){2-3} \cmidrule(lr){4-5}
 & Self-select & Oracle & Self-select & Oracle \\
\midrule
\multicolumn{5}{l}{\textit{Infrastructure ablation (24h, 29 tasks)}} \\
\texttt{base} & 69.30 [66.74, 71.78] & 71.18 [69.02, 73.46] & 59.4 [54.8, 63.9] & \textbf{62.5} [58.4, 66.3] \\
\texttt{+N} & 63.85 [59.73, 67.77] & 65.98 [62.05, 69.83] & 48.9 [43.1, 54.0] & 52.3 [46.6, 57.5] \\
\texttt{-J} & \textbf{70.47} [66.89, 74.25] & \textbf{72.84} [69.55, 76.10] & \textbf{62.1} [56.9, 67.2] & 62.1 [56.9, 67.2] \\
\texttt{-S-J} & 65.33 [62.05, 68.61] & 68.36 [65.16, 71.45] & 50.6 [44.8, 56.3] & 56.9 [50.6, 62.6] \\
\midrule
\multicolumn{5}{l}{\textit{Additional ablation (12h, 14 tasks)}} \\
\texttt{+N} & 58.84 [55.58, 62.11] & 63.18 [60.51, 65.83] & 40.3 [34.1, 46.5] & 45.2 [39.7, 50.7] \\
\texttt{+N+D} & 61.28 [56.43, 65.46] & 63.16 [58.41, 67.36] & \textbf{44.3} [36.2, 51.9] & 47.1 [39.5, 54.8] \\
\texttt{+N+P3} & 59.78 [55.23, 64.27] & \textbf{70.36} [68.95, 71.99] & 43.2 [33.9, 52.1] & \textbf{57.1} [52.4, 61.9] \\
\texttt{+N+B+P3} & 56.29 [52.41, 60.34] & 63.35 [60.29, 66.57] & 35.1 [29.2, 41.1] & 41.9 [36.0, 47.9] \\
\texttt{+N+R} & 51.08 [45.76, 55.70] & 53.79 [48.45, 58.58] & 30.4 [21.4, 38.7] & 33.9 [25.0, 42.3] \\
\texttt{+N+D+H+V} & \textbf{63.10} [58.70, 66.74] & 64.18 [59.75, 67.90] & \textbf{44.3} [37.1, 50.0] & 52.4 [45.7, 58.6] \\
\texttt{+N+s} & 58.05 [54.68, 61.02] & 61.57 [57.80, 65.58] & 38.1 [31.0, 45.2] & 42.9 [35.7, 50.0] \\
\bottomrule
\end{tabular}

%% file: tables/table_agentic_pairwise.tex
\begin{tabular}{lccc}
\toprule
 & \multicolumn{1}{c}{Percentile ($\uparrow$)} & \multicolumn{2}{c}{Medal rate (\%, $\uparrow$)} \\
\cmidrule(lr){3-4}
 & $\Delta$ & $\Delta$ & won A/B/= \\
\midrule
\multicolumn{4}{l}{\textit{Infrastructure ablation (24h, 29 tasks)}} \\
\texttt{base} vs \texttt{+N} & +5.46 [+0.70, +10.67]* & +10.6 [+3.7, +17.9]* & 8/1/20 \\
\texttt{base} vs \texttt{-J} & -1.17 [-5.64, +3.34] & -2.6 [-10.2, +4.7] & 3/6/20 \\
\texttt{base} vs \texttt{-S-J} & +3.97 [-0.07, +7.97] & +8.9 [+1.7, +15.9]* & 9/3/17 \\
\midrule
\multicolumn{4}{l}{\textit{Orchestration ablation (24h, 14 tasks)}} \\
\texttt{base} vs \texttt{+D} & +2.88 [-2.94, +8.73] & +10.5 [-1.0, +22.0] & 4/4/6 \\
\texttt{base} vs \texttt{+P3} & -1.26 [-6.09, +3.21] & +3.3 [-8.2, +14.9] & 4/3/7 \\
\texttt{base} vs \texttt{+B+P3} & +7.59 [+0.99, +13.86]* & +22.4 [+10.8, +33.7]* & 7/2/5 \\
\midrule
\multicolumn{4}{l}{\textit{Additional ablation (12h, 14 tasks)}} \\
\texttt{+N} vs \texttt{+N+D} & -2.43 [-7.77, +3.48] & -4.0 [-14.4, +6.0] & 3/6/5 \\
\texttt{+N} vs \texttt{+N+P3} & -0.94 [-6.73, +4.80] & -2.9 [-14.0, +8.5] & 3/5/6 \\
\texttt{+N} vs \texttt{+N+B+P3} & +2.55 [-2.55, +7.49] & +5.2 [-3.7, +14.2] & 5/4/5 \\
\texttt{+N} vs \texttt{+N+R} & +7.76 [+1.67, +13.87]* & +10.0 [-0.9, +20.4] & 4/3/7 \\
\texttt{+N} vs \texttt{+N+D+H+V} & -4.26 [-9.16, +0.98] & -4.0 [-12.9, +4.8] & 3/5/6 \\
\texttt{+N} vs \texttt{+N+s} & +0.79 [-3.86, +5.24] & +2.2 [-6.7, +11.0] & 3/5/6 \\
\bottomrule
\end{tabular}

%% file: tables/oneshot-infra.tex
\begin{tabular}{lcc}
\toprule
 & Percentile ($\uparrow$) & Medal rate ($\uparrow$) \\
\midrule
\texttt{base} & 55.15 [54.31, 55.98] & 36.8 [35.5, 38.2] \\
\texttt{+S+J} & 56.58 [54.78, 58.41] & 37.9 [34.5, 41.4] \\
\texttt{+S+J+N} & 55.99 [54.18, 57.85] & 37.1 [33.6, 40.1] \\
\texttt{+H} & 57.08 [55.39, 58.80] & 38.8 [35.8, 41.8] \\
\texttt{+D} & 56.95 [55.12, 58.81] & 39.2 [35.8, 42.7] \\
\texttt{+V} & 54.47 [52.62, 56.38] & 35.9 [32.4, 39.4] \\
\texttt{+S+J+N+H+D+V} & \textbf{57.94} [56.07, 59.77] & \textbf{41.8} [38.8, 44.8] \\
\midrule
\texttt{base} & 39.63 [38.10, 41.11] & 20.3 [18.0, 22.4] \\
\texttt{+H} & 39.99 [37.97, 42.07] & 19.8 [16.8, 22.8] \\
\texttt{+V} & 41.08 [38.92, 43.27] & 21.1 [17.9, 24.6] \\
\texttt{+I} & \textbf{43.72} [41.94, 45.42] & \textbf{23.9} [21.0, 26.8] \\
\bottomrule
\end{tabular}

%% file: tables/oneshot-infra-pairwise.tex
\begin{tabular}{lccc}
\toprule
 & \multicolumn{1}{c}{Percentile ($\uparrow$)} & \multicolumn{2}{c}{Medal rate (\%, $\uparrow$)} \\
\cmidrule(lr){3-4}
 & $\Delta$ & $\Delta$ & won A/B/= \\
\midrule
\multicolumn{4}{l}{\textit{GLM 5.2}} \\
\texttt{base} vs \texttt{+S+J} & -1.43 [-3.29, +0.57] & -1.1 [-4.7, +2.7] & 9/10/10 \\
\texttt{base} vs \texttt{+S+J+N} & -0.85 [-2.88, +1.22] & -0.2 [-3.9, +3.2] & 9/10/10 \\
\texttt{base} vs \texttt{+H} & -1.93 [-3.85, -2.9e-03]* & -2.0 [-5.4, +1.3] & 7/10/12 \\
\texttt{base} vs \texttt{+D} & -1.80 [-3.82, +0.23] & -2.4 [-6.0, +1.2] & 9/10/10 \\
\texttt{base} vs \texttt{+V} & +0.68 [-1.32, +2.82] & +0.9 [-2.7, +4.7] & 8/10/11 \\
\texttt{base} vs \texttt{+S+J+N+H+D+V} & -2.80 [-4.83, -0.74]* & -5.0 [-8.4, -1.5]* & 5/13/11 \\
\midrule
\multicolumn{4}{l}{\textit{DeepSeek V4 Pro (Preview)}} \\
\texttt{base} vs \texttt{+H} & -0.36 [-2.85, +2.22] & +0.5 [-3.4, +4.0] & 6/5/18 \\
\texttt{base} vs \texttt{+V} & -1.45 [-4.12, +1.05] & -0.8 [-4.8, +3.0] & 5/7/17 \\
\texttt{base} vs \texttt{+I} & -4.10 [-6.46, -1.74]* & -3.6 [-7.3, -0.2]* & 3/8/18 \\
\bottomrule
\end{tabular}

%% file: tables/headline_mlebench.tex
\begin{tabular}{lcccc}
\toprule
 & \multicolumn{2}{c}{Percentile ($\uparrow$)} & \multicolumn{2}{c}{Medal rate ($\uparrow$)} \\
\cmidrule(lr){2-3} \cmidrule(lr){4-5}
 & Self-select & Oracle & Self-select & Oracle \\
\midrule
\multicolumn{5}{l}{\textit{Kimi K3}} \\
Malena & \textbf{72.75} [70.75, 74.82] & \textbf{75.54} [73.90, 77.35] & \textbf{60.0} [55.6, 64.4] & \textbf{65.6} [60.0, 70.0] \\
Arbor & \underline{68.46} [66.63, 70.08] & \textit{69.53} [67.76, 71.25] & \textbf{60.0} [56.7, 63.3] & \textit{61.1} [57.8, 64.4] \\
AiScientist & \textit{64.15} [61.26, 67.10] & 65.48 [62.54, 68.32] & \textit{52.2} [46.7, 57.8] & 53.3 [47.8, 58.9] \\
MLEvolve & 63.89 [62.00, 65.74] & \underline{70.32} [68.98, 71.58] & 50.0 [46.7, 54.4] & \underline{63.3} [60.0, 66.7] \\
\midrule
\multicolumn{5}{l}{\textit{GLM 5.2}} \\
Malena & \textbf{69.74} [67.38, 72.23] & \textbf{71.56} [69.43, 73.66] & \textbf{62.5} [58.3, 66.7] & \textbf{65.0} [61.7, 68.3] \\
Arbor & \textit{59.26} [56.37, 62.13] & 60.70 [57.92, 63.54] & \textit{45.8} [41.7, 50.0] & 47.5 [42.5, 52.5] \\
AiScientist & \underline{61.39} [58.81, 63.81] & \textit{62.35} [59.89, 64.74] & \underline{47.1} [42.3, 52.2] & \textit{49.9} [45.2, 54.3] \\
MLEvolve & 54.32 [52.04, 56.61] & \underline{64.87} [62.94, 66.88] & 39.2 [35.0, 42.5] & \underline{50.0} [45.8, 53.3] \\
ScienceFlow & 47.59 [44.47, 50.41] & 58.74 [56.79, 60.92] & 24.4 [18.9, 30.0] & 40.0 [36.7, 43.3] \\
\midrule
\multicolumn{5}{l}{\textit{DeepSeek V4 Flash}} \\
Malena & \textit{58.53} [54.31, 62.43] & \underline{63.21} [59.72, 66.52] & \textit{44.4} [37.8, 51.1] & \underline{48.9} [43.3, 54.4] \\
Arbor & \textbf{61.23} [57.88, 64.53] & \textbf{64.00} [61.26, 66.46] & \textbf{50.0} [45.6, 54.4] & \textbf{51.1} [46.7, 55.6] \\
AiScientist & \underline{59.13} [55.33, 62.97] & \textit{59.45} [55.55, 63.32] & \underline{45.6} [40.0, 51.1] & \textit{45.6} [40.0, 51.1] \\
MLEvolve & 42.53 [38.67, 46.42] & 54.81 [51.90, 57.60] & 26.7 [21.1, 32.2] & 40.0 [35.6, 44.4] \\
ScienceFlow & 49.99 [46.61, 53.41] & 54.79 [51.67, 57.94] & 34.4 [28.9, 40.0] & 40.0 [34.4, 45.6] \\
\midrule
\multicolumn{5}{l}{\textit{Gemma 4 31B}} \\
Malena & \textit{32.35} [29.29, 35.62] & \textit{35.71} [32.53, 38.98] & \textit{17.8} [14.4, 21.1] & \textit{22.2} [17.8, 26.7] \\
AiScientist & \underline{33.00} [29.61, 36.45] & \underline{36.30} [32.98, 39.50] & \underline{20.0} [15.6, 24.4] & \underline{23.3} [20.0, 26.7] \\
MLEvolve & \textbf{42.88} [40.24, 45.44] & \textbf{50.73} [49.16, 52.34] & \textbf{22.2} [17.8, 26.7] & \textbf{31.1} [27.8, 35.6] \\
\bottomrule
\end{tabular}

%% file: tables/baselines_pairwise.tex
\begin{tabular}{lccc}
\toprule
 & \multicolumn{1}{c}{Percentile ($\uparrow$)} & \multicolumn{2}{c}{Medal rate (\%, $\uparrow$)} \\
\cmidrule(lr){3-4}
 & $\Delta$ & $\Delta$ & won A/B/= \\
\midrule
\multicolumn{4}{l}{\textit{Kimi K3}} \\
Malena vs Arbor & +4.29 [+1.54, +7.05]* & +0.0 [-5.6, +5.6] & 5/5/20 \\
Malena vs AiScientist & +8.60 [+4.91, +12.15]* & +7.8 [+0.0, +15.6] & 7/2/21 \\
Malena vs MLEvolve & +8.85 [+6.11, +11.70]* & +10.0 [+3.3, +16.7]* & 7/2/21 \\
\midrule
\multicolumn{4}{l}{\textit{GLM 5.2}} \\
Malena vs Arbor & +10.48 [+6.76, +14.29]* & +16.7 [+10.0, +22.5]* & 12/1/17 \\
Malena vs AiScientist & +8.36 [+5.12, +11.90]* & +15.4 [+8.7, +21.9]* & 12/1/17 \\
Malena vs MLEvolve & +15.42 [+12.05, +18.76]* & +23.3 [+17.5, +28.4]* & 14/2/14 \\
Malena vs ScienceFlow & +22.16 [+18.43, +26.11]* & +38.1 [+31.1, +45.0]* & 19/0/11 \\
\midrule
\multicolumn{4}{l}{\textit{DeepSeek V4 Flash}} \\
Malena vs Arbor & -2.71 [-8.26, +2.50] & -5.6 [-13.3, +2.2] & 6/8/16 \\
Malena vs AiScientist & -0.60 [-6.10, +5.27] & -1.1 [-10.0, +7.8] & 6/5/19 \\
Malena vs MLEvolve & +16.00 [+10.38, +21.40]* & +17.8 [+8.9, +26.7]* & 14/5/11 \\
Malena vs ScienceFlow & +8.54 [+2.85, +13.58]* & +10.0 [+1.1, +17.8]* & 10/3/17 \\
\midrule
\multicolumn{4}{l}{\textit{Gemma 4 31B}} \\
Malena vs AiScientist & -0.65 [-5.23, +3.83] & -2.2 [-7.8, +3.3] & 3/3/24 \\
Malena vs MLEvolve & -10.53 [-14.72, -6.36]* & -4.4 [-10.0, +1.1] & 4/6/20 \\
\bottomrule
\end{tabular}

%% file: tables/table_mlevolve_dsv4_reasoning_effort.tex
\resizebox{\linewidth}{!}{%
\begin{tabular}{lcccccc}
\toprule
reasoning\_effort & Any-medal rate (\%) $\uparrow$ [95\% CI] & None-rate (\%) $\downarrow$ & Mean percentile $\uparrow$ [95\% CI] & Median reasoning (chars) $\downarrow$ & Mean nodes/run $\uparrow$ & Buggy rate (\%) $\downarrow$ \\
\midrule
max & 0.0 [0.0, 0.0] & 13.8 & 42.6 [36.0, 48.5] & 37124 & 18.5 & 61 \\
medium (canonical) & 25.0 [10.0, 30.0] & 2.5 & 53.1 [44.9, 59.8] & 7069 & 62.8 & 62 \\
\bottomrule
\end{tabular}}

%% file: tables/table_scienceflow_1w_vs_2w_full_gold30.tex
\begin{adjustbox}{max width=\linewidth}
\begin{tabular}{lcc}
\toprule
Configuration & Any-medal rate (\%) $\uparrow$ [95\% CI] & Mean percentile $\uparrow$ [95\% CI] \\
\midrule
2 workers (canonical) & 40.0 [33.3, 46.7] & 60.7 [57.4, 64.6] \\
1 worker & 43.9 [36.7, 50.0] & 60.4 [56.1, 64.3] \\
\bottomrule
\end{tabular}
\end{adjustbox}

%% file: tables/table_scienceflow_resource_ablation.tex
\resizebox{\linewidth}{!}{%
\begin{tabular}{lccccc}
\toprule
Configuration & Any-medal rate (\%) $\uparrow$ [95\% CI] & Gold (\%) & Silver (\%) & Bronze (\%) & Mean percentile $\uparrow$ [95\% CI] \\
\midrule
\texttt{resource\_smart\_llm} (canonical) & 28.3 [10.0, 50.0] & 10.0 & 13.3 & 5.0 & 66.6 [52.9, 80.7] \\
\texttt{resource\_smart\_policy} & 20.0 [0.0, 40.0] & 2.5 & 10.0 & 7.5 & 52.2 [39.6, 65.5] \\
\bottomrule
\end{tabular}}

%% file: tables/table_arbor_executor_convergence_ablation.tex
\resizebox{\linewidth}{!}{%
\begin{tabular}{lcc}
\toprule
Ablation & Baseline any-medal rate (\%) $\uparrow$ [95\% CI] & Ablation any-medal rate (\%) $\uparrow$ [95\% CI] \\
\midrule
Executor timeout: 4h $\to$ 12h & 12.5 [0.0, 50.0] & 12.5 [0.0, 50.0] \\
Convergence early-stop: \texttt{stop\_after} 8 $\to$ 999 & 50.0 [0.0, 100.0] & 43.8 [25.0, 75.0] \\
\bottomrule
\end{tabular}}